\documentclass[opre,nonblindrev]{informs3} 

\DoubleSpacedXI 

\usepackage{endnotes}
\let\footnote=\endnote

\usepackage{natbib}
 \bibpunct[, ]{(}{)}{,}{a}{}{,}%
 \def\bibfont{\small}%
\TheoremsNumberedThrough     
\ECRepeatTheorems

\EquationsNumberedThrough    

\usepackage{hyperref, bbm, nicefrac, makecell, svg, subcaption}
\usepackage{algorithm, algorithmicx, float}
\usepackage[noend]{algpseudocode}
\def\arginf{\mathop{\rm arg\,inf}}%
\def\argsup{\mathop{\rm arg\,sup}}%
\newcommand\omicron{o}
\begin{document}


\RUNAUTHOR{Dogan et. al.}

\RUNTITLE{When the Agent Learns Too: Learning to Incentivize with Unobserved Rewards}

\TITLE{When the Agent Learns Too: Learning to Incentivize with Unobserved Rewards}

\ARTICLEAUTHORS{%
\AUTHOR{Ilgin Dogan}
\AFF{Department of Industrial Engineering and Operations Research, University of California, Berkeley, Berkeley, California 94720. Current affiliation: INSEAD, Fontainebleau, France, \EMAIL{ilgin.dogan@insead.edu}}  
\AUTHOR{Zuo-Jun Max Shen}
\AFF{Department of Industrial Engineering and Operations Research, University of California, Berkeley, Berkeley, California 94720, \EMAIL{maxshen@berkeley.edu}}
\AUTHOR{Anil Aswani}
\AFF{Department of Industrial Engineering and Operations Research, University of California, Berkeley, Berkeley, California 94720, \EMAIL{aaswani@berkeley.edu}}
} 

\ABSTRACT{%
In practice, incentive providers (i.e., principals) often cannot observe the reward realizations of incentivized agents, which is in contrast to many principal-agent models that have been previously studied. This information asymmetry challenges the principal to consistently estimate the agent's unknown rewards by solely watching the agent's decisions, which becomes even more challenging when the agent has to learn its own rewards. This complex setting is observed in various real-life scenarios ranging from renewable energy storage contracts to personalized healthcare incentives. Hence, it offers not only interesting theoretical questions but also wide practical relevance. This paper explores a repeated adverse selection game between a self-interested learning agent and a learning principal. The agent tackles a multi-armed bandit (MAB) problem to maximize their expected reward plus incentive. On top of the agent's learning, the principal trains a parallel algorithm and faces a trade-off between consistently estimating the agent's unknown rewards and maximizing their own utility by offering adaptive incentives to lead the agent. For a non-parametric model, we introduce an estimator whose only input is the history of principal's incentives and agent's choices. We unite this estimator with a proposed data-driven incentive policy within a MAB framework. Without restricting the type of the agent's algorithm, we prove finite-sample consistency of the estimator and a rigorous regret bound for the principal by considering the sequential externality imposed by the agent. Lastly, our theoretical results are reinforced by simulations justifying applicability of our framework to green energy aggregator contracts.  
}%



\maketitle

%


\section{Introduction}
Sequential decision-making in many operations management problems requires studying the complex interactions between decision-makers who are challenged with optimizing their own objectives under the uncertainty inherent in the environment of interest. \emph{Repeated principal-agent theory} is a well-established paradigm that studies  sequential interactions between two self-interested decision-makers. In particular, it offers a framework to analyze the problem of a primary party (i.e., principal) in a system who seeks to optimize the ultimate performance of the system by repeatedly delegating some operational control to another strategic party (i.e., agent) with a private decision-making process. This privacy imposes an information asymmetry between the principal and the agent that can appear as either an \emph{adverse selection} setting, in which the information about the agent's true preferences or rewards are hidden from the principal, or a \emph{moral hazard} setting, in which the actions chosen by the agent are hidden from the principal \citep{bolton2004contract}. In either case, the principal's problem can be defined along two main dimensions: 
\begin{enumerate} 
    \item[] \textit{i)} learning some private information about the agent by training a consistent estimator,
    \item[] \textit{ii)} designing an incentive mechanism to lead the agent's algorithm in favor of the principal.
\end{enumerate} 
In this paper, we study these two research problems for an unexplored adverse selection setting by marrying the classical principal-agent theory to statistics and reinforcement learning. 

In a repeated principal-agent game, the main theoretical challenge is sourced from the dynamic and sequential interactions taking place between the two strategic decision-makers. In each play of the game, first the principal offers a menu of incentives to the agent, and then the agent makes a choice from a finite set of actions, which in turn determines the rewards collected by both players. In other words, there is a two-sided sequential externality in this setting, whereby the agent's imperfect knowledge imposes additional costs on the principal and the principal's incentives impose a more challenging decision-making environment for the agent with imperfect knowledge. This paper considers that both the principal and the agent observe stochastic rewards with unknown (to both) expectations, and that both parties aim to maximize their own cumulative expected rewards at the end of the game. This implies they are not only utility-maximizers but also learning players who are estimating their own reward models while interacting with each other. 

Existing work on repeated principal-agent theory mostly studies the moral hazard setting, with a more recent focus on the problem of estimating agent's unknown model parameters (e.g., \citealt{ho2016}, \citealt{kaynar2022estimating}). On the other hand, the adverse selection model is typically studied either for single-period static games \citep{navabi2018optimal, 
 chade2019disentangling, gottlieb2022simple} or else for the repeated dynamic games where restrictive assumptions are made on, for example, dimension of the agent's action space, utility function of the agent, and relationship between principal's rewards and agent's unknown type (e.g., \citealt{halac2016optimal, esHo2017dynamic, maheshwari2022inducing}). Furthermore, the statistical estimation and learning problem has not yet been explored for the repeated adverse selection setting. To the best of our knowledge, we are not aware of any related work on this setting except a recent study by \cite{dogan2023repeated}. They analyze both the estimation and the incentive design problems faced by the principal under a specific repeated adverse selection setting which we call here the \emph{unobserved rewards} model. In this model, the principal can observe the past sequence of the actions chosen by the agent, but is uninformed about the associated reward realizations observed by the agent. Therefore, the principal needs to undertake their two main problems by only leveraging the data of their past incentives and the agent's past decisions. 

In this paper, we will study the two research problems described above under the repeated unobserved rewards setting, but in a more challenging environment than the one in \cite{dogan2023repeated}. The main difference between this paper and that work is regarding the agent's behavior. That work assumes the agent has perfect knowledge of their reward model and picks the true utility-maximizing action at each period. However, here we consider a learning agent that tackles a multi-armed bandit (MAB) problem to acquire the knowledge of their true utility-maximizers under the offered incentives. The implication is that the principal's learning algorithm is trained on top of the agent's learning process in the considered setting of this paper. Our statistical and regret analyses show that this consideration yields a substantially higher complexity than the ``perfect agent'' setting. The main challenge comes from the uncertainty in the agent's choices for which now there is no guarantee to be the true maximizers of the agent's rewards. Both the estimation errors and cost of explorations incurred by the agent over the horizon directly contribute to the cumulative regret of the principal. For this complex setting, we offer a generic and data-driven framework without necessarily restricting the type of the agent's MAB algorithm and prove that our incentive policy attains a sublinear regret for the principal. 
\subsection{Motivating Real-Life Applications}
\subsubsection{Adaptive Aggregation for Sustainable Energy}\label{subsubsec:aggregator}
Among many attempts to address the devastating and growing impact of the climate crisis,  transition to clean green energy stands as one of the most effective and widespread processes. Today, there is a rising awareness on the value of renewable and large-scale distributed energy storage for the clean energy transition. With this motivation, the so-called ``aggregators'' have started to play a key role in electricity markets. An aggregator is a company operating a grid-scale virtual power plant that pools energy supply available in their distributed battery systems and sells this capacity in electricity markets during peak demand or emergency periods \citep{irena2019, sunrun2021, berntzen2021aggregator}. This aggregation operation provides substantial advantages to the market stakeholders, the utility companies, and the independent aggregator firms on the supply side and the residential and commercial customers on the demand side. The primary benefits can be summarized as reducing costs for utilities and communities, decreasing carbon emissions, and improving power reliability. To achieve these benefits, the aggregator has to motivate the customers to be flexible and allow the aggregator to use the backup power available in their electric vehicles or solar energy storage systems \citep{bindra2018storage, biggins2022going}. For this purpose, the aggregator offers benefits to the participating customers such as a compensation for their contribution to the energy supply in the grid. On the higher level, it falls to the utility company to encourage the aggregator to initiate flexibility in their storage capacity through increasing their investments and managing voluntary customer participation in the aggregation program. 

The sequential game between the utility company (i.e., principal) and independent aggregator company (i.e., agent) can be effectively modeled as a repeated unobserved rewards setting with two strategic and learning players. In this model, the actions of the aggregator is defined as the amount of storage capacity (MW/h) (sourced from electric vehicles or household heat pumps) reserved for the use of utility grid, and the payments of the utility company are defined as the service fees offered for purchased storage (MW/h) in an aggregation contract (which is organized typically on an hourly basis). The realized profits of the utility and the aggregator observed as a result of these contracts depend on several sources of uncertainty including variations in renewable energy generation (e.g., wind, solar, hydro, etc.), electricity demand, and market prices. Such contracts are utilized in recent applications such as the Emergency Load Reduction program launched by Tesla and PG\&E \citep{utility2022} and the Resilient Home program initiated by Sunrun and East Bay Community Energy in California \citep{sunrun2020}. Taking into account the conflicting objectives of multiple stakeholders and unknown stochastic components in these systems, we believe that the adaptive incentive policies proposed in this paper can help the utility company to offer smart contracts that explicitly considers their sequential interactions with the aggregator and expands the participation of the aggregator.

\subsubsection{Personalized Incentive Design for Medical Adherence}\label{subsubsec:medical}
The problem of patients not following medication dosing instructions is recognized as a major and widespread issue around the world. Such lack of adherence to a medication regime leads to not only poor health outcomes but also substantial financial costs \cite{osterberg2005adherence}. According to \cite{world2003adherence}, medication non-adherence is observed 50\% of the time, which may increase up to 80\% for relatively asymptomatic diseases such as hypertension \citep{brown2016medication}. Research reveals various reasons for this problem including individual-level factors (e.g., medication side effects), social factors (e.g., cultural beliefs), and economic factors (e.g., transportation costs to clinics) \citep{world2003adherence, bosworth2010medication, long2011patient}. To overcome some of these concerns, incentive programs that provide financial rewards to the patients are commonly employed and shown to effectively improve medical adherence. There is a related literature in medicine and economics on examining the effects of these monetary incentives using empirical analyses \citep{lagarde2007conditional, gneezy2011and} and in operations management on quantitatively designing the financial incentives for different market contexts \citep{lobo2017using, aswani2018inverse, ghamat2018contracts, guo2019impact, suen2022design}. 

The design of financial incentives throughout a medication regime with finite length adequately features a repeated principal-agent problem under the unobserved rewards setting that we introduce in this work. Given their personal preferences and characteristics (i.e., type), the patient (i.e., agent) exhibits certain adherence behaviors in order to maximize their total utility which is comprised of benefits obtained through the improvements in their health conditions, costs incurred due the adherence, and incentives offered by the healthcare provider. On the other hand, the goal of the healthcare provider (i.e., principal) is to maximize the clearance rate, that is the rate at which the infected patient is recovered, by designing motivating payments to the utility-maximizing patient to improve their adherence actions. This payment design problem is nontrivial due to scarce clinical resources and the information asymmetry between the provider and the patient. Although the healthcare provider can often observe the patient's adherence decisions, the type of the patient (and hence the patient's utilities) often stands as private information to the provider. Because the data-driven incentive design framework presented in this study is based on a generic model without any restrictive technical assumptions, we believe that it is useful and fits well to the practical setting for the problem of medical non-adherence. 
\subsection{Related Literature}
\subsubsection{Repeated Principal-Agent Models}
There is a rich and extensive literature on principal-agent models in economics \citep{holmstrom1979moral, grossman1983, hart1987theory} and in operations management \citep{martimort2009theory}. For repeated models, most existing studies focuses on the moral hazard setting \citep{radner1981monitoring, rogerson1985repeated, spear1987repeated, abreu1990toward, plambeck2000performance, conitzer2006, sannikov2008, sannikov2013}. Several study the problem of estimating the agent's model when actions are hidden \citep{vera2003structural, misra2005salesforce, misra2011structural, ho2016, kaynar2022estimating}. On the other hand, related work on the design of incentives under the adverse selection setting is relatively scarce. In many of them, the agent's type (e.g., level of effort or probability of being successful) is considered as an additional, unknown information on top of a moral hazard setting \citep{dionne85, banks1993adverse, gayle2015identifying, williams2015solvable, halac2016optimal, esHo2017dynamic}. Only a few of these works study the estimation problem for the hidden type setting, and they use statistical estimation methods such as least squares approximation \citep{leezenios2012}, minimization of a sum of squared criterion function \citep{gayle2015identifying}, and simulation-based maximum likelihood estimation \citep{aswani2019data, mintz2023behavioral}. However, the adverse selection setting studied in these papers comes with limiting assumptions such as that the agent's type parameter belongs to a discrete set. 

Our work differs from these studies in several ways. The estimation problem we consider in our setting involves estimating the reward expectation values which belong to a bounded continuous space. Differently from the existing work summarized above, our estimation approach consists of solving a practical, slack-minimization optimization problem. Furthermore, regarding the incentive design problem, past papers do not consider the exploration-exploitation trade-off faced by the principal, and hence, they are not able to provide guarantees on how close to optimal their solutions are. In this paper, we take a sequential learning approach to solve the incentive design problem and perform a regret analysis for the considered repeated adverse selection models. The closest existing work to ours is a recent study by \citealt{dogan2023repeated} in which the repeated unobserved rewards setting is first introduced. As highlighted in the introduction, the estimation and incentive design problems that we study in this paper are significantly more complex than the ones in \citealt{dogan2023repeated}, because in this past work the agent has perfect knowledge of its own rewards whereas in this paper we consider a setting where the agent must also learn its own rewards. The setting of this paper is more complex mainly because the input data to the principal's estimator and MAB algorithm carries the uncertainty and learning costs of a separate MAB trained by the agent, which essentially requires a different theoretical analysis than that of \citealt{dogan2023repeated}. 

\subsubsection{Multi-Armed Bandits for Auction Design}
A related line of research from sequential decision-making includes the use of a multi-armed bandit (MAB) framework for mechanism design. MAB's are widely applied to dynamic auction design problems which are closely related with the incentive design in dynamic principal-agent problems \citep{Nazerzadeh2008DynamicCM, Devanur2009ThePO, jain2014multiarmed, NIPS2014, biswas2015truthful, ho2016, braverman2019multi, bhat2019optimal, abhishek2020designing, shweta2020multiarmed, han2020learning, simchowitz2021, wang2022, gao2022combination}. 

In the repeated unobserved rewards game studied in this paper, both the principal's problem and the agent's problem are directly applicable to the MAB framework. The agent with imperfect knowledge of their reward expectations associated with a finite set of actions aims to gain this knowledge by exploring their action space, but without damaging their cumulative expected utility (i.e., agent's reward plus incentive) at the end of the game. In parallel to the agent, the principal intends to maximize their own cumulative expected utility (i.e., principal's reward minus incentive) by providing the minimum possible incentives to motivate the agent to select the arms estimated to yield the highest expected rewards, while consistently improving their estimates for every bandit arm by providing incentives that will direct the agent to select various arms. From these perspectives, the MAB framework is regarded to be useful in effectively managing the exploration-exploitation trade-offs faced by the two parties in the game. 

\subsubsection{Inverse Optimization} 
Inverse optimization is a framework for inferring parameters of an optimization model from the observed solution data \citep{ahuja2001inverse, heuberger2004inverse}. More recent work in this area probes into estimating the model of a decision-making agent by formulating the agent's model as a linear or a convex optimization problem in offline settings where data are available a priori \citep{keshavarz2011imputing, bertsimas2015data, esfahani2018data, aswani2018inverse, chan2019inverse, chan2022inverse} or in online settings where data arrive sequentially \citep{barmann2018online, dong2018generalized, dong2020inverse, maheshwari2023convergent}. Different from these studies, we do not assume any specific structure of the agent's decision-making problem, but instead we consider a utility-maximizing agent with finite action space. This case of estimating the non-parametric model of a utility-maximizing agent is also addressed by \cite{kaynar2022estimating}, who study the offline static setting of the principal-agent problem under moral hazard. A key distinction between our paper and their work is that we study the online dynamic setting of the repeated principal-agent problem under adverse selection. In accordance with the unobserved rewards setting that we examine, we design an estimator for the expected rewards of the agent's arms, whose only input is the data of actions chosen by a reward-maximizer agent in response to the provided incentives in the past. In that respect, the principal's estimation problem under the sequential unobserved rewards setting can be regarded as an analogy of online inverse optimization in MAB's.
\subsection{Outline and Contributions}
We next present an outline of this paper by featuring our main contributions to the theory and applications in the related literature areas covered above.   
\begin{description}
\item[\textit{Consistent estimator fed by the agent's MAB.}] We start by introducing our repeated principal-agent model with unobserved agent rewards in Section \ref{subsec:model}. To enhance the practical relevance of our approach, we design a generic and simple model. In accordance with this model, in Section \ref{subsec:estimator} we propose a novel estimator for the agent's expected reward for each bandit arm – which uses as data the sequence of incentives offered and subsequently chosen arms by the agent's MAB algorithm. Our estimator is formulated exactly as a linear optimization model without assuming any functional form or any specific distributional property. Using this formulation, we next prove an identifiability property and a finite-sample concentration bound of the proposed estimator under a mild assumption on the probability that the agent's MAB algorithm does not select the true utility-maximizer arms in response to the incentives offered over the time horizon. We present these statistical results in Sections \ref{subsec:identify} and \ref{subsec:consistency}.

\item[\textit{Data-driven and sequential incentive policy.}] Section \ref{subsec:algorithm} embeds our consistent estimator into a MAB framework and presents the principal's adaptive and data-driven incentive policy using a practical and computationally efficient $\epsilon$-greedy approach. By utilizing the finite-sample consistency results we derived for our estimator, we compute the regret of the proposed policy with respect to an oracle incentive policy that maximizes the principal's expected net reward at each time step under  perfect knowledge of all reward expectations. Section \ref{subsec:regret} presents a rigorous sublinear regret bound for the principal under the sequential uncertainty imposed through the agent's choices.   
\item[\textit{Agent's behavior.}] In Section \ref{sec:agent}, we present a theoretical analysis and a discussion from the perspective of the selfish learning agent. We highlight that our statistical consistency and regret bound results for the principal are proven without restricting the type or structure of the agent's MAB algorithm. As mentioned above, our only assumption about the agent's behavior is a probability bound associated with the inaccuracy of the arms chosen by their algorithm. In Section \ref{subsec:validation}, we show the mildness of our assumption by proving that it is satisfied when the agent uses a naive $\epsilon$-greedy algorithm to make their decisions throughout the sequential game. Furthermore, in Section \ref{subsec:information}, we discuss that the learning framework proposed for the principal considers a self-interested agent with particular sophistication where they are not knowledgeable about or attempting to learn the principal’s model. In that respect, our approach can be regarded as the worst case bound for the circumstances in which the agent has enough sophistication to maximize their information rent. However, regardless of the level of agent's complexity, our data-driven incentive mechanism is designed in a way that maximizes the principal's cumulative expected net reward subject to the existence of the agent's information rent. 
\item[\textit{Numerical results.}] Lastly, we conduct simulation experiments on an instance of designing sequential aggregator contracts for the green energy storage operations described earlier in the introduction. In Section \ref{sec:numerical}, we share details of our experimental setting and show that the numerical results support our theoretical results on finite-sample
concentration of the proposed estimator and on finite-sample convergence and regret of the proposed incentive policy. We demonstrate the applicability and efficiency of our framework in obtaining a renewable, reliable, and smart utility grid for communities. 
\end{description}

We conclude in Section \ref{sec:conclusion} by discussing future directions led by our analyses in this study. The proofs for all theoretical results provided in the main text are included in Appendices.
\subsection{Mathematical Notation}
We first clarify our notational conventions throughout the paper. All vectors are denoted by boldfaced lowercase letters. A vector $\mathbf{x}$ whose entries are indexed by a set $\mathcal{M} = [1, \ldots, M]$ is defined as $\mathbf{x} = (x_m)_{m \in \mathcal{M}}$. If each entry $x_m$ belongs to a set $\mathcal{X}$, then we have $\mathbf{x} \in \mathcal{X}^M$. The $\ell_\infty$-norm of the vector $\mathbf{x}$ is defined by $\|\mathbf{x} \|_\infty = \max (|x_1|, \ldots, |x_M|)$. Further, the cardinality of a set $\mathcal{X}$ is denoted by $|\mathcal{X}|$, and $\mathbbm{1}(\cdot)$ denotes the indicator function that takes value $1$ when its argument $(\cdot)$ is true, and $0$ otherwise. Lastly, the notations  $\mathbf{0}_n$ and $\mathbf{1}_n$ are used for the all-zeros and all-ones vectors of size $n$, respectively, and $\mathbb{P}(\cdot)$ is used for probabilities. 
\section{Principal's Consistent Estimator}\label{sec:estimator}
This section introduces the considered repeated adverse selection model and our novel estimator along with the associated statistical consistency results. 

\subsection{The Repeated Game with Unobserved Agent Rewards}\label{subsec:model}
We consider a sequential game between an incentive-provider principal and an incentivized agent over a finite time horizon $\mathcal{T} = [1, \ldots, T]$. The agent solves a MAB problem with a discrete set of arms (or actions) $\mathcal{A} = \{1, \ldots, n\}$. At each time step $t \in \mathcal{T}$, the principal first chooses an incentive amount $\pi_{t,a}$ for each bandit arm of the agent and offers the vector of incentives $\boldsymbol{\pi}_t = (\pi_{t,a})_{a \in \mathcal{A}}$. Then, the agent's MAB algorithm selects the arm $\upsilon_t(\boldsymbol{\pi}_t)$ which brings \textit{i)} a stochastic reward outcome to the principal denoted by $\mu_{t,\upsilon_t(\boldsymbol{\pi}_t)}$ that follows a distribution $\mathbb{F}^{\text{pr}}_{\theta^0_{\upsilon_t(\boldsymbol{\pi}_t)},\upsilon_t(\boldsymbol{\pi}_t)}$ associated with the arm $\upsilon_t(\boldsymbol{\pi}_t)$ with expectation $\theta^0_{\upsilon_t(\boldsymbol{\pi}_t)} \in \Theta$ where $\Theta$ is a known compact set, and \textit{ii)} a stochastic reward outcome to the agent denoted by $\rho_{t,\upsilon_t(\boldsymbol{\pi}_t)}$ that follows a distribution $\mathbb{F}^{\mathrm{ag}}_{r^0_{\upsilon_t(\boldsymbol{\pi}_t)}, \upsilon_t(\boldsymbol{\pi}_t)}$ associated with the arm $\upsilon_t(\boldsymbol{\pi}_t)$ with expectation $r^0_{\upsilon_t(\boldsymbol{\pi}_t)} \in \mathcal{R}$ where $\mathcal{R} = [R_{\min}, R_{\max}] \subset \mathbb{R}$ is a known compact set such that $R_{\max} - R_{\min} \geq 1$. We highlight that the principal can only observe the selected arm $\upsilon_t(\boldsymbol{\pi}_t)$ and their own net utility realization $\mu_{t, \upsilon_t(\boldsymbol{\pi}_t)} - \sum_{a \in \mathcal{A}} \pi_{t, a}$ at the end of each period. 

In this setting, the ground truth mean reward vectors $\mathbf{r}^0 = (r^0_a)_{a \in \mathcal{A}}$ and $\boldsymbol{\theta}^0 = (\theta^0_a)_{a \in \mathcal{A}}$ are unknown both to the agent and to the principal. To ensure that our research problems are well-posed, it suffices to assume that the feasible range of the principal's incentives subsumes the range of the agent's reward expectations.
\begin{assumption} \label{assm1}
The incentives $\pi_{t,a}$ for all $a \in \mathcal{A}$ belongs to a compact set $\mathcal{C} = [\underline{C}, \overline{C}]$ where $\underline{C} = R_{\min}$  and $\overline{C} = R_{\max} + \gamma$ for some constant $0 < \gamma \leq R_{\max} - R_{\min} - 1$. 
\end{assumption}

Assumption \ref{assm1} ensures that the magnitudes of the principal's incentives can be chosen sufficiently large to change the relative ordering of the bandit arms with respect to their expected rewards after adding the incentives. This assures that the principal is able to provide incentives that will steer the agent's decisions into the desirable ones. 

\subsection{The Consistent Estimator}\label{subsec:estimator}
The problem of designing adaptive incentives throughout the described repeated principal-agent play involves another challenge for the principal: estimating the agent's mean rewards $\mathbf{r}^0$ by only using the data of past incentives offered and chosen arms in response to these incentives. Before proceeding to the principal's estimator and presenting its finite-sample concentration bounds, we need to discuss and resolve an ambiguity inherent in this estimation problem. 

Because we consider an imperfect-knowledge agent, the agent also trains their own estimator to predict $\mathbf{r}^0$ and chooses their arms by following a sequential learning algorithm that aims to maximize their estimated mean reward plus the incentives offered by the principal. Therefore, the principal tries to estimate a certain reward vector such that the sequence of chosen arms are the maximizers of that vector plus the incentive vectors offered at the corresponding time periods. Suppose the principal gives the incentive vector $\boldsymbol{\pi}_\tau$ at time $\tau$. We can trivially find two different reward vectors for which the principal's estimation problem will be ill-defined. To see this, we consider the vectors $\mathbf{r}' \in \mathcal{R}^n$ and $\mathbf{r}'' = \mathbf{r}' + m \mathbf{1}_n$, where $m$ is any constant such that $\mathbf{r}'' \in \mathcal{R}^n$. Then, the key is to notice that $\argmax \ \mathbf{r}' + \boldsymbol{\pi}_\tau = \argmax \ \mathbf{r}'' + \boldsymbol{\pi}_\tau$. Since these two vectors $\mathbf{r}'$ and $\mathbf{r}''$ yield the same maximizer arm, they are not distinguishable in the considered affine space. To resolve this issue and ensure that our estimator satisfies an identifiability property, we apply a dimensionality reduction to the agent's model and define the \emph{normalized} mean reward vector $\mathbf{s}$.
\begin{definition} For a mean reward vector $\mathbf{r} = (r_1, r_2, \ldots, r_n) \in \mathcal{R}^n$, we define $\mathbf{s}$ as the normalized mean reward vector that is without loss of generality defined by $\mathbf{s}:= \mathbf{r} - r_1 \mathbf{1}_n = (0, r_2 - r_1, \ldots, r_n -r_1)$ and belongs to the compact set $\mathcal{S}^n = [R_{\min} - R_{\max}, R_{\max}- R_{\min}]^n$. 
\end{definition}

The purpose of this normalization is to eliminate one redundant dimension from the considered estimation problem and decrease our degrees of freedom by one by setting all the differences of pairs of $\mathbf{r}$'s entries with respect to a reference point 0. Here, it is essential to observe that what matters for the consistency of the principal's estimation is the differences of pairs of entries of $\mathbf{r}^0$, rather than its individual entries. This observation will allow as to derive the identifiability result for our estimator in the next subsection.

We clarify that this dimensionality reduction does not change the accuracy of our estimation because the maximizer entries of $\mathbf{r}^0 + \boldsymbol{\pi}_\tau$ and $\mathbf{s}^0 + \boldsymbol{\pi}_\tau$ are the same with each other. That is why hereby we define our estimator and present our statistical analyses with respect to the normalized reward vectors. 

As highlighted before, the principal's estimator has only two sequences of inputs: $\boldsymbol{\Pi}_t = \{\boldsymbol{\pi}_1, \boldsymbol{\pi}_2, \ldots, \boldsymbol{\pi}_{t-1}\}$ is the sequence of the incentives chosen by the principal up to time $t$ and $\Upsilon_t(\boldsymbol{\Pi}_t) = \{\upsilon_1(\boldsymbol{\pi}_1), \upsilon_2(\boldsymbol{\pi}_2), \ldots, \upsilon_{t-1}(\boldsymbol{\pi}_{t-1})\}$ is the sequence of the arms chosen by the agent up to time $t$. As we stated above, these chosen arms are based on the agent's own estimator which is trained in parallel to the principal's estimator, and hence, there is no guarantee that these arms are the true maximizers for the agent under the offered incentives. The implication of this case is that there is an additional source of uncertainty carried to the principal's estimator through the arms chosen by the agent's learning algorithm. Taking into account this uncertainty in the behaviors of the agent, we formalize the principal's estimate $\widehat{\mathbf{s}}^{\mathrm{pr}}_t\left(\Upsilon_t(\boldsymbol{\Pi}_t), \boldsymbol{\Pi}_t\right)$ at time $t$ for the agent's true normalized mean reward vector $\mathbf{s}^0$:
\begin{align}   \widehat{\mathbf{s}}^{\mathrm{pr}}_t\left(\Upsilon_t(\boldsymbol{\Pi}_t), \boldsymbol{\Pi}_t\right) \in &\argmin \hspace{0.5em}  \sum_{\tau = 1}^{t-1} y_\tau \allowdisplaybreaks  \label{estimator-1} \\
            &\hspace{0.2em} \mathrm{s.t.} \hspace{0.5em} s_{\upsilon_\tau(\boldsymbol{\pi}_\tau)}  + \pi_{\tau,\upsilon_\tau(\boldsymbol{\pi}_\tau)}  +  y_\tau \geq s_a  + \pi_{\tau,a} && \quad \forall a \in \mathcal{A}, \ \tau = 1, \ldots, t-1  \allowdisplaybreaks  \\
            &\hspace{1.9em}  y_\tau \in \mathbb{R} && \quad  \tau = 1, \ldots, t-1  \allowdisplaybreaks  \\
            &\hspace{1.9em} s_1 = 0, \ s_a \in \mathcal{S} && \quad \forall a \in \mathcal{A}  \allowdisplaybreaks \label{estimator-last}
    \end{align} 
where $y_\tau$'s are the slack variables used to contend with the agent's unknown behavior.  By introducing the loss function 
\begin{align}
    L\left(\mathbf{s}, \Upsilon_t(\boldsymbol{\Pi}_t), \boldsymbol{\Pi}_t\right) = \sum_{\tau = 1}^{t-1} \ell\left(\mathbf{s}, \upsilon_\tau(\boldsymbol{\pi}_\tau), \boldsymbol{\pi}_\tau\right) \allowdisplaybreaks \label{eq:lossfunc}
\end{align}
where $\ell\left(\mathbf{s}, \upsilon_\tau(\boldsymbol{\pi}_\tau), \boldsymbol{\pi}_\tau\right) = \max_{a \in \mathcal{A}} \left( s_a + \pi_{\tau, a} - s_{\upsilon_\tau(\boldsymbol{\pi}_\tau)} - \pi_{\tau, \upsilon_\tau(\boldsymbol{\pi}_\tau)} \right)$, we can reformulate the linear optimization problem above as 
\begin{equation} 
\label{prblm:estimator}
  \begin{alignedat}{4}
           \widehat{\mathbf{s}}^{\mathrm{pr}}_t\left(\Upsilon_t(\boldsymbol{\Pi}_t), \boldsymbol{\Pi}_t\right) \in & \argmin_{s_1 = 0, \ s_a \in \mathcal{S}, \forall a \in \mathcal{A}} L\left(\mathbf{s}, \Upsilon_t(\boldsymbol{\Pi}_t), \boldsymbol{\Pi}_t\right) \allowdisplaybreaks 
  \end{alignedat}
\end{equation}
For notational simplicity, we use the simplified notation $\widehat{\mathbf{s}}^{\mathrm{pr}}_t$ throughout the rest of this paper. 
\subsection{Identifiability}\label{subsec:identify}
The first step of our finite-sample convergence analysis for the principal's estimator (\ref{prblm:estimator}) is to prove that our estimator satisfies an identifiability property that ensures the loss function (\ref{eq:lossfunc}) is minimized uniquely by the true reward vector $\mathbf{s}^0$ \citep{van2000asymptotic}. The implication of this condition is that our estimator should be able to distinguish between $\mathbf{s}^0$ and an incorrect estimate $\mathbf{\widehat{s}}^{\mathrm{pr}}_{t}$ for a given set of incentives. As we pointed out in the previous subsection, the characterization of such incentives is based on the differences of pairs of entries of $\mathbf{s}^0$. With this observation on hand, we give our identifiability result in Proposition \ref{prop:iden4} that is proven using intermediate results in Propositions \ref{prop:iden1} -- \ref{prop:iden3}. Though these intermediate results may superficially look similar to identifiability results in Section 2.3 of \citealt{dogan2023repeated} for a perfect-knowledge agent, the results here differ in a fundamental way because we must take into account the unknown decision rule of the imperfect-knowledge agent in our paper. 

Before we present these results, we introduce the sets and notation required for our statistical analysis. We define $\mathcal{N}(\mathbf{s}^0, \beta) := \{\mathbf{s} : \| \mathbf{s} - \mathbf{s}^0 \|_\infty \leq \beta\} \subset \mathcal{S}^n$ as an open neighborhood centered around $\mathbf{s}^0$ with diameter $\beta > 0$, which further specifies the set $\mathcal{F} := \mathcal{S}^n \setminus \mathcal{N}(\mathbf{s}^0, \beta)$. We note that $\mathcal{F}$ is compact and denote an open ball centered around a vector $\mathbf{s}^j$ with diameter $d > 0$ by $\mathcal{B}(\mathbf{s}^j, d) := \{\mathbf{s} : \| \mathbf{s} - \mathbf{s}^j \|_\infty < d\}$. Then, there is a finite subcover $\{\mathcal{B}(\mathbf{s}^j, d) : \mathbf{s}^j \in \mathcal{F}\}_{j = 1}^q$ of a collection of open balls covering $\mathcal{F}$ for finite $q > 0$ and $d < \beta$. In that subcover, we consider a normalized reward vector $ \mathbf{s} \in \mathcal{B}(\mathbf{s}^j, d), j \in \{1, \ldots, q\}$ and specify the indices 
\begin{itemize}
    \item $K := \argmax_{a \in \mathcal{A}} s_{a}$ (the set of indices corresponding to the maximizers of $\mathbf{s}$)
    \item $K^0 := \argmax_{a \in \mathcal{A}} s^0_a$ (the set of indices corresponding to  the maximizers of $\mathbf{s}^0$)
    \item $b \in \argmax_{a \in \mathcal{A}} |s^0_a - s_{a}|$ (an index corresponding to a highest absolute value entry in $\mathbf{s}^0 - \mathbf{s}$)
\end{itemize}
\begin{proposition} \label{prop:iden1} 
Suppose that $K^0 \cap K = \emptyset$ for a given vector $\mathbf{s} \in \mathcal{B}(\mathbf{s}^j, d), j \in \{1, \ldots, q\}$. Consider the incentives chosen uniformly randomly from the compact set $\mathcal{C}$, i.e., $\pi_{t,a} \sim U(\underline{C}, \overline{C}), \forall a \in \mathcal{A}$. Then, we bound the following probability conditioned on the case that the agent chooses the true maximizer arm at time $t \in \mathcal{T}$. 
 \begin{multline}
        \mathbb{P}\left(\ell\left(\mathbf{s}, \upsilon_{t}(\boldsymbol{\pi}_t), \boldsymbol{\pi}_t\right) \geq \omicron \Big | \upsilon_t(\boldsymbol{\pi}_t) = \argmax_{a \in \mathcal{A}}  \left(s^0_a + \pi_{t, a} \right)\right) \\ \geq \left(\frac{(s^0_{\kappa^0} - s^0_{\kappa})^2 -  \omicron^2}{2(\overline{C} - \underline{C})^2} \right)\left(1 - \frac{\gamma + \beta - d}{\overline{C} - \underline{C}} \right)^2  \left( \frac{\gamma}{\overline{C} - \underline{C}} \right)^{n-2}  \label{eq:identf-1}
\end{multline}
which holds for any two indices $\kappa^0 \in K^0$, $\kappa \in K$, and any constant $\omicron \in (0, \delta)$ where $\delta := \max_{a \in \mathcal{A}} s^0_{a} -  \max_{a \in \mathcal{A} \setminus \{K^0\}} s^0_{a}$ is the difference between the largest and second largest entries of $\mathbf{s}^0$. 
\end{proposition}
\begin{proposition} \label{prop:iden2}
Suppose that $K^0 \cap K \neq \emptyset$, $b \notin K^0 \cap K$ for a given vector $\mathbf{s} \in \mathcal{B}(\mathbf{s}^j, d), j \in \{1, \ldots, q\}$. Consider that  $\pi_{t,a} \sim U(\underline{C}, \overline{C}), \forall a \in \mathcal{A}$ and the agent chooses the true maximizer arm at time $t \in \mathcal{T}$. Then, 
 \begin{align}
        &\mathbb{P}\left(\ell\left(\mathbf{s}, \upsilon_{t}(\boldsymbol{\pi}_t), \boldsymbol{\pi}_t\right) \geq \omicron \Big | \upsilon_t(\boldsymbol{\pi}_t) = \argmax_{a \in \mathcal{A}} \left(s^0_a + \pi_{t, a} \right)\right) \geq \frac{(\beta - \omicron)^2}{(\overline{C}-\underline{C})^2} \left(1 - \frac{\gamma + \omega}{\overline{C} - \underline{C}} \right)^2 \left( \frac{\gamma}{\overline{C} - \underline{C}}\right)^{n-2} \label{eq:identf-2} \allowdisplaybreaks 
\end{align} 
for any constant $\omicron \in (0, \beta)$ where $\omega = \sup_{\mathbf{s} \in \mathcal{B}(\mathbf{s}^j, d)} \max_{a \in \mathcal{A}} (|s^0_a|, |s_a| )$ is the largest absolute value observed among the entries of $\mathbf{s}^0$ and of all the vectors in $\mathcal{B}(\mathbf{s}^j, d)$.
\end{proposition}
\begin{proposition} \label{prop:iden3}
Suppose that $K^0 \cap K \neq \emptyset$, $b \in K^0 \cap K$ for a given vector $\mathbf{s} \in \mathcal{B}(\mathbf{s}^j, d), j \in \{1, \ldots, q\}$. Consider that  $\pi_{t,a} \sim U(\underline{C}, \overline{C}), \forall a \in \mathcal{A}$ and the agent chooses the true maximizer arm at time $t \in \mathcal{T}$. Then, 
\begin{multline}
        \mathbb{P}\left(\ell\left(\mathbf{s}, \upsilon_{t}(\boldsymbol{\pi}_t), \boldsymbol{\pi}_t\right) \geq \omicron \Big | \upsilon_t(\boldsymbol{\pi}_t) = \argmax_{a \in \mathcal{A}} \left(s^0_a + \pi_{t, a} \right)\right) \\ \geq
        \frac{(\beta-\omicron)^2}{(\overline{C}-\underline{C})^2} \left(1 - \frac{\gamma + \beta - d}{\overline{C} - \underline{C}} \right) \left(1 - \frac{\gamma + \omega}{\overline{C} - \underline{C}} \right) \left( \frac{\gamma}{\overline{C} - \underline{C}}  \right)^{n-2} \label{eq:identf-3} \allowdisplaybreaks 
\end{multline}
for any constant $\omicron \in (0, \beta)$.
\end{proposition}

Our goal in these propositions is to show that the principal's estimator (\ref{prblm:estimator}) is able to differentiate between the true mean reward vector $\mathbf{s}^0$ and a different reward vector $\mathbf{s}$ that is at least $\beta$ away from $\mathbf{s}^0$ in terms of the $\ell_\infty$ norm. Due to the structure of our model, we prove this result separately for three mutually exclusive cases defined based on the sets of indices $K$ and $K^0$ and the index $b$ introduced above. However, each of these results are proven for the event that the agent's algorithm picks the true maximizer arm in response to the given random incentives. As stated in the introduction, we aim to offer a generic approach in this paper without limiting the type of the algorithm used by the agent. To achieve this, we consider a mild assumption on the learning behavior of the agent. In particular, we assume that after a transient period of learning, the agent's algorithm will choose an incorrect (i.e., different than the true maximizer) arm at a decreasing rate as the game move forwards over the considered time horizon. We specify this rate in the following statement. 
\begin{assumption} \label{assm:agent}
    Let $p_t := \mathbb{P}\left(\upsilon_t(\boldsymbol{\pi}_t) \neq \argmax_{a \in \mathcal{A}} \left( s^0_a + \pi_{t, a} \right)\right)$ be the probability that the agent does not select the true utility-maximizer arm at time $t$. There exists a constant $k \geq 1$ such that $p_t \leq k \dfrac{\sqrt{\log 2t}}{\sqrt{t}}$ at any time step $t \in [\widetilde{k}, T]$ where $\widetilde{k} \geq 2$ is the minimum value satisfying $k \textstyle \sqrt{\log 2\widetilde{k}} < \sqrt{\widetilde{k}}$. 
\end{assumption}

We later provide a validation of this assumption by showing that it is satisfied when the agent uses a classical MAB algorithm and the principal uses the proposed data-driven incentive policy presented in the next section. Now, we unite the results in Propositions \ref{prop:iden1} -- \ref{prop:iden3} and obtain our final identifiability statement. 
\begin{proposition} \label{prop:iden4}
At time $t \in [\widetilde{k}, T]$, suppose that $\pi_{t,a} \sim U(\underline{C}, \overline{C}), \forall a \in \mathcal{A}$. Then, for any normalized reward vector $ \mathbf{s} \in \mathcal{F}$, we have 
 \begin{align}
        &\mathbb{P}\left( \ell\left(\mathbf{s}, \upsilon_{t}(\boldsymbol{\pi}_t), \boldsymbol{\pi}_t\right) \geq \omicron \right) \geq \alpha (\beta - \omicron)^2 \left(1 - k \frac{ \sqrt{\log 2t}}{\sqrt{t}}\right) \label{eq:identf-4} \allowdisplaybreaks 
\end{align}
for any constant $\omicron \in (0, \beta)$ and $\alpha > 0$. 
\end{proposition}

This result proves that our estimator (\ref{prblm:estimator}) can identify and refute an incorrect estimate $\mathbf{s} \in \mathcal{F}$ with a strictly positive probability whose rate is proportional to the square of the estimation error $\beta$. Therefore, as the game proceeds and the agent learns their own reward model, the principal is also capable of effectively learning $\mathbf{s}^0$ from the arms chosen by the agent in response to the offered random incentives that explore the agent's bandit model.
\subsection{Finite-Sample Concentration Bound}\label{subsec:consistency}
We derive our statistical consistency result for the principal's estimator (\ref{prblm:estimator}) in several intermediate steps. In these steps, we prove finite-sample concentration inequalities with respect to the behavior of the loss function (\ref{eq:lossfunc}) evaluated at any incorrect reward vector $\mathbf{s} \in \mathcal{F}$ and the loss function evaluated at the true reward vector $\mathbf{s}^0$. 
\begin{proposition} \label{prop:concen1}
Let $\eta(\widetilde{k}, t)$ be the number of time steps that the principal chooses each incentive $\pi_{t,a}$ uniformly randomly from the compact set $\mathcal{C}$ within the time interval $[\widetilde{k}, t-1]$, i.e.,
\begin{equation}
    \eta(\widetilde{k}, t) = \left| \Lambda(\widetilde{k}, t) \right| \ \text{ where } \ \Lambda(\widetilde{k}, t) := \{\tau: \widetilde{k} \leq \tau \leq t-1, \ \pi_{\tau,a} \sim U(\underline{C}, \overline{C}), \forall a \in \mathcal{A}\}  \label{eq:eta} \allowdisplaybreaks
\end{equation}
For the given sequences of incentives $\boldsymbol{\Pi}_t$ and chosen arms $\Upsilon_t(\boldsymbol{\Pi}_t)$, the total estimation loss over these time steps is defined as
\begin{align}
     L^{\Lambda(\widetilde{k}, t)}\left(\mathbf{s}, \Upsilon_t(\boldsymbol{\Pi}_t), \boldsymbol{\Pi}_t\right) &=  \sum_{\tau\in \Lambda(\widetilde{k}, t)} \ell\left(\mathbf{s}, \upsilon_\tau(\boldsymbol{\pi}_\tau), \boldsymbol{\pi}_\tau \right)   \label{eq:l-stoc-lambda} \allowdisplaybreaks
\end{align}
and satisfies
\begin{multline}
    \mathbb{P} \left(\left| L^{\Lambda(\widetilde{k}, t)}\left(\mathbf{s}, \Upsilon_t(\boldsymbol{\Pi}_t), \boldsymbol{\Pi}_t\right) - \mathbb{E} L^{\Lambda(\widetilde{k}, t)}\left(\mathbf{s}, \Upsilon_t(\boldsymbol{\Pi}_t), \boldsymbol{\Pi}_t\right)\right| \geq \nu  \right) \\ \leq 2 \exp \left(- \frac{2\nu^2}{(\eta(\widetilde{k}, t)-1) n \left(6R_{\max} - 6R_{\min} + 2\gamma\right)^2 } \right)  \allowdisplaybreaks
\end{multline}
for any constant $\nu > 0$ and mean reward vector $\mathbf{s} \in \mathcal{S}$. 
\end{proposition}

The proof of this result follows by using the bounded differences inequality (i.e., McDiarmid’s inequality) \citep{boucheron2013concentration} and the definition of our single-step loss function $\ell\left(\mathbf{s}, \upsilon_\tau(\boldsymbol{\pi}_\tau), \boldsymbol{\pi}_\tau \right)$. 
\begin{proposition} \label{prop:concen2}
   For the given sequences of incentives $\boldsymbol{\Pi}_t$ and chosen arms $\Upsilon_t(\boldsymbol{\Pi}_t)$, the concentration of the loss function $L^{\Lambda(\widetilde{k}, t)}\left(\mathbf{s}, \Upsilon_t(\boldsymbol{\Pi}_t), \boldsymbol{\Pi}_t\right)$ in (\ref{eq:l-stoc-lambda}) within the the compact set $\mathcal{F} = \{\mathbf{s} \in \mathcal{S}^n: \| \mathbf{s} - \mathbf{s}^0 \|_\infty > \beta\}$ is given as 
   \begin{multline}
    \mathbb{P} \left(\sup_{\mathbf{s} \in \mathcal{F}} \left| L^{\Lambda(\widetilde{k}, t)}\left(\mathbf{s}, \Upsilon_t(\boldsymbol{\Pi}_t), \boldsymbol{\Pi}_t\right) - \mathbb{E} L^{\Lambda(\widetilde{k}, t)}\left(\mathbf{s}, \Upsilon_t(\boldsymbol{\Pi}_t), \boldsymbol{\Pi}_t\right)\right| \geq \nu  \right) \allowdisplaybreaks  \\ 
    \leq 2\exp \left(-\frac{2 \nu^2}{(\eta(\widetilde{k}, t)-1)n(6R_{\max} - 6R_{\min} + 2\gamma)^2} - \log \beta + n \log (R_{\max} - R_{\min})\right) \allowdisplaybreaks
\end{multline}
for any constant $\nu > 0$.
\end{proposition}

This proposition is proven by using the result of Proposition \ref{prop:concen1} and bounding the covering number $q$ for $\mathcal{F}$ by volume ratios. We now compute a lower bound for the minimum possible expected loss over $\Lambda(\widetilde{k}, t)$ achieved within $\mathcal{F}$, the set of feasible reward vectors that are at least $\beta$ away from $\mathbf{s}^0$.
\begin{lemma} \label{lem:concen1}
We define the minimizer of the loss (\ref{eq:l-stoc-lambda}) within the compact set $\mathcal{F}$ as $\mathbf{s}^{\mathcal{F}}_t := \arginf_{\mathbf{s} \in \mathcal{F}} L^{\Lambda(\widetilde{k}, t)}\left(\mathbf{s}, \Upsilon_t(\boldsymbol{\Pi}_t), \boldsymbol{\Pi}_t\right)$. Then, given the sequences of incentives $\boldsymbol{\Pi}_t$ and chosen arms $\Upsilon_t(\boldsymbol{\Pi}_t)$, we have 
\begin{align}
    \mathbb{E} L^{\Lambda(\widetilde{k}, t)}\left(\mathbf{s}^{\mathcal{F}}_t, \Upsilon_t(\boldsymbol{\Pi}_t), \boldsymbol{\Pi}_t\right) &\geq  \frac{4\alpha \left(1 - k \sqrt{\log 2\widetilde{k}}/ \sqrt{\widetilde{k}}\right)^2}{27} \beta^3 \mathbb{E}\eta(\widetilde{k}, t)   
\end{align}
for any $t \in [\widetilde{k}, T]$.
\end{lemma}

We prove this result by using Assumption \ref{assm:agent} and our identifiability result in Proposition \ref{prop:iden4}. We continue by deriving an upper bound for the expected total loss up to time $t$ evaluated at the true mean reward vector $\mathbf{s}^0$.
\begin{lemma} \label{lem:concen2}
The expectation of the total loss of the principal's estimator (\ref{prblm:estimator}) computed for the true mean reward vector $\mathbf{s}^0$ is bounded by 
\begin{align}
    \mathbb{E} L\left(\mathbf{s}^0, \Upsilon_t(\boldsymbol{\Pi}_t), \boldsymbol{\Pi}_t\right) &\leq 3 k \left(3 (R_{\max} - R_{\min}) + \gamma\right) \sqrt{t\log (2t)} 
\end{align}    
\end{lemma}
In the last lemma of this section, we present the concentration inequality for $\mathbf{s}^0$. 
\begin{lemma}\label{lem:concen3}
For the given sequences of incentives $\boldsymbol{\Pi}_t$ and chosen arms $\Upsilon_t(\boldsymbol{\Pi}_t)$, the concentration of the total loss of the principal's estimator evaluated at $\mathbf{s}^0$ is given as
\begin{align}
  \mathbb{P}\left( L\left(\mathbf{s}^0, \Upsilon_t(\boldsymbol{\Pi}_t), \boldsymbol{\Pi}_t\right) - \mathbb{E}  L\left(\mathbf{s}^0, \Upsilon_t(\boldsymbol{\Pi}_t), \boldsymbol{\Pi}_t\right) \geq \nu \right) &\leq \exp\bigg(-\frac{2\nu^2}{(t-1)\left(3R_{\max} - 3R_{\min} + \gamma\right)^2}\bigg) 
\end{align}
for any $\nu > 0$.
\end{lemma}

This lemma is proven by first observing that the loss for $\mathbf{s}^0$ at any time step becomes 0 when the agent selects the true maximizer arm, and thus, it suffices to only consider the time steps where the agent's algorithm makes an inaccurate decision. Then, the proof follows by using Hoeffding's Inequality \citep{boucheron2013concentration}. 

The last step is to combine Proposition \ref{prop:concen2} and Lemmas \ref{lem:concen1}-\ref{lem:concen3} to obtain the final finite-sample concentration bound for the principal's estimator with respect to the total loss function (\ref{eq:lossfunc}). 
\begin{theorem} \label{thm:concen}
We introduce the quantity
\begin{align}
    \lambda_t = \frac{4\alpha \Big(1 - k \sqrt{\log 2\widetilde{k}}/ \sqrt{\widetilde{k}}\Big)^2}{27} \beta^3 \mathbb{E}\eta(\widetilde{k}, t) - 3 k \left(3 (R_{\max} - R_{\min}) + \gamma\right) \sqrt{t\log (2t)}. \label{eq:lambda_t}
\end{align}
Given the sequences of incentives $\boldsymbol{\Pi}_t$ and agent's choices $\Upsilon_t(\boldsymbol{\Pi}_t)$, we show that
\begin{multline}
     \mathbb{P}\left(\inf_{\mathbf{s} \in \mathcal{F}} L\left(\mathbf{s}, \Upsilon_t(\boldsymbol{\Pi}_t), \boldsymbol{\Pi}_t\right) \leq L\left(\mathbf{s}^0, \Upsilon_t(\boldsymbol{\Pi}_t), \boldsymbol{\Pi}_t\right) \right) \allowdisplaybreaks  \\ \leq 2\exp \left(-\frac{2\lambda_t^2}{(t-1)16n(6R_{\max} - 6R_{\min} + 2\gamma)^2} - \log \beta + n \log (2R_{\max} - 2R_{\min})\right)
 \end{multline}
for $t \in [\widetilde{k}, T]$.
\end{theorem}

Alternatively, we can reinterpret Theorem \ref{thm:concen} and obtain a concentration with respect to the $\ell_\infty$ distance between our estimates $\widehat{\mathbf{s}}_t$ and the true reward vector $\mathbf{s}^0$. 
\begin{corollary} {\mdseries \scshape (Finite-Sample Concentration Bound)} \label{cor:concen}
The principal's estimator (\ref{prblm:estimator}) satisfies 
\begin{align}
    \mathbb{P} \left(\|\mathbf{s}^0 - \widehat{\mathbf{s}}^{\mathrm{pr}}_t \|_\infty > \beta \right) \leq 2\exp \left(-\frac{2\lambda_t^2}{(t-1)16n(6R_{\max} - 6R_{\min} + 2\gamma)^2} - \log \beta + n \log (2R_{\max} - 2R_{\min})\right) 
\end{align} 
for any $\beta > 0$ and $t \in [\widetilde{k}, T]$ where $\lambda_t$ is as defined in (\ref{eq:lambda_t}).
\end{corollary}


It is important to note that in this bound the learning rate of the principal directly depends on $\eta(\widetilde{k}, t)$, which is the number of periods at which they offer random incentives from an offset time $\widetilde{k}$ and beyond to explore the arm space of the agent. More importantly, the existence of this offset point shows that the rate that the principal explores the agent's bandit model must be greater than the agent's exploration rate. Recall that we consider a selfish agent who is only interested in learning their own reward model whereas the principal needs to learn both their own rewards and the agent's rewards to be able to design effective incentives. Because the principal's learning process is fed by the agent's decisions over the considered repeated game, the principal's learning will be possible if and only if the agent's learning is successful. To restate, the principal has to wait for a sufficient time after which the agent will start playing the right arm the most fraction of the time. That is why our finite-sample concentration bound holds for the time periods $t \geq \widetilde{k}$ where $\widetilde{k}$ is an increasing function of the agent's parameter $k$ as given in Assumption \ref{assm:agent}. (We note that the offset point of time $\widetilde{k}$ can be computed numerically when $k$ is known.) The implication of this result is that for higher $k$ values, it will take a longer time for the agent to start playing correctly in a consistent way and for the principal's estimator to converge to $\mathbf{s}^0$. We conclude this section by emphasizing this discussion in the following remark. 
\begin{remark} \label{rem:learningrate}
   The rationale behind the condition $t \geq \widetilde{k}$ in Corollary \ref{cor:concen} relies on an essential dynamic of the repeated adverse selection games we study in this paper. It reflects the fact that the finite-sample consistency of the principal's estimator is attainable only after a transient learning period for the agent's algorithm. 
\end{remark}

This fundamental observation once again brings us back to the main theoretical challenge we pointed out in the introduction of the paper. The adverse impact of the agent's learning process on the principal's inferences immensely accumulates the costs of the principal, as we will show in our regret analysis in the next section. 
\section{The Data-Driven Sequential Incentive Policy}\label{sec:policy}
Our novel and consistent estimator allows the principal to design an adaptive and easy to compute menu of incentives for the agent's bandit arms. To achieve this, we propose a MAB framework for the principal within which we unify the principal's estimator and the data-driven incentives we propose in this section. 

Before we present the details of the proposed learning framework, we recall that the principal's problem involves estimating their own reward expectations $\boldsymbol{\theta}^0$ in addition to the agent's rewards. However, the former estimation is a more manageable problem than the latter because the principal can fully observe their own reward outcomes $\mu_{t, \upsilon_t(\boldsymbol{\pi}_t)}$ realized through the arms chosen by the agent. 
\begin{assumption}\label{assm:subgaussian}
For each agent arm $a \in \mathcal{A}$, the principal observes independent reward realizations $\mu_{t,a}, t \in \mathcal{T}$ from a sub-Gaussian distribution $\mathbb{F}^{\mathrm{pr}}_{\theta^0_a, a}$ for all $\theta^0_a \in \Theta$. Similarly, the agent's rewards $\rho_{t, a}, t \in \mathcal{T}$ are independent from each other and follow a sub-Gaussian distribution for all $r^0_a \in \mathcal{R}$. 
\end{assumption}
This assumption about the reward distribution families of the principal and the agent is a mild and common condition encountered in many MAB models. Under this assumption, we define the quantity $T(a,t) = \left|\{\tau: 1 \leq \tau \leq t-1, \upsilon_\tau(\boldsymbol{\pi}_\tau) = a\} \right|$ as the number of time points when the agent selects arm $a$ up to time $t$. Then, we consider the sample mean of the principal's reward outcomes up to time $t$ as the principal's estimate for $\theta^0_a, \forall a \in \mathcal{A}$.
\begin{align}
    \widehat{\theta}_{t,a} = \frac{1}{T(a,t)}\sum_{\tau = 1}^{t-1} \mu_{\tau,a} \mathbbm{1} \left(\upsilon_\tau(\boldsymbol{\pi}_\tau) = a\right) \label{eq:theta-hat}
\end{align}
We note that $\widehat{\theta}_{t,a}$ is an unbiased estimator and is the same as the maximum likelihood estimator for many common exponential family distributions where the sufficient statistic is equal to the random variable itself, including the Gaussian, Bernoulli, Poisson, and  multinomial distributions. 
\subsection{Principal's $\epsilon$-Greedy Policy}\label{subsec:algorithm}
We next present a sequential learning framework that utilizes the principal's estimators $\widehat{\mathbf{s}}^{\mathrm{pr}}_t$ and $\widehat{\boldsymbol{\theta}}_{t}$ to propose an adaptive and efficient incentive policy. Keeping in mind practicality, we develop an $\epsilon$-greedy algorithm for which we provide pseudocode in Algorithm \ref{alg:pr}. 

To initiate the principal's and agent's learning processes, we consider an initialization period over the first $n = |\mathcal{A}|$ time periods throughout which the agent is induced to pick each of the $n$ arms once to observe a reward realization and compute a starting estimate of the associated reward expectation. To achieve this, the principal offers the maximum possible incentive ($\overline{C}$) for the desired arm in each step that would be sufficient to make that arm utility-maximizer for the agent by Assumption \ref{assm1}. 

At each time step $t \in [n+1, \ldots, T]$ after the initialization period, the principal first updates their estimate for their own mean reward associated with the most recently chosen arm $\theta^0_{\upsilon_{t-1}(\boldsymbol{\pi}_{t-1})}$. Then, the principal samples a Bernoulli random variable $x^{\mathrm{pr}}_t$ with success probability $\epsilon^{\mathrm{pr}}_t$ which corresponds to the principal's exploration probability. In accordance with our observation highlighted in Remark \ref{rem:learningrate} following our statistical analysis in Section \ref{subsec:consistency}, here we clarify that the principal's rate of the exploration is designed to be greater than the agent's learning rate in accordance with Assumption \ref{assm:agent}. 

If $x^{\mathrm{pr}}_t = 1$, then the principal performs an exploration step where they offer incentives $\boldsymbol{\pi}_t = (\pi_{t,a})_{a \in \mathcal{A}}$ that are uniformly randomly selected from the feasible range of incentives $\mathcal{C} = [\underline{C}, \overline{C}]$. Otherwise, for $x^{\mathrm{pr}}_t = 0$ the principal prefers a less-risky, greedy exploitation play to maximize their expected net reward in that period. They update their estimate $\widehat{\mathbf{s}}^{\mathrm{pr}}_t$ for the agent's mean rewards by solving (\ref{prblm:estimator}) and use it to compute the incentives $\mathbf{c}(\widehat{\boldsymbol{\theta}}_t, \widehat{\mathbf{s}}^{\mathrm{pr}}_t)$ maximizing their estimated expected net reward at time $t$. Recall that the principal's expected net reward in a period is equal to their fixed expected reward for the chosen arm minus the sum of incentives offered for each bandit arm. Therefore, our goal in an exploitation period is to compute the minimum vector of incentives that will steer the agent's choice into the desired arm (that is estimated to yield the highest expected net reward to the principal) in that period while inducing the agent's incentive compatibility. Using the most recent estimates $\widehat{\boldsymbol{\theta}}_t$ and $\widehat{\mathbf{s}}^{\mathrm{pr}}_t$, for each arm $j \in \mathcal{A}$, we compute the minimum amount of incentives $(\widetilde{c}^j_a)_{a \in \mathcal{A}}$ that make $j$ the maximizer arm for the agent and the corresponding expected net reward  $\widetilde{V}(j, \widehat{\mathbf{s}}^{\mathrm{pr}}_t; \widehat{\boldsymbol{\theta}}_t)$ of the principal in case $j$ is actually chosen by the agent in response to these incentives. Further, to contend with the principal's estimation error in $\widehat{\mathbf{s}}^{\mathrm{pr}}_t$ and the unknown behavior of the learning agent, we add a buffer amount to the computed minimum incentives and obtain 
\begin{align}
    &\widetilde{c}^j_j = \left(\textstyle \max\limits_{a \in \mathcal{A}} \widehat{s}^{\mathrm{pr}}_{t,a}\right) - \widehat{s}^{\mathrm{pr}}_{t,j} + 2\beta_t \label{eq:incentive1}  \allowdisplaybreaks  \\
    &\widetilde{c}^j_a = 0, \quad \forall a \in \mathcal{A}, \ a \neq j  \label{eq:incentive2}  \allowdisplaybreaks  \\
    &\widetilde{V}(j, \widehat{\mathbf{s}}^{\mathrm{pr}}_t; \widehat{\boldsymbol{\theta}}_t) = \widehat{\theta}_{t, j} - \sum\limits_{a \in \mathcal{A}} \widetilde{c}^j_a =  \widehat{\theta}_{t, j} - \left(\max\limits_{a \in \mathcal{A}} \widehat{s}^{\mathrm{pr}}_{t,a}\right) + \widehat{s}^{\mathrm{pr}}_{t,j} - 2\beta_t \allowdisplaybreaks \label{eq:incentive3}
\end{align}
where $\beta_t = B \frac{\sqrt{\log 2t}}{t^{w/3}}$ with $B = \frac{3 k \left(3 (R_{\max} - R_{\min}) + \gamma\right)^n \sqrt[6]{32n}}{1 - k \sqrt{\log 2\widetilde{k}}/ \sqrt{\widetilde{k}}}$. After computing these values for each arm $j \in \mathcal{A}$, the principal chooses the set of incentives corresponding to the arm $j^*_t$ that is estimated to yield the highest $\widetilde{V}(j, \widehat{\mathbf{s}}^{\mathrm{pr}}_t; \widehat{\boldsymbol{\theta}}_t)$ to the principal. To reiterate, the exploitation incentives are purposefully designed to make the desired arm $j^*_t$ utility-maximizer for the agent with high probability by leveraging the statistically consistent estimator proposed in Section \ref{sec:estimator}. Lastly, we clarify that the principal only observes the arm $\upsilon_t(\boldsymbol{\pi}_t)$ chosen by the agent in response to the offered incentives and their own net reward realization $\mu_{t,\upsilon_t(\boldsymbol{\pi}_t)} - \sum_{a \in \mathcal{A}} \pi_{t, a}$ at the end of each period $t$. The total utility collected by the agent $\rho_{t, \upsilon_t(\boldsymbol{\pi}_t)} + \pi_{t, \upsilon_t(\boldsymbol{\pi}_t)}$ remains as private knowledge under the considered hidden rewards setting.
\begin{remark} Computation of the exploitation incentives through lines \ref{alg-steps1}--\ref{alg-stepslast} in Algorithm \ref{alg:pr} includes arithmetic operations with a linear computational complexity $O(n)$ in terms of the dimension of the agent's bandit model $n = |\mathcal{A}|$. 
\end{remark}
\begin{algorithm*}
\caption{Principal's $\epsilon$-Greedy Algorithm}
\label{alg:pr} 
\begin{algorithmic}[1]
\State Set: $m^{\mathrm{pr}} \geq 1$ and $w \in (0, 1/4)$
\For{$t \in [1, \ldots, n]$} \label{alg-initial1}
\State Set: $\boldsymbol{\pi}_t = (\pi_{t,a})_{a \in \mathcal{A}}$ where $\pi_{t, a} = \overline{C}$ for $a = t$ and $\pi_{t, a} = 0$ for all $a \neq t$
\State Observe: $\upsilon_t(\boldsymbol{\pi}_t) = t$ and $\mu_{t, \upsilon_t(\boldsymbol{\pi}_t)}$
\If {$t \geq 2$} $\widehat{\theta}_{t, \upsilon_{t-1}(\boldsymbol{\pi}_{t-1})} = \mu_{t-1, \upsilon_{t-1}(\boldsymbol{\pi}_{t-1})}$ \label{alg-initiallast}
\EndIf
\EndFor
\For{$t \in [n+1, \ldots, T]$} 
\State Compute: $\widehat{\theta}_{t, \upsilon_{t-1}(\boldsymbol{\pi}_{t-1})} \in \dfrac{1}{T(\upsilon_{t-1}(\boldsymbol{\pi}_{t-1}), t)} \sum\limits_{\tau = 1}^{t-1} \mu_{\tau,\upsilon_{t-1}(\boldsymbol{\pi}_{t-1})} \mathbbm{1} (\upsilon_\tau(\boldsymbol{\pi}_\tau) = \upsilon_{t-1}(\boldsymbol{\pi}_{t-1})) $ 
\State Set: $\epsilon^{\mathrm{pr}}_t = \min \left\{1, \dfrac{m^{\mathrm{pr}}}{t^{1/2 - w}}\right\}$ 
\State Sample: $x^{\mathrm{pr}}_t \sim \mathrm{Bernoulli} (\epsilon^{\mathrm{pr}}_t)$
\If {$x^{\mathrm{pr}}_t = 1$}
\State Sample: $\pi_{t,a} \sim \mathcal{U}\left(\underline{C}, \overline{C}\right)$ for all $a \in \mathcal{A}$ \label{alg-explore1}
\State Set: $\boldsymbol{\pi}_t = (\pi_{t,a})_{a \in \mathcal{A}}$ \label{alg-explorelast}
\Else {} 
\State Compute: $\beta_t = B \dfrac{\sqrt{\log 2t}}{t^{w/3}}$ \label{alg-exploit1}
\State Compute: $\widehat{\mathbf{s}}^{\mathrm{pr}}_t \in \argmin \left\{L\left(\mathbf{s}, \Upsilon_t(\boldsymbol{\Pi}_t), \boldsymbol{\Pi}_t\right) \big | s_1 = 0, \ s_a \in \mathcal{S} \ \forall a \in \mathcal{A}\right\}$ 
\For{$j \in \mathcal{A}$}  \label{alg-steps1}
\State Compute: $\widetilde{V}(j, \widehat{\mathbf{s}}^{\mathrm{pr}}_t; \widehat{\boldsymbol{\theta}}_t) = \widehat{\theta}_{t,j} - \left(\max\limits_{a \in \mathcal{A}} \widehat{s}^{\mathrm{pr}}_{t,a}\right) + \widehat{s}^{\mathrm{pr}}_{t,j} - 2\beta_t$
\EndFor
\State Compute: $j^*_t = \argmax\limits_{j \in \mathcal{A}} \widetilde{V}(j, \widehat{\mathbf{s}}^{\mathrm{pr}}_t; \widehat{\boldsymbol{\theta}}_t)$ \label{eq:jstar}
\State Set: $c_{j^*_t}(\widehat{\boldsymbol{\theta}}_t, \widehat{\mathbf{s}}^{\mathrm{pr}}_t) = \left(\max\limits_{a \in \mathcal{A}} \widehat{s}^{\mathrm{pr}}_{t,a}\right) - \widehat{s}^{\mathrm{pr}}_{t,j^*_t} + 2\beta_t$ and $c_{a}(\widehat{\boldsymbol{\theta}}_t, \widehat{\mathbf{s}}^{\mathrm{pr}}_t) = 0 $ for all $a \neq j^*_t$ \label{alg-stepslast}
\State Set: $\boldsymbol{\pi}_t = (c_{a}(\widehat{\boldsymbol{\theta}}_t, \widehat{\mathbf{s}}^{\mathrm{pr}}_t))_{a \in \mathcal{A}}$ \label{alg-exploitlast}
\EndIf
\State Observe: $\upsilon_t(\boldsymbol{\pi}_t)$ and $\mu_{t, \upsilon_t(\boldsymbol{\pi}_t)}$ \label{alg-action}
\EndFor
\end{algorithmic}
\end{algorithm*}
\subsection{Principal's Finite-Sample Regret}\label{subsec:regret}
To show the efficiency and effectiveness of the designed incentive mechanism, we conduct a rigorous regret analysis where regret is defined in terms of the expected net reward of the principal accumulated over a finite time horizon. 
\subsubsection{Oracle Incentive Policy}\label{subsubsec:oracle}
As a benchmark to the proposed incentive policy, we define an oracle incentive policy that maximizes the principal's expected net reward under full knowledge of all reward expectations $\mathbf{s}^0$ and $\boldsymbol{\theta}^0$. In other words, the oracle policy computes the exploitation incentives by using the true mean reward values and assuming a perfect-knowledge agent. Similar to the procedure described in the previous subsection, for each arm $j\in \mathcal{A}$, we solve for the minimum incentives $(\widetilde{c}^j_a)_{a \in \mathcal{A}}$ required to make $j \in \mathcal{A}$ maximizer of the agent's total utility and compute the corresponding expected net reward of the principal $\widetilde{V}(j, \mathbf{s}^0;\boldsymbol{\theta}^0)$ as follows. 
\begin{align}
    &\widetilde{c}^j_j = \left(\max_{a \in \mathcal{A}} s^0_a \right) - s^0_j \allowdisplaybreaks  \\
    &\widetilde{c}^j_a = 0, \quad \forall a \in \mathcal{A}, \ a \neq j \allowdisplaybreaks  \\
    &\widetilde{V}(j, \mathbf{s}^0;\boldsymbol{\theta}^0) = \theta^0_j - \sum_{a \in \mathcal{A}} \widetilde{c}^j_a = \theta^0_j - \left(\max_{a \in \mathcal{A}} s^0_a \right) + s^0_j
\end{align}
Then, the oracle policy chooses the set of incentives corresponding to desired arm $j^{*,0} := \argmax_{j \in \mathcal{A}} \widetilde{V}(j, \mathbf{s}^0;\boldsymbol{\theta}^0)$. The computed oracle incentives are denoted by $\mathbf{c}(\boldsymbol{\theta}^0, \mathbf{s}^0)$ and given as
\begin{align}
    &c_{j^{*,0}}(\boldsymbol{\theta}^0, \mathbf{s}^0) = \left(\max_{a \in \mathcal{A}} s^0_a\right)  - s^0_{j^{*,0}} + \varsigma \allowdisplaybreaks  \label{eq:oracleincentive1} \\
    &c_a(\boldsymbol{\theta}^0, \mathbf{s}^0) =  0, \quad \forall a \neq j^{*,0} \label{eq:oracleincentive2}\allowdisplaybreaks
\end{align}
where $\varsigma > 0$ is an arbitrarily small constant that helps avoiding the occurrence of multiple maximizer arms for the agent. Because the oracle policy fully knows the agent's expected rewards and hence the true utility-maximizer arm, the main difference between oracle incentives and the exploitation incentives computed by Algorithm \ref{alg:pr} is the absence of the buffer amount $\beta_t$. Further, we observe that the desired arm $j^{*,0}$ of the oracle policy is equal to the agent's true maximizer arm in response to the oracle incentives, i.e., $j^{*,0} = \upsilon(\mathbf{c}(\boldsymbol{\theta}^0, \mathbf{s}^0)) = \argmax_{a \in \mathcal{A}} s^0_a + c_a(\boldsymbol{\theta}^0, \mathbf{s}^0)$.

\subsubsection{Regret Bound} \label{subsubsec:regret}
At any time period $t \in \mathcal{T}$, the expected net reward of the principal under the oracle incentive policy is given as 
\begin{align}
    V(\mathbf{c}(\boldsymbol{\theta}^0, \mathbf{s}^0); \boldsymbol{\theta}^0) &= \theta^0_{j^{*,0}} - \max_{a \in \mathcal{A}} s^0_a  + s^0_{j^{*,0}} - \varsigma \label{eq:stoc-oracle} \allowdisplaybreaks 
\end{align}
and under the proposed incentive policy generated by Algorithm \ref{alg:pr} is given as 
\begin{align}
    V_t(\boldsymbol{\pi}_t; \boldsymbol{\theta}^0) = \theta^0_{\upsilon_t(\boldsymbol{\pi}_t)} - \sum_{a \in \mathcal{A}} \pi_{t,a}.\label{eq:stoc-epsilon}
\end{align}
Accordingly, we define the regret of the proposed $\epsilon$-greedy incentive policy $\Pi_{\epsilon, T} = \{\boldsymbol{\pi}_t\}_{t\in \mathcal{T}}$ with respect to the oracle incentive policy over a finite time horizon $\mathcal{T}$ as  
\begin{align}
    \mathrm{Regret}\left(\Pi_{\epsilon, T} \right) &= \sum_{t \in \mathcal{T}} V(\mathbf{c}(\boldsymbol{\theta}^0, \mathbf{s}^0); \boldsymbol{\theta}^0) - V_t(\boldsymbol{\pi}_t; \boldsymbol{\theta}^0).  \label{eq:regretdefn}
\end{align}

Before we present our upper bound for this regret notion, we first prove a useful theoretical result showing a probability bound on the accuracy of the arm $j^*_t$ that is estimated by Algorithm \ref{alg:pr} to yield the maximum expected net reward to the principal at any time period.
\begin{proposition} \label{prop:regret}
    The probability that the estimated best arm ($j^*_t$) for the principal in an exploitation step of Algorithm \ref{alg:pr} (see lines \ref{alg-steps1}--\ref{alg-stepslast}) is the same as the true best arm ($j^{*,0}$) chosen by the oracle policy is bounded from above by
    \begin{align}
        \mathbb{P} \left(j^*_t \neq j^{*,0}\right) \leq \frac{n}{t} + \frac{n^{5/6} 2^{n+1}}{3^{n + 1} k \sqrt[6]{32}} \frac{1}{\sqrt{t \log 2t}}\allowdisplaybreaks 
    \end{align}
for $t \in [\widetilde{k}, T]$ where $\widetilde{k}$ is as introduced in Assumption \ref{assm:agent}.
\end{proposition}

This probability bound on the accuracy of our estimate for the principal's desired arm at each period has two main components corresponding to the estimation gap of $\widehat{\mathbf{s}}^{\mathrm{pr}}_t$ and the error in $\widehat{\boldsymbol{\theta}}_t$. Therefore, the proof follows by mainly using our finite-sample concentration bound for $\widehat{\mathbf{s}}^{\mathrm{pr}}_t$ given in Corollary \ref{cor:concen} and Hoeffding's Inequality \citep{boucheron2013concentration} for consistency of $\widehat{\boldsymbol{\theta}}_t$. 

Lastly, we combine this result with the results of our statistical analysis and prove a rigorous regret bound for the proposed learning framework of the principal. 
\begin{theorem} {\mdseries \scshape (Finite-Sample Regret Bound)} \label{thm:regret}
The finite-sample regret bound of a policy $\Pi_{\epsilon, T}$ computed by the principal's $\epsilon$-greedy algorithm (\ref{alg:pr}) is proven to be 
    \begin{align}
    \mathrm{Regret}\left(\Pi_{\epsilon, T} \right) &\leq \frac{12 B}{3 - w}T^{1 - w/3} \sqrt{\log 2T}  + m^\mathrm{pr} \left(n (\overline{C} - \underline{C}) + \Theta^{\max}\right) \left(\frac{2}{2w + 1} T^{w + 1/2} + \frac{2w - 1}{2w + 1}\right) \notag \allowdisplaybreaks \\
    & + 2k \Theta^{\mathrm{\max}} \sqrt{T \log 2T} + \frac{2^{n+1} \left(\Theta^{\mathrm{\max}} (2n^{11/6} + 1/\sqrt[6]{n}) + n^{5/6} (\overline{C} - \underline{C})\left(1 + 2n \right) \right)}{3^{n + 1} k \sqrt[6]{32}}  \sqrt{T} \notag \allowdisplaybreaks \\
    &+  n^2 \left(\overline{C} - \underline{C} + \Theta^{\mathrm{\max}}\right) \log T + \Theta^{\mathrm{\max}} \widetilde{k}  \allowdisplaybreaks 
    \end{align}
where $\Theta^{\max}$ is defined as the upper bound on $\theta^0_a$'s.
\end{theorem}
\begin{remark} \label{rem:regret}
    The finite-sample regret bound proved for the proposed data-driven and adaptive incentive policy implies an asymptotic regret bound of order $O\left(T^{11/12 + \sigma} \sqrt{\log T}\right)$ where $\sigma$ can be made arbitrarily close to 0.  
\end{remark}

We emphasize that the principal's regret also comprises the agent's regret through the uncertainty in the arms chosen over the course of the repeated \emph{hidden rewards} game. Our regret bound reflects the substantial complexity and adverse impact resulting from the two parallel learning algorithms that are dynamically interacting with each other throughout the considered time horizon. For this challenging setting, our generic and practically relevant incentive mechanism is able to achieve a sublinear regret performance without restricting the type of the learning algorithm used by the incentivized agent. 
\section{The Learning Agent's Behavior} \label{sec:agent}

Although our goal in this paper is to mainly address the estimation and incentive design problems faced by the principal in the considered repeated hidden rewards game, this section provides a theoretical analysis of the agent's learning behavior and a discussion on the information rent gained by the self-interested agent. 

\subsection{Validation of Assumption \ref{assm:agent}}\label{subsec:validation}

As discussed earlier and highlighted in Remark \ref{rem:learningrate}, convergence of the principal's learning policy is only attainable after convergence of the agent's learning policy. Therefore, our theoretical analysis for statistical consistency of the proposed estimator in Section \ref{sec:estimator} and regret bound of the proposed incentive policy in Section \ref{sec:policy} needed to assume a probability bound on the learning behavior of the agent. We specified this bound in Assumption \ref{assm:agent}, and here we show the mildness of this assumption by proving that it is satisfied by a naive $\epsilon$-greedy algorithm used by the agent to make their decisions throughout the repeated game.

We consider the $\epsilon$-greedy algorithm presented in Algorithm \ref{alg:ag}. The first $n = |\mathcal{A}|$ time steps constitute the initialization period where the agent selects each of their $n$ arms to obtain a random reward realization and have an initial estimate for their associated mean reward. Because the agent can fully observe their own rewards, we consider the sample mean of their random reward outcomes as their estimate $\widehat{\boldsymbol{s}}^{\mathrm{ag}}_{t}$ for the mean rewards $\mathbf{s}^0$. We note that this corresponds to an unbiased estimator under Assumption \ref{assm:subgaussian}.

After that period, at each time step $t \in [n+1, \ldots, T]$, the agent first observes the menu of incentives $\boldsymbol{\pi}_t$ offered by the principal and updates their estimate for their own mean reward associated with the most recently chosen arm $\widehat{s}^{\mathrm{ag}}_{t, \upsilon_{t-1}(\boldsymbol{\pi}_{t-1})}$. Then, the algorithm samples a Bernoulli random variable $x^{\mathrm{ag}}_t$ based on the agent's exploration probability $\epsilon^{\mathrm{ag}}_t$. If $x^{\mathrm{ag}}_t = 1$, the agent explores their MAB model by randomly selecting an arm for the current time step. Otherwise, the agent performs a greedy exploitation by picking the arm that maximizes their estimated expected reward plus the offered incentive. At the end of each play, the self-interested agent only observes the stochastic reward they collect from the chosen arm.   
\begin{algorithm*}
\caption{Agent's $\epsilon$-Greedy Algorithm}
\label{alg:ag} 
\begin{algorithmic}[1]
\State Set: $m^{\mathrm{ag}} \geq 1$
\For{$t \in [1, \ldots, n]$} \label{agent-alg-initial1}
\State Observe: $\boldsymbol{\pi}_t$
\State Set: $\upsilon_t(\boldsymbol{\pi}_t) = t$ 
\State Observe: $\rho_{t, \upsilon_t(\boldsymbol{\pi}_t)}$
\If {$t \geq 2$} $\widehat{s}^{\mathrm{ag}}_{t, \upsilon_{t-1}(\boldsymbol{\pi}_{t-1})} = \rho_{t-1, \upsilon_{t-1}(\boldsymbol{\pi}_{t-1})}$ \label{agent-alg-initiallast}
\EndIf
\EndFor
\For{$t \in [n+1, \ldots, T]$} 
\State Observe: $\boldsymbol{\pi}_t$
\State Compute: $\widehat{s}^{\mathrm{ag}}_{t, \upsilon_{t-1}(\boldsymbol{\pi}_{t-1})} \in \dfrac{1}{T(\upsilon_{t-1}(\boldsymbol{\pi}_{t-1}), t)} \sum\limits_{\tau = 1}^{t-1} \rho_{\tau,\upsilon_{t-1}(\boldsymbol{\pi}_{t-1})} \mathbbm{1} (\upsilon_\tau(\boldsymbol{\pi}_\tau) = \upsilon_{t-1}(\boldsymbol{\pi}_{t-1})) $  \label{def:s-hat-agent}
\State Set: $\epsilon^{\mathrm{ag}}_t = \min \left\{1, \dfrac{m^{\mathrm{ag}}}{\sqrt{t}}\right\}$ 
\State Sample: $x^{\mathrm{ag}}_t \sim \mathrm{Bernoulli} (\epsilon^{\mathrm{ag}}_t)$
\If {$x^{\mathrm{ag}}_t = 1$} Sample: $\upsilon_t(\boldsymbol{\pi}_t) \in \mathcal{A}$ 
\Else \  Set: $\upsilon_t(\boldsymbol{\pi}_t) = \argmax_{a \in \mathcal{A}} \widehat{s}^{\mathrm{ag}}_{t, a} + \pi_{t, a}$ 
\EndIf
\State Observe: $\rho_{t, \upsilon_t(\boldsymbol{\pi}_t)}$ 
\EndFor
\end{algorithmic}
\end{algorithm*}

We show that Algorithm \ref{alg:ag} is consistent with Assumption \ref{assm:agent} by first proving two useful lemmas and then combining them in Proposition \ref{prop:validate}. For notational convenience, we let $\mathcal{T}^{\mathrm{pr-xplore}} \in \mathcal{T}$ and $\mathcal{T}^{\mathrm{pr-xploit}} \in \mathcal{T}$ be the set of random time steps that the principal's algorithm (\ref{alg:pr}) performs exploration (lines \ref{alg-explore1}--\ref{alg-explorelast}) and exploitation (lines \ref{alg-exploit1}--\ref{alg-exploitlast}), respectively. Similarly, we define $\mathcal{T}^{\mathrm{ag-xplore}}, \mathcal{T}^{\mathrm{ag-xploit}} \in \mathcal{T}$ as the set of random time steps that the agent's algorithm (\ref{alg:ag}) performs exploration and exploitation, respectively. 

\begin{lemma} \label{lem:validate1}
    Consider a time step $t \in [\widetilde{k}, T]$ where the principal offers the  exploitation incentives computed according to Algorithm \ref{alg:pr} and the agent performs a greedy exploitation by picking the utility-maximizer arm according to Algorithm \ref{alg:ag}. Then, the probability that the agent does not select the true utility-maximizer arm at time $t$ is bounded by
\begin{align}
     \mathbb{P}\Big(\upsilon_t(\boldsymbol{\pi}_t) \neq &\argmax_{a \in \mathcal{A}} 
 s^0_a + \pi_{t, a} \Big | t \in \mathcal{T}^{\mathrm{ag-xploit}} \cap  \mathcal{T}^{\mathrm{pr-xploit}} \Big) \leq \frac{2n^3 \left(\exp\Big(\frac{2(m^{\mathrm{ag}})^2B^2}{4n(R_{\max} - R_{\min})^2}\Big) + 1\right)}{t-1}  \notag \allowdisplaybreaks \\
     & + 8n^2\exp \left(-\frac{2\lambda_t^2}{(t-1)16n(6R_{\max} - 6R_{\min} + 2\gamma)^2} - \log \frac{\beta_t}{2} + n \log(2R_{\max} - 2R_{\min})\right)   \allowdisplaybreaks
\end{align}
where $\lambda_t$ is as defined in (\ref{eq:lambda_t}) and $\beta_t$ is as specified in (\ref{alg-exploit1}).
\end{lemma}
\begin{lemma} \label{lem:validate2}
    Consider a time step $t \in \mathcal{T}$ where the principal offers random exploration incentives according to Algorithm \ref{alg:pr} and the agent performs a greedy exploitation by picking the utility-maximizer arm according to Algorithm \ref{alg:ag}. Then, the probability that the agent does not select the true utility-maximizer arm at time $t$ is bounded by
\begin{align}
     \mathbb{P}\left(\upsilon_t(\boldsymbol{\pi}_t) \neq \argmax_{a \in \mathcal{A}} s^0_a + \pi_{t, a} \Big | t \in \mathcal{T}^{\mathrm{ag-xploit}} \cap \mathcal{T}^{\mathrm{pr-xplore}}\right) &\leq  \frac{4n^2\sqrt{n}\sqrt{\log 2t}}{\sqrt{m^{\mathrm{ag}}}\sqrt[4]{t}} + \frac{2n^2(\exp(2m^{\mathrm{ag}}) + 1)}{t-1}
\end{align}
\end{lemma}

In our analysis for the agent's learning process, we need to take into account the externality imposed by the principal's incentives in each period which requires considering the principal's exploration/exploitation plays separately. Lemmas \ref{lem:validate1} and \ref{lem:validate2} correspond to our results for each of these cases, and their detailed proofs are provided in Appendix \ref{appendix3}. Next, we combine these intermediate results and obtain our final proposition.  
\begin{proposition} \label{prop:validate}
    The probability that the agent's algorithm (\ref{alg:ag}) does not select the true utility-maximizer arm at time $t \in [\widetilde{k}, T]$ satisfies $\mathbb{P}\left(\upsilon_t(\boldsymbol{\pi}_t) \neq \argmax_{a \in \mathcal{A}} s^0_a + \pi_{t, a} \right) = O\left( \dfrac{\sqrt{\log 2t}}{\sqrt{t}}\right)$. 
\end{proposition}

Proposition \ref{prop:validate} is proven by using the mathematical induction technique. Our theoretical analysis in this section shows that the probability bound associated with the inaccuracy of the arms chosen by the agent directly depends on: i) exploration rates of the agent and the principal, ii) the agent's estimation gap between $\widehat{\mathbf{s}}^\mathrm{ag}_t$ and $\mathbf{s}^0$, and iii) the principal's estimation gap between $\widehat{\mathbf{s}}^{\mathrm{pr}}_t$ and $\mathbf{s}^0$. The detailed proof is provided in Appendix \ref{appendix3}. 
\begin{remark}
    Proposition \ref{prop:validate} shows the validity and mildness of Assumption \ref{assm:agent} by proving that it is satisfied by a classical MAB algorithm. 
\end{remark}
\subsection{Agent's Information Rent}\label{subsec:information}
As in every adverse selection model, there is an unavoidable information rent that the agent extracts from the principal due to the information asymmetry inherent in the considered hidden rewards setting. Our data-driven incentive design framework aims to minimize the extra amount principal the pays to the agent on top of this minimal amount of information rent by inducing the agent's incentive compatibility. We implicitly ensure this when we optimize the principal's exploitation incentives (\ref{alg-steps1})--(\ref{alg-stepslast}) which are designed to drive the agent pick the arm that the principal wants them to pick. However, our analysis assumes that the incentivized agent acts truthfully in the sense that their choices are made in a consistent way with their estimated expected reward vector $\widehat{\mathbf{s}}^\mathrm{ag}_t$ – which may not be the case in practice. 

From the standpoint of the self-interested agent, even though the agent is able to consistently learn their true rewards throughout the sequential game, we observe that they could just pretend that their learned rewards are different and make their choices in accordance with these ``pretended'' rewards to extract a higher information rent from the principal and maximize their total utilities. In that regard, the agent's problem can be formalized as an optimization model that maximizes their information rent where the optimum objective value can be achieved when the agent is also knowledgeable about the principal's rewards so that they can misinform the principal and demand extra payment from them. It is essential to notice that there is no way for the principal to prohibit the agent from this misbehavior under the hidden rewards setting whereas it could be possible in other adverse selection models where the principal has access to more information about the agent's utility model. Analyzing such settings is beyond the scope of this paper, yet it stands as an interesting future research direction.

Our framework considers a self-interested agent with particular sophistication who is not attempting to learn the principal’s rewards. Therefore, our solution can be regarded as yielding the worst-case result for an agent with higher sophistication. On the other hand, in most of the real-life applications that we consider, the incentivized agents do not have enough elaboration on learning the principal's model. Further, as their knowledge level decreases, they can only get smaller amounts of information rent from the principal. In any case, our approach is designed to minimize the principal's payments and maximize their cumulative expected net reward subject to any amount of information rent that the agent takes regardless of their level of complexity. 
\section{Numerical Simulations} \label{sec:numerical}
This section presents the numerical results supporting our theoretical bounds on the finite-sample concentration of the proposed estimator and on the convergence of the proposed incentive policy. We conduct simulation experiments on an instance of the contract design problem described in Section \ref{subsubsec:aggregator} for grid-scale distributed storage and aggregation of renewable energy.

Consider an independent aggregator company that operates distributed rechargeable battery storage systems to offer an affordable and environmentally sustainable alternative to the utility grid, and a utility company who acknowledges that they can reduce their costs and carbon footprint and improve reliability of their service by contracting the extra supply capacity available in the battery systems of the aggregator during the peak demand or extreme weather emergency periods. These contracts are done by participation of the aggregator in the secondary reserve market that allows the aggregator generate income by providing upward reserve capacities to the utility. The utility needs to consistently estimate the aggregator's net income to design smart and efficient payments that will expand the aggregator's participation on the upward reserve. 

We model this problem by using the repeated adverse selection model given in Section \ref{subsec:model}. Each contract is done on an hourly basis where the utility company (i.e., principal) offers a price for energy delivered (dollar/MW) by the aggregator (i.e., agent) as secondary upward reserve for each hour of the next day. Each time step $t$ in our repeated model corresponds to an hourly reserve session over the entire period $\mathcal{T}$ of contract signed between the aggregator and the utility. The aggregator's problem is formalized as a MAB model where each arm $a \in \mathcal{A}$ corresponds to a different range of battery capacity (MW) they reserve for the use of the utility. The maximum amount of upward reserve capacity is bounded by the maximum total amount of energy that can be consumed by the aggregator's entire battery systems. For example, if an electric vehicle is consuming 2 kW in an hour and the aggregator owns one thousand electric vehicles connected in residential charging points of the clients who allow the aggregator to control their charging processes, then the aggregator can provide upward reserve up to 2 MW per hour \citep{bessa2011optimized}. In this scenario, we can divide this feasible interval of upward reserve capacity $[0, 2]$ into a discrete number of subintervals to define the aggregator's arms, for example $\mathcal{A} = \{[0, 0.5], [0.5, 1], [1, 1.5], [1.5, 2]\}$. In each reserve session, the demand response software managed by the aggregator chooses an arm $\upsilon_{t}(\boldsymbol{\pi}_t)$ in response to the payment scheme $\boldsymbol{\pi}_t$ offered by the utility company. This choice and several sources of uncertainty that are uncontrollable, such as variations in renewable energy generation, observed electricity demand in the community, and market prices, collectively determine the realized profits of the aggregator ($\rho_{t,\upsilon_{t}(\boldsymbol{\pi}_t)}$) and the utility company ($\mu_{t,\upsilon_{t}(\boldsymbol{\pi}_t)}$) in that specific reserve session. 

We run our experiments for multiple combinations of the dimension of the aggregator's MAB model $n \in \{5, 10\}$ and the length of the contract period $T \in \{10^3, 5\cdot10^3, 10\cdot10^3, 20\cdot10^3, 40\cdot10^3\}$. Each setting is replicated five times, and the average and standard deviation of our performance metrics are reported across these replicates. The realizations of the profits of the aggregator and those of the utility company are assumed to follow Gaussian distributions, $\mathcal{N}(\mathbf{r}^0, 10)$ and $\mathcal{N}(\boldsymbol{\theta}^0, 10)$, respectively, where the values used for the expectations $\mathbf{r}^0$ and $\boldsymbol{\theta}^0$ are given in Table \ref{table:parameters} in Appendix \ref{appendix4}. Further, we take the compact set to which the aggregator's expected profits belong as $\mathcal{R} = [-20, 50]$ and the feasible range of the utility's payments as $\mathcal{C} = [-20, 60]$ (which makes $\gamma = 10$ as introduced in Assumption \ref{assm1}). The input parameters for the utility's algorithm (\ref{alg:pr}) are chosen as $m^{\mathrm{pr}} = 5$ and $w = 1/5$ in all settings which implies that the utility explores with probability one during the first $214-n$ reserve sessions after the initialization period (\ref{alg-initial1}-\ref{alg-initiallast}). To simulate the aggregator's choices, we use algorithm (\ref{alg:ag}) with parameter $m^{\mathrm{ag}} = 10$ which implies that the aggregator explores with probability one during the first $100-n$ reserve sessions after the initialization period (\ref{agent-alg-initial1}-\ref{agent-alg-initiallast}). 

We consider three main metrics to demonstrate the performance of our approach in the considered experimental setting: 1) concentration of our estimator towards the true expected profit vector $\mathbf{s}^0$, 2) convergence of our payment policy to the oracle payment policy introduced in Section \ref{subsubsec:oracle}, and 3) cumulative regret of the proposed united data-driven framework based on the defined regret notion (\ref{eq:regretdefn}). 

\begin{figure}\captionsetup[subfigure]{font=footnotesize}
    \centering
    \begin{subfigure}{.49\textwidth}
    \includegraphics[width=1.11\textwidth]{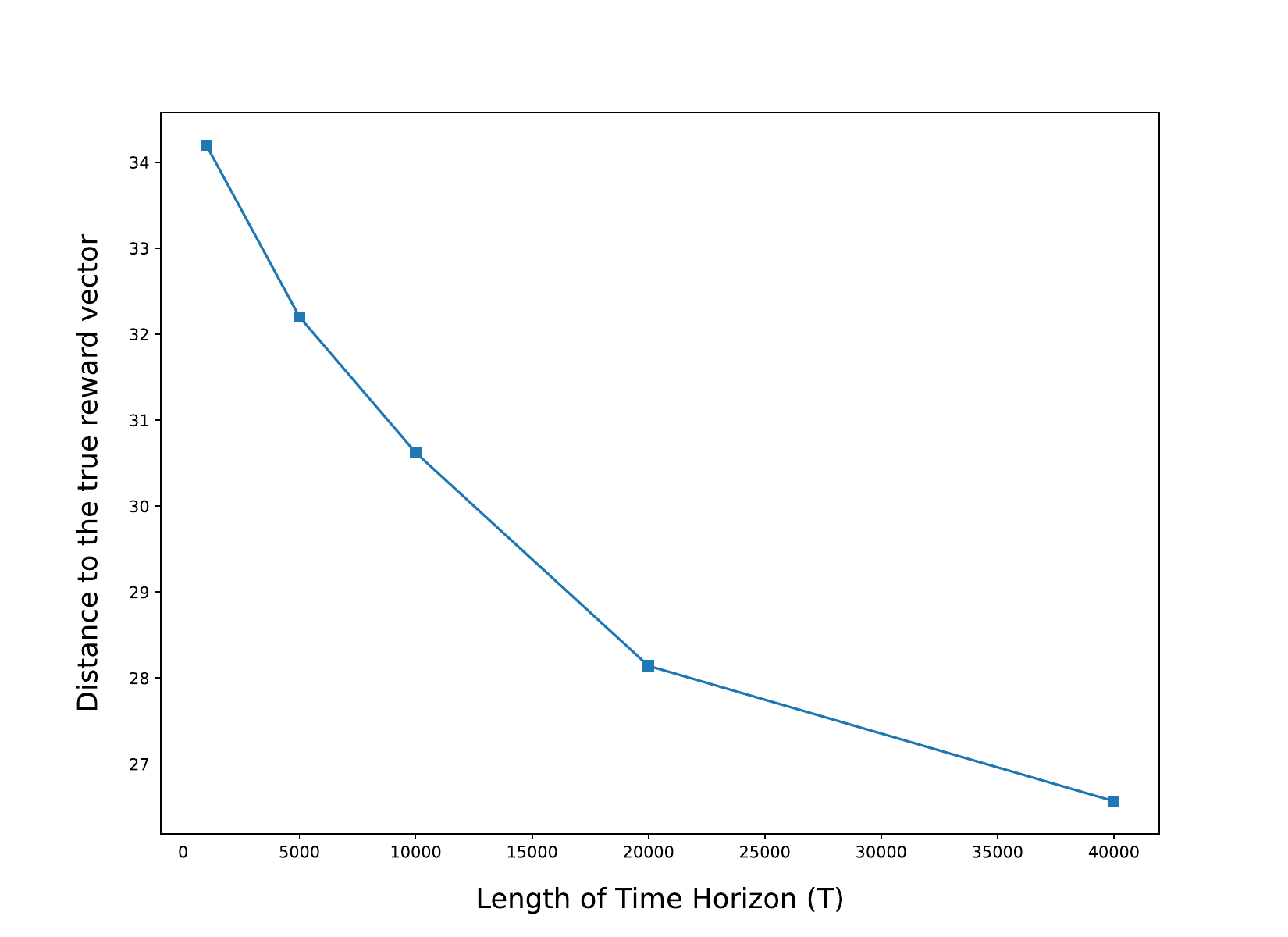}
    \caption{$\|\mathbf{s}^0 - \widehat{\mathbf{s}}^{\mathrm{pr}}_T \|_\infty$ for $n=5$}
    \label{fig:error-5}
    \end{subfigure}
    \begin{subfigure}{.49\textwidth}
    \includegraphics[width=1.11\textwidth]{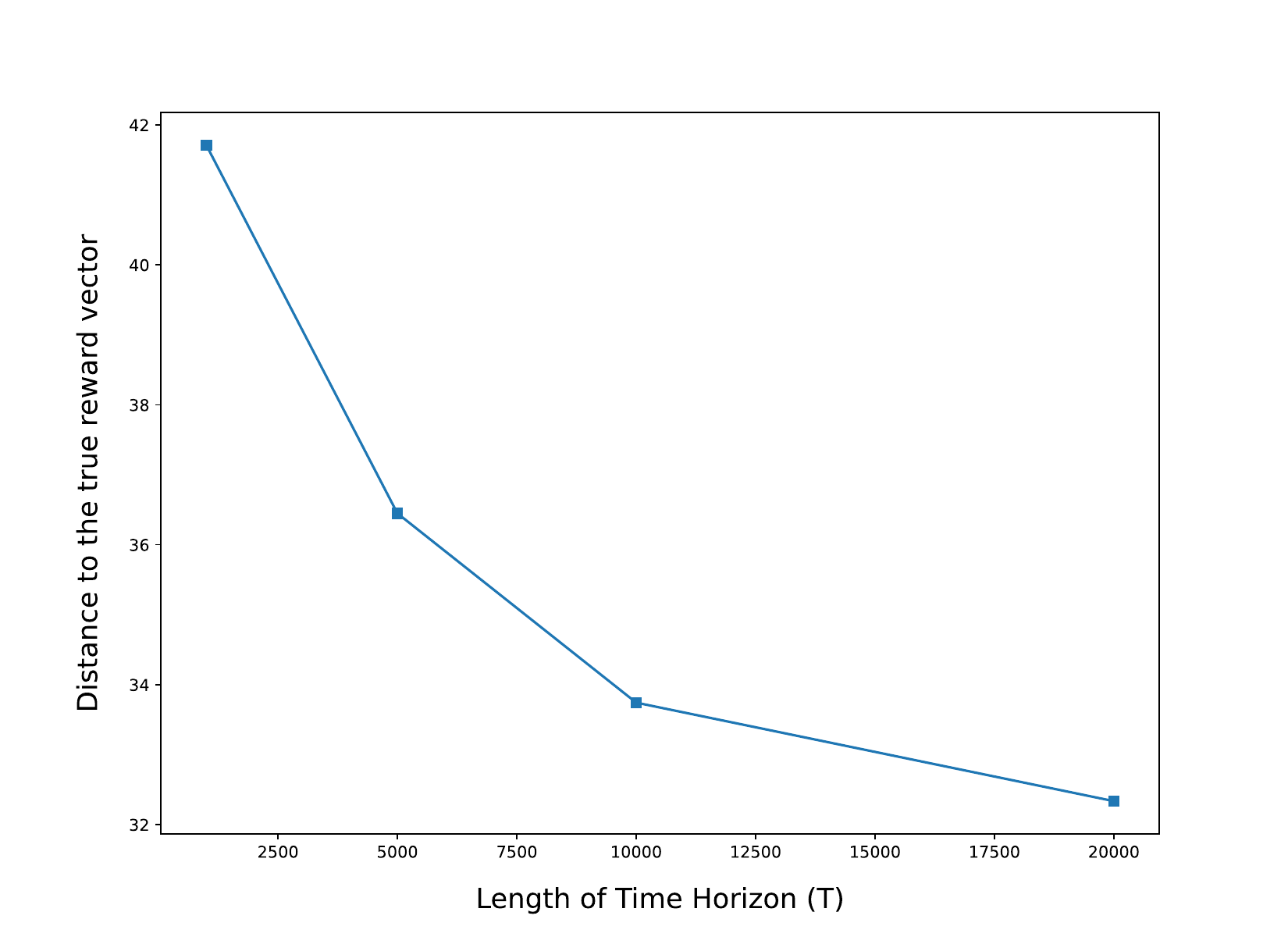}
    \caption{$\|\mathbf{s}^0 - \widehat{\mathbf{s}}^{\mathrm{pr}}_T \|_\infty$ for $n=10$}
    \label{fig:error-10}
    \end{subfigure}
    \caption{Estimator concentration measured in terms of the $\ell_\infty$ distance between the true mean profit vector $\mathbf{s}^0$ and the final mean profit vector $\widehat{\mathbf{s}}^{\mathrm{pr}}_T$ obtained by the proposed estimator.}
    \label{fig:error}
\end{figure}

First, we provide a direct measure of the accuracy of the proposed estimator (\ref{prblm:estimator}) for the aggregator's mean profits to support our finite-sample concentration results proven in Section \ref{sec:estimator}. In Figure \ref{fig:error}, we present the $\ell_\infty$-distance between the final estimated mean profit vector $\widehat{\mathbf{s}}^{\mathrm{pr}}_T$ and the true mean profit vector $\mathbf{s}^0$. These results display the consistency of our novel estimation approach and the achieved accuracy even when the size of the considered MAB model is doubled.   

\begin{figure}\captionsetup[subfigure]{font=footnotesize}
    \centering
    \begin{subfigure}{.49\textwidth}
    \includegraphics[width=1.11\textwidth]{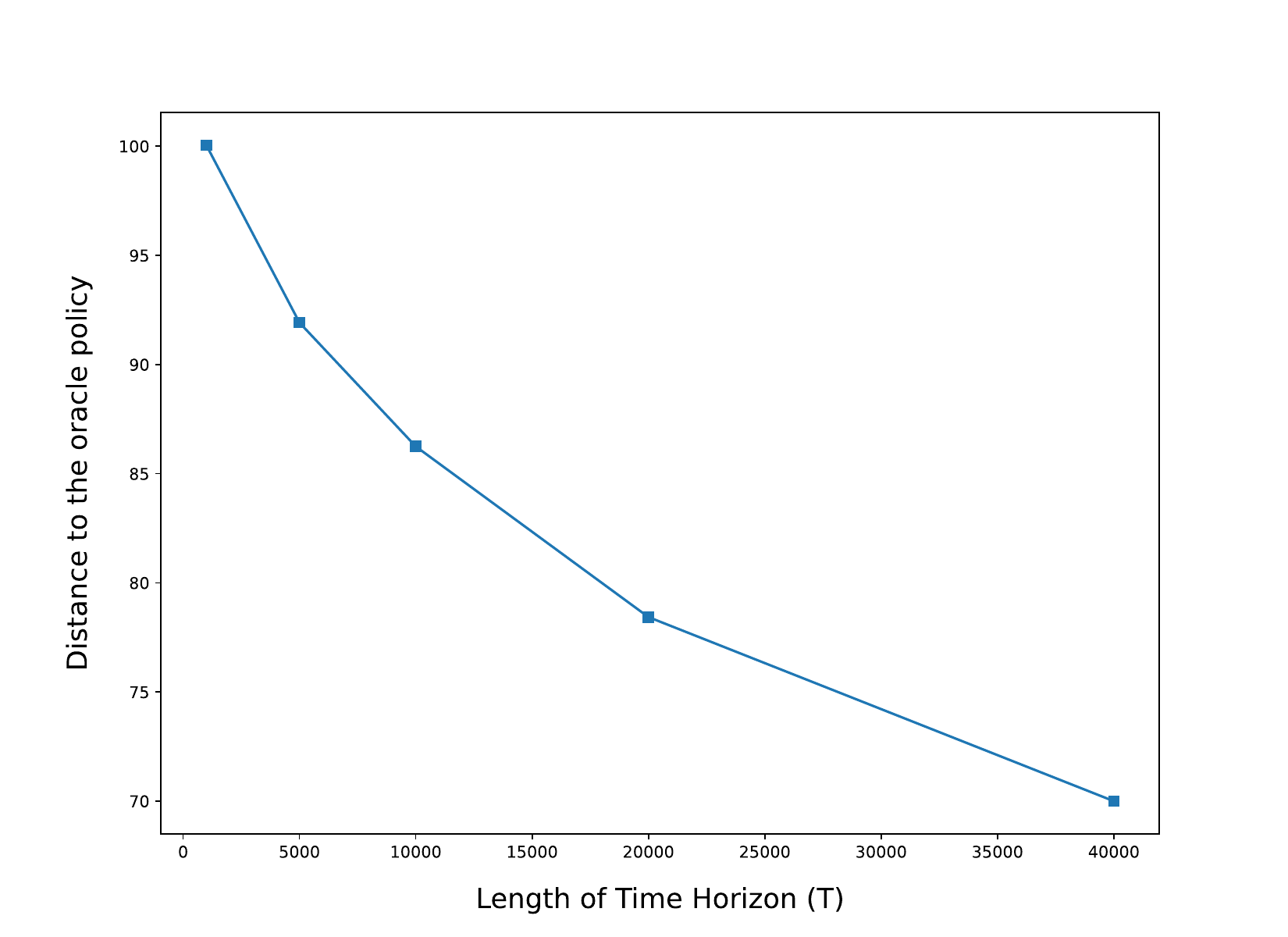}
    \caption{$\|\boldsymbol{\pi}_T - \mathbf{c}(\boldsymbol{\theta}^0, \mathbf{s}^0)\|_1$ for $n=5$}
    \label{fig:l1norm-5}
    \end{subfigure}
    \begin{subfigure}{.49\textwidth}
    \includegraphics[width=1.11\textwidth]{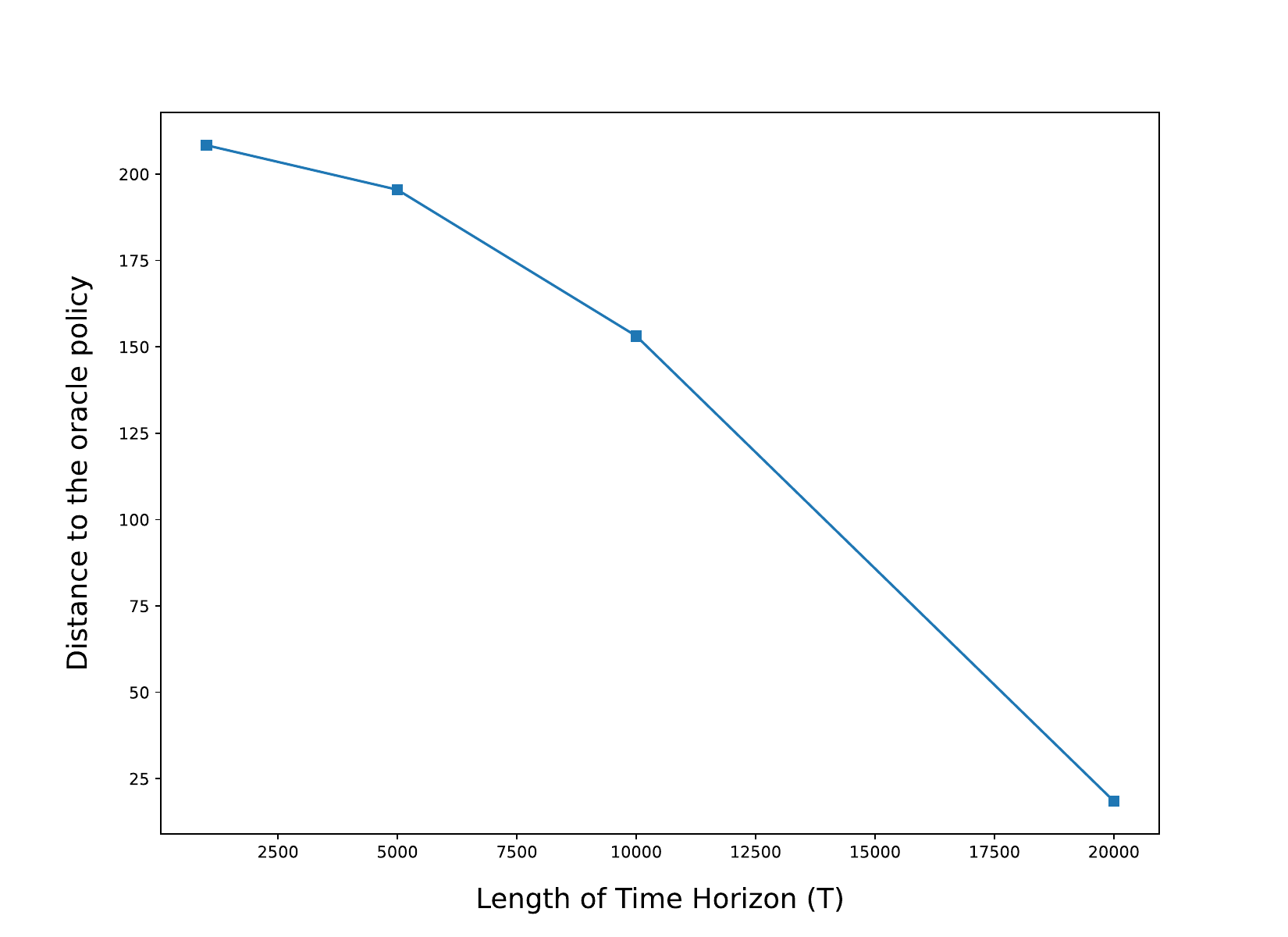}
    \caption{$\|\boldsymbol{\pi}_T - \mathbf{c}(\boldsymbol{\theta}^0, \mathbf{s}^0)\|_1$ for $n=10$}
    \label{fig:l1norm-10}
    \end{subfigure}
    \caption{Policy convergence measured in terms of the $\ell_1$ distance between the oracle payments $\mathbf{c}(\boldsymbol{\theta}^0, \mathbf{s}^0)$ and the payments $\boldsymbol{\pi}_T$ reached by Algorithm \ref{alg:pr} at the end of the contract horizon. $\big($For any two vectors $\mathbf{x}, \mathbf{y} \in \mathbb{R}^M$, the $\ell_1$ distance is defined by $\| \mathbf{x} - \mathbf{y}\|_1 = \sum_{m = 1}^M |x_m - y_m|$.$\big)$}
    \label{fig:l1norm}
\end{figure}

Second, we measure the convergence of our payment policy generated by Algorithm \ref{alg:pr} to the oracle payment policy. A significant challenge in the utility's contract design problem is that the inherent learning problem gets harder as the number of arms (different upward reserve ranges) of the aggregator gets larger. That is because they need to compute a payment amount for each possible arm (not only for the desired one) as accurately as possible in order to steer the aggregator's participation in the sequential secondary reserve sessions and optimize their ultimate objective. Because every alternative arm matters the same, we measure the distance between the proposed payment vector and the oracle payment vector in terms of the $\ell_1$ norm metric which weights all the entries of the vectors equally. As seen in Figure \ref{fig:l1norm}, the proposed payment mechanism is able to consistently converge to the oracle mechanism, and it achieves a better convergence as the length of the contract horizon gets longer. Further, a comparison of Figures \ref{fig:l1norm-5} and \ref{fig:l1norm-10} reveals the highlighted challenge regarding the relation between the difficulty of the contract design problem and the dimension of the considered MAB model. 

\begin{figure}[h]\captionsetup[subfigure]{font=footnotesize}
    \centering
    \begin{subfigure}{.49\textwidth}
    \includegraphics[width=1.11\textwidth]{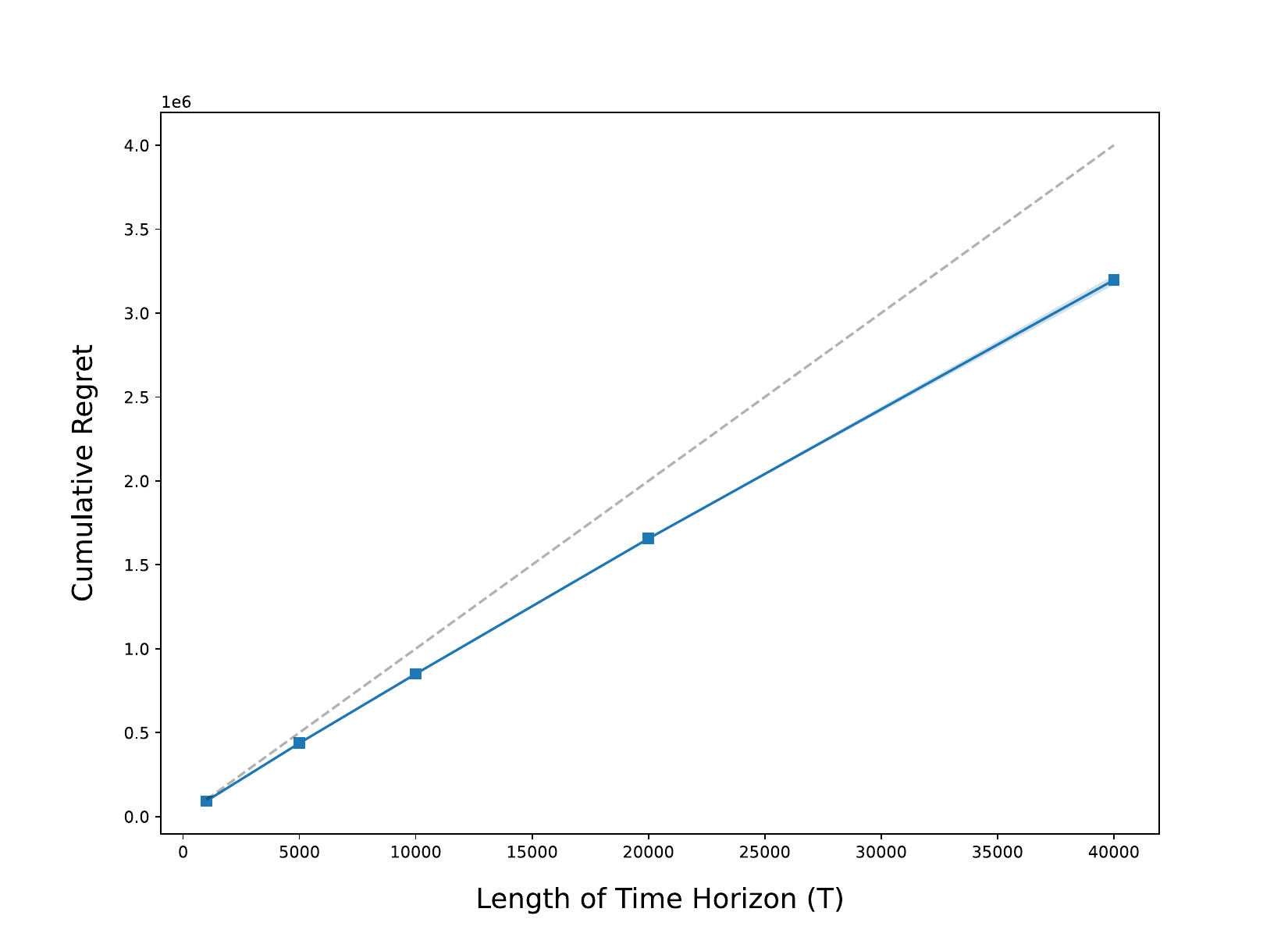}
    \caption{Regret for $n=5$}
    \end{subfigure}
    \begin{subfigure}{.49\textwidth}
    \includegraphics[width=1.11\textwidth]{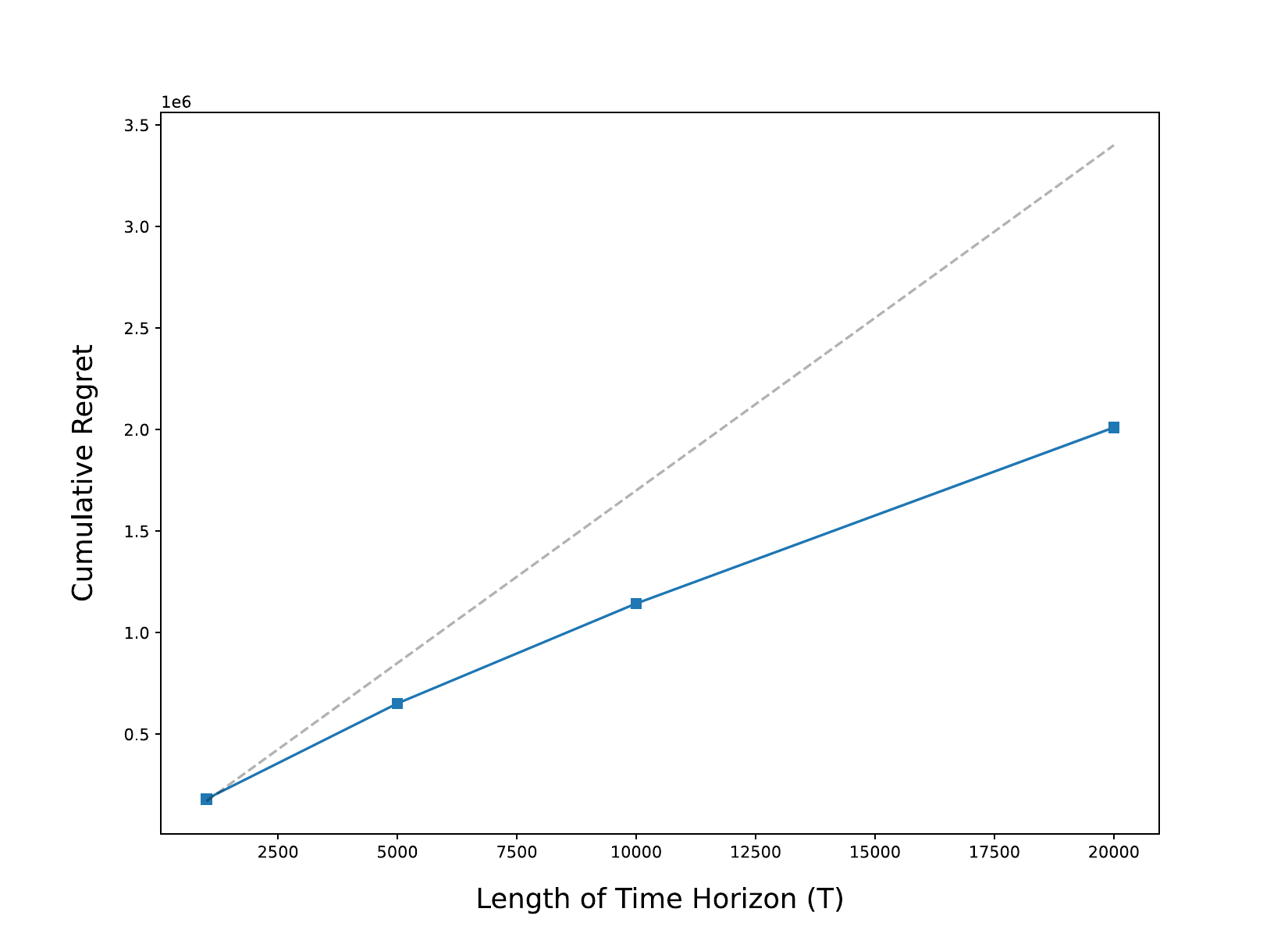}
    \caption{Regret for $n=10$}
    \end{subfigure}
    \caption{The average cumulative regret of the policies generated by Algorithm \ref{alg:pr}.}
    \label{fig:regret}
\end{figure}

Last, we present the overall performance of our adaptive framework where we unite our consistent estimator with the proposed efficient payments. Figure \ref{fig:regret} shows the cumulative regret accrued by the utility company for different values of $n$ and $T$. As expected, our approach achieves a sublinear regret that matches with the asymptotic order proven by our theoretical analysis as highlighted in Remark \ref{rem:regret}. 
\section{Conclusions and Foresight} \label{sec:conclusion}

Motivated by a variety of unexplored incentive design applications, this study aims to bridge the well-established principal-agent theory with  recent advances in online sequential learning theory. We model a generic and practically relevant repeated adverse selection game within a MAB framework where a principal incentivizes a utility-maximizer imperfect-knowledge agent by only watching the agent's decisions in response to the offered incentives over a finite time horizon. To lead the agent effectively, the principal needs to consistently learn the agent's true utilities (which are also unknown to the agent) without observing the agent's random reward realizations in each play of the game, and then they need to utilize these estimated values to optimize their incentives while ensuring the agent's incentive compatibility. Because we consider an imperfect-knowledge agent in this setting, we engage with two parallel learning algorithms that are trained by the principal and the agent under a two-way dynamic interaction with each other over the course of the repeated game. In this challenging setting, we address both the estimation/learning problem and the sequential incentive design problem that the principal faces by primarily focusing on their exploration/exploitation trade-off and providing convergence guarantees without restricting the type of the agent's algorithm. Our main theoretical contributions can be summarized as: 1) introducing a novel estimator and conducting a statistical analysis that proves its finite-sample concentration bound, 2) designing efficient data-driven incentives and uniting them with the proposed estimator within a practical $\epsilon$-greedy algorithm, and 3) performing a rigorous regret analysis for the united adaptive incentive design framework. We also provide insights from the perspective of the selfish learning agent whose goal is to maximize their utilities by extracting higher information rents from the principal. Lastly, we reinforce our theoretical results with numerical experiments that justify the applicability, effectiveness, and efficiency of the proposed framework to manage the renewable energy aggregation operations and achieve a smart and reliable grid for communities. 

The \emph{hidden agent rewards} setting remains mostly unexplored in the literature due to the complexity of jointly studying the learning and incentive design problems. However, we believe our contributions in this paper can channel promising future work for the data-driven contract design literature. One interesting direction would be to consider a multi-agent setting where a principal gets to collaborate with multiple utility-maximizer selfish agents. We believe that our model and approach is applicable to scenarios where the agents collectively work as a team and the principal provides team incentives based on the observed team-level decisions, whereas studying a scenario where the incentives are needed to be designed separately for each individual selfish-agent (that might be also communicating with other agents) would require a completely different approach and theoretical analysis. Furthermore, the repeated adverse selection model that we introduce in this paper is designed purposefully to be as generic and simple as possible to improve practical relevance for various domains. However, our model can be further extended and specialized to accommodate features of a particular application that may attract the operations management practitioners. 

\ACKNOWLEDGMENT{This material is based upon work partially supported by the National Science Foundation under Grant CMMI-184766.}


\bibliographystyle{informs2014} 
\bibliography{Stochastic_Refs} 


\begin{APPENDICES}
\section{Proofs of All Results in the Main Text}
\subsection{Results in Section \ref{sec:estimator}} \label{appendix1}
\proof{\textbf{Proof of Proposition \ref{prop:iden1}.}}

Given that the agent selects the true utility-maximizer arm in the considered period $t$, i.e., $\upsilon_t(\boldsymbol{\pi}_t) = \argmax_{a \in \mathcal{A}} \left( s^0_a + \pi_{t, a} \right)$, the construction of our estimator (\ref{prblm:estimator}) implies that  the one-stage loss function $\ell\left(\mathbf{s}, \upsilon_{t}(\boldsymbol{\pi}_t), \boldsymbol{\pi}_t\right)$ can be strictly positive if and only if the selected arm satisfies $\upsilon_{t}(\boldsymbol{\pi}_t) \neq \argmax_{a \in \mathcal{A}} \left(s_a + \pi_{t,a}\right)$ for the given vector $\mathbf{s} \in \mathcal{B}(\mathbf{s}^j, d), j \in \{1, \ldots, q\}$. 

Because we consider the scenario when $K^0 \cap K = \emptyset$, we already observe different maximizer indices for the true rewards $\mathbf{s}^0$ and the considered rewards $\mathbf{s}$ before adding the incentives. Therefore, we can simply choose a set of incentives such that the new maximizers after adding the incentives will still belong to the sets $K$ and $K^0$. For that purpose, we consider the incentives 
\begin{align}
    &\pi_{t,a} < R_{\min} + \gamma + \beta - d \ \text{for all }  a \in \mathcal{A} \setminus \{\kappa, \kappa^0\} \label{eq:s-smallc's-part1} \allowdisplaybreaks  \\
    &\pi_{t,a} \geq R_{\min} + \gamma + \beta - d \ \text{for } a \in \{\kappa, \kappa^0\} \label{eq:s-smallc's-part2} 
\end{align} 
for any $\kappa \in K, \ \kappa^0 \in K^0$ where $\gamma$ is as introduced in Assumption \ref{assm1}. We note that (\ref{eq:s-smallc's-part1}) and (\ref{eq:s-smallc's-part2}) are valid conditions based on Assumption \ref{assm1}. 

For the rest of our analysis, we define $\widetilde{\mathbf{s}}^j := \arginf_{\mathbf{s} \in \mathcal{B}(\mathbf{s}^j, d)} \|\mathbf{s}^0 - \mathbf{s}\|_\infty $ as the closest vector (with respect to the $\ell_\infty$-norm) in ball $\mathcal{B}(\mathbf{s}^j, d)$ to the true reward vector $\mathbf{s}^0$. Then, we have $\|\mathbf{s}^0 - \widetilde{\mathbf{s}}^j\|_\infty \geq \beta - d$ by construction, and we obtain
\begin{align}
    &\mathbb{P}\left(\ell\left(\mathbf{s}, \upsilon_{t}(\boldsymbol{\pi}_t), \boldsymbol{\pi}_t\right) \geq \omicron \Big | \upsilon_t(\boldsymbol{\pi}_t) = \argmax_{a \in \mathcal{A}} \left( s^0_a + \pi_{t, a} \right)\right) \nonumber  \allowdisplaybreaks  \\
    &= \mathbb{P}\left(\max_{a \in \mathcal{A}} \left(s_a + \pi_{t,a}\right) - s_{\upsilon_{t}(\boldsymbol{\pi}_t)} - \pi_{t, \upsilon_{t}(\boldsymbol{\pi}_t)} \geq \omicron \Big | \upsilon_t(\boldsymbol{\pi}_t) = \argmax_{a \in \mathcal{A}} \left( s^0_a + \pi_{t, a} \right)\right)  \allowdisplaybreaks  \\
    &\geq \mathbb{P}\bigg( \bigcup_{x \in \mathcal{A}, y \in \mathcal{A}, y \neq x} x = \argmax_{a \in \mathcal{A}}\left(s^0_a  + \pi_{t,a} \right), y = \argmax_{a \in \mathcal{A}}\left(s_a  + \pi_{t,a} \right), s_y + \pi_{t,y} - s_x - \pi_{t, x} \geq \omicron \bigg)   \allowdisplaybreaks  \\
    &\geq \mathbb{P}\left(\kappa = \argmax_{a \in \mathcal{A}} (s_a + \pi_{t,a}),\kappa^0 = \argmax_{a \in \mathcal{A}} (s^0_a + \pi_{t,a}), s_{\kappa} + \pi_{t,\kappa} - s_{\kappa^0} - \pi_{t, \kappa^0} \geq \omicron\right) \allowdisplaybreaks  \\
    & \geq \mathbb{P}\left(\kappa = \argmax_{a \in \mathcal{A}} (s_a + \pi_{t,a}),\kappa^0 = \argmax_{a \in \mathcal{A}} (s^0_a + \pi_{t,a}), s_{\kappa} + \pi_{t,\kappa} - s_{\kappa^0} - \pi_{t, \kappa^0} \geq \omicron \Big | (\ref{eq:s-smallc's-part1}), (\ref{eq:s-smallc's-part2})\right) \notag  \allowdisplaybreaks  \\
    &\hspace{.5cm} \cdot \mathbb{P}\left((\ref{eq:s-smallc's-part1}), (\ref{eq:s-smallc's-part2})\right)   \allowdisplaybreaks  \\
    & = \mathbb{P}\left(s_{\kappa^0} - s_{\kappa} + \omicron \leq \pi_{t,\kappa} - \pi_{t,\kappa^0} < s^0_{\kappa^0} - s^0_{\kappa} \right) 
    \prod_{a \in \{\kappa, \kappa^0 \}} \mathbb{P}\left(\pi_{t,a} \geq R_{\min} + \gamma + \beta - d \right) \notag  \allowdisplaybreaks  \\
    &\hspace{.5cm} \cdot \prod_{a \in \mathcal{A} \setminus \{\kappa, \kappa^0 \}}  \mathbb{P} \left(\pi_{t,a} < R_{\min} + \gamma + \beta - d \right)   \label{eq:s-byindepc's} \allowdisplaybreaks  \\
    &\geq \mathbb{P}\left(s_{\kappa^0} - s_{\kappa} + \omicron \leq \pi_{t,\kappa} - \pi_{t,\kappa^0} < s^0_{\kappa^0} - s^0_{\kappa} \right) 
    \prod_{a \in \{\kappa, \kappa^0 \}} \mathbb{P}\left(\pi_{t,a} \geq R_{\min} + \gamma + \beta - d \right) \notag  \allowdisplaybreaks  \\
    &\hspace{.5cm} \cdot \prod_{a \in \mathcal{A} \setminus \{\kappa, \kappa^0 \}} \mathbb{P} \left(\pi_{t,a} \leq R_{\min} + \gamma \right)   \allowdisplaybreaks  \\
    &= \mathbb{P}\left(s_{\kappa^0} - s_{\kappa} + \omicron \leq \pi_{t,\kappa} - \pi_{t,\kappa^0} < s^0_{\kappa^0} - s^0_{\kappa} \right)  \prod_{a \in \{\kappa, \kappa^0 \}} \left(1 - \frac{R_{\min} + \gamma + \beta - d - \underline{C}}{\overline{C} - \underline{C}} \right) \notag  \allowdisplaybreaks  \\
    &\hspace{.5cm} \cdot \prod_{a \in \mathcal{A} \setminus \{\kappa, \kappa^0 \}} \frac{R_{\min} + \gamma - \underline{C}}{\overline{C} - \underline{C}}   \label{eq:s-byuniformc's} \allowdisplaybreaks  \\
    &= \mathbb{P}\left(s_{\kappa^0} - s_{\kappa} + \omicron \leq \pi_{t,\kappa} - \pi_{t,\kappa^0} < s^0_{\kappa^0} - s^0_{\kappa} \right) \prod_{a \in \{\kappa, \kappa^0 \}} \left(1 - \frac{\gamma + \beta - d}{\overline{C} - \underline{C}} \right)  \prod_{a \in \mathcal{A} \setminus \{\kappa, \kappa^0 \}} \frac{\gamma}{\overline{C} - \underline{C}}   \label{eq:s-byrangeofc's}
\end{align}
where (\ref{eq:s-byindepc's}) follows since $\pi_{t,a}$'s are considered to be independent random variables, (\ref{eq:s-byuniformc's}) follows since $\pi_{t,a} \sim \mathcal{U}(\underline{C}, \overline{C}), \forall a \in \mathcal{A}$, and (\ref{eq:s-byrangeofc's}) follows since $\underline{C} = R_{\min}$ by Assumption \ref{assm1}. 

For the first term in (\ref{eq:s-byrangeofc's}), notice that the case that $s_{\kappa^0} - s_{\kappa} = s^0_{\kappa^0} - s^0_{\kappa} = 0$ cannot occur. This can only happen if $\kappa^0 \in K$ and $\kappa \in K^0$ which contradicts with the condition $K^0 \cap K_t = \emptyset$. Similarly,  $\mathbf{s}^0$ cannot be the all-zeros vector under the given condition $K^0 \cap K_t = \emptyset$. Further, since $0 < \omicron <  \delta = s^0_{\kappa^0} -  \max_{a \in \mathcal{A} \setminus \{K^0\}} s^0_a$ by definition, we know that $s_{\kappa^0} - s_{\kappa} + \omicron < s^0_{\kappa^0} - s^0_{\kappa}$ holds. This implies that the first probability term in (\ref{eq:s-byrangeofc's}) has a nonzero value and can be bounded by using the cumulative distribution function (cdf) of $\pi_{t,a} - \pi_{t,a'}$ which is the difference of two identically and independently distributed (iid) Uniform random variables. The difference $\pi_{t,a} - \pi_{t,a'}$ follows a triangular distribution whose cdf can be explicitly computed as follows. 
\begin{align}
    \mathbb{P}\left(\pi_{t,a} - \pi_{t,a'} \leq \Delta \right)  \label{eq:delta}
  &= \left\{\begin{array}{ll}
        0, & \text{for } \Delta < \underline{C}-\overline{C} \allowdisplaybreaks  \\
        \int\limits_{\underline{C}}^{\overline{C}+\Delta}  \int\limits_{\pi_{t,a} - \Delta}^{\overline{C}} \frac{1}{(\overline{C}-\underline{C})^2} d\pi_{t,a} d\pi_{t,a'}, & \text{for } \underline{C}-\overline{C} \leq \Delta < 0 \allowdisplaybreaks  \\
        1 - \int\limits_{\underline{C} + \Delta}^{\overline{C}}  \int\limits_{\underline{C}}^{\pi_{t,a} - \Delta} \frac{1}{(\overline{C}-\underline{C})^2} d\pi_{t,a} d\pi_{t,a'} , & \text{for } 0 \leq \Delta \leq \overline{C}-\underline{C} \allowdisplaybreaks  \\
        1 , & \text{for } \Delta \geq \overline{C}-\underline{C} \allowdisplaybreaks  \\
        \end{array}\right\} \allowdisplaybreaks  \\
    &= \left\{\begin{array}{ll}
        0, & \text{for } \Delta < \underline{C}-\overline{C} \allowdisplaybreaks  \\
        \frac{(\Delta + \overline{C} - \underline{C})^2 }{2(\overline{C}-\underline{C})^2}, & \text{for } \underline{C}-\overline{C} \leq \Delta < 0 \allowdisplaybreaks  \\
        1-\frac{\left(\Delta + \underline{C} - \overline{C}\right)^2 }{2(\overline{C}-\underline{C})^2} , & \text{for } 0 \leq \Delta \leq \overline{C}-\underline{C} \allowdisplaybreaks  \\
        1 , & \text{for } \Delta \geq \overline{C}-\underline{C} \allowdisplaybreaks  \\
        \end{array}\right\} \allowdisplaybreaks   \label{eq:cdf}
\end{align}
Using this cdf, we derive a strictly positive lower bound for the first term in (\ref{eq:s-byrangeofc's}).
\begin{align}
    &\mathbb{P}\left(s_{\kappa^0} - s_{\kappa} + \omicron \leq \pi_{t,\kappa} - \pi_{t,\kappa^0} < s^0_{\kappa^0} - s^0_{\kappa} \right) \nonumber  \allowdisplaybreaks  \\
    &\geq \mathbb{P}\left(s_{\kappa^0} - s_{\kappa} + \omicron \leq \pi_{t,\kappa} - \pi_{t,\kappa^0} < s^0_{\kappa^0} - s^0_{\kappa}, \ s_{\kappa^0} - s_{\kappa} + \omicron < 0 \right) \allowdisplaybreaks  \\
    &= 1 - \frac{(s^0_{\kappa^0} - s^0_{\kappa} + \underline{C} - \overline{C})^2}{2(\overline{C} - \underline{C})^2} - \frac{(s_{\kappa^0} - s_{\kappa} + \omicron + \overline{C} - \underline{C})^2}{2(\overline{C} - \underline{C})^2}   \allowdisplaybreaks  \\
    &= 1 - \frac{(s^0_{\kappa^0} - s^0_{\kappa})^2 + (s_{\kappa^0} - s_{\kappa} + \omicron)^2 + 2(\overline{C} - \underline{C})^2 + 2(\overline{C} - \underline{C})(s_{\kappa^0} - s_{\kappa} + \omicron - s^0_{\kappa^0} + s^0_{\kappa})}{2(\overline{C} - \underline{C})^2} \allowdisplaybreaks  \\
    &= \frac{-(s^0_{\kappa^0} - s^0_{\kappa})^2 - (s_{\kappa^0} - s_{\kappa} + \omicron)^2 + 2(\overline{C} - \underline{C})(s_{\kappa} - s_{\kappa^0} - \omicron + s^0_{\kappa^0} - s^0_{\kappa})}{2(\overline{C} - \underline{C})^2}   \allowdisplaybreaks  \\
    &\geq \frac{-(s^0_{\kappa^0} - s^0_{\kappa})^2 - (s_{\kappa^0} - s_{\kappa} + \omicron)^2 + 2(\overline{C} - \underline{C})(s^0_{\kappa^0} - s^0_{\kappa})}{2(\overline{C} - \underline{C})^2}   
    \intertext{where the last line follows since we consider the case $s_{\kappa^0} - s_{\kappa} + \omicron < 0$ for this lower bound. Then, }
    &\geq \frac{(s^0_{\kappa^0} - s^0_{\kappa})^2 -  \omicron^2}{2(\overline{C} - \underline{C})^2} > 0   
\end{align}
which follows since $s^0_{\kappa^0} - s^0_{\kappa} \leq \overline{C} - \underline{C}$ by definition and $s_{\kappa^0} - s_{\kappa} < 0$ based on the considered case. Combining this last result with (\ref{eq:s-byrangeofc's}), we obtain the following nonzero lower bound for the desired probability. 
\begin{multline}
    \mathbb{P}\left(\ell\left(\mathbf{s}, \upsilon_{t}(\boldsymbol{\pi}_t), \boldsymbol{\pi}_t\right) \geq \omicron \Big | \upsilon_t(\boldsymbol{\pi}_t) = \argmax_{a \in \mathcal{A}} \left( s^0_a + \pi_{t, a} \right)\right) \\ \geq \left(\frac{(s^0_{\kappa^0} - s^0_{\kappa})^2 -  \omicron^2}{2(\overline{C} - \underline{C})^2} \right)\left(1 - \frac{\gamma + \beta - d}{\overline{C} - \underline{C}} \right)^2  \left( \frac{\gamma}{\overline{C} - \underline{C}} \right)^{n-2}  
\end{multline}
\Halmos \endproof

\proof{\textbf{Proof of Proposition \ref{prop:iden2}.}}
Similar to our observation in the proof of Proposition \ref{prop:iden1}, the construction of our estimator (\ref{prblm:estimator}) implies that the one-stage loss function $\ell\left(\mathbf{s}, \upsilon_{t}(\boldsymbol{\pi}_t), \boldsymbol{\pi}_t\right)$ can be strictly positive if and only if we have $\upsilon_{t}(\boldsymbol{\pi}_t) \neq \argmax_{a \in \mathcal{A}} \left(s_a + \pi_{t,a}\right)$ for the given vector $\mathbf{s} \in \mathcal{B}(\mathbf{s}^j, d), j \in \{1, \ldots, q\}$ under the condition of $\upsilon_t(\boldsymbol{\pi}_t) = \argmax_{a \in \mathcal{A}} \left( s^0_a + \pi_{t, a} \right)$. Therefore, to prove the lower bound in (\ref{eq:identf-2}), we will consider the event where $\argmax_{a \in \mathcal{A}} (s_a + \pi_{t,a}) = 1$ and $\argmax_{a \in \mathcal{A}} (s^0_a + \pi_{t,a}) = b$ because of the fact that $b \neq 1$. 
As we have $s_1 = s^0_1 = 0$ by construction, having $b = 1$ would imply that $\mathbf{s}^0 = \mathbf{s} = \mathbf{0}_n$, and that would contradict with the definition of $\mathbf{s}$ which implies $\|\mathbf{s}^0 - \mathbf{s}\|_\infty = |s^0_b - s_b| > \beta$.

Let $\omega = \sup_{\mathbf{s} \in \mathcal{B}(\mathbf{s}^j, d)} \max_{a \in \mathcal{A}} \{|s^0_a|, |s_a| \}$ be the largest absolute value observed among the entries of $\mathbf{s}^0$ and of all vectors in $\mathcal{B}(\mathbf{s}^j, d)$. Then, we suppose
\begin{align}
    & \pi_{t,a} < R_{\min} + \gamma + \beta - d \ \text{for all }  a \in \mathcal{A} \setminus \{1, b\} \label{eq:s-smallc's-2-part1} \allowdisplaybreaks  \\
    & \pi_{t,a} \geq R_{\min} + \gamma + \omega  \ \text{for } a \in \{1, b\}. \label{eq:s-smallc's-2-part2}
\end{align} 
which are consistent with Assumption \ref{assm1}. These conditions imply that $s^0_{\kappa^0} + \pi_{t, \kappa^0} < s^0_a + \pi_{t, a}$ and $s_{\kappa} + \pi_{t, \kappa} < s_{a} + \pi_{t, a}$ hold for any $\kappa^0 \in K^0$, $\kappa \in K$, $a \in \{1, b\}$, and that the indices in the sets $K^0$ and $K$ are no more maximizers after adding the incentives in (\ref{eq:s-smallc's-2-part1})-(\ref{eq:s-smallc's-2-part2}). Thus, we will obtain the desired case (that is $\argmax_{a \in \mathcal{A}} (s_a + \pi_{t,a}) = 1$ and $\argmax_{a \in \mathcal{A}} (s^0_a + \pi_{t,a}) = b$) if the events $s^0_{1} + \pi_{t, 1} < s^0_{b} + \pi_{t, b}$ and $s_b + \pi_{t, b} < s_1 + \pi_{t, 1}$ hold.

Further, because $|s^0_b - s_b| > \beta$ by definition, we know that $|s^0_b - s_b| > |s^0_1 - s_1| = 0$. Suppose that without loss of generality, we have $s^0_b - s_b > s^0_1 - s_1 = 0$ and $s^0_b - s_b > \beta$. Then, our proof follows as 
\begin{align}
    &\mathbb{P}\left(\ell\left(\mathbf{s}, \upsilon_{t}(\boldsymbol{\pi}_t), \boldsymbol{\pi}_t\right) \geq \omicron \Big | \upsilon_t(\boldsymbol{\pi}_t) = \argmax_{a \in \mathcal{A}} \left( s^0_a + \pi_{t, a} \right)\right)  \nonumber \allowdisplaybreaks \\
    &= \mathbb{P}\left(\max_{a \in \mathcal{A}} \left(s_a + \pi_{t,a}\right) - s_{\upsilon_{t}(\boldsymbol{\pi}_t)} - \pi_{t, \upsilon_{t}(\boldsymbol{\pi}_t)} \geq \omicron \Big | \upsilon_t(\boldsymbol{\pi}_t) = \argmax_{a \in \mathcal{A}} \{ s^0_a + \pi_{t, a} \}\right)  \allowdisplaybreaks  \\
    &\geq \mathbb{P}\bigg( \bigcup_{x \in \mathcal{A}, y \in \mathcal{A}, y \neq x} x = \argmax_{a \in \mathcal{A}}\left(s^0_a  + \pi_{t,a} \right), y = \argmax_{a \in \mathcal{A}}\left(s_a + \pi_{t,a} \right), s_y + \pi_{t,y} - s_x - \pi_{t, x} \geq \omicron \bigg)   \allowdisplaybreaks  \\
    &\geq \mathbb{P}\left(1 = \argmax_{a \in \mathcal{A}} (s_a + \pi_{t,a}), b = \argmax_{a \in \mathcal{A}} (s^0_a + \pi_{t,a}), s_1 + \pi_{t,1} - s_b - \pi_{t, b} \geq \omicron\right)   \allowdisplaybreaks  \\
    &\geq \mathbb{P}\left(1 = \argmax_{a \in \mathcal{A}} (s_a + \pi_{t,a}), b = \argmax_{a \in \mathcal{A}} (s^0_a + \pi_{t,a}), s_1 + \pi_{t,1} - s_b - \pi_{t, b} \geq \omicron \Big |  (\ref{eq:s-smallc's-2-part1}), (\ref{eq:s-smallc's-2-part2}) \right) \mathbb{P}\left((\ref{eq:s-smallc's-2-part1}), (\ref{eq:s-smallc's-2-part2}) \right)  \allowdisplaybreaks  \\ 
    &= \mathbb{P}\left(s_b - s_1 + \omicron \leq \pi_{t, 1} - \pi_{t, b} < s^0_{b} - s^0_{1}\right) \prod_{a \in \{1, b\}} \mathbb{P} \left(\pi_{t,a} \geq R_{\min} + \gamma + \omega \right) \notag  \allowdisplaybreaks  \\
    &\hspace{.5cm} \cdot \prod_{a \in \mathcal{A} \setminus \{1, b\}} \mathbb{P} \left(\pi_{t,a} < R_{\min} + \gamma + \beta - d \right)  \allowdisplaybreaks  \\ 
    &\geq \mathbb{P}\left(s_b - s_1 + \omicron \leq \pi_{t, 1} - \pi_{t, b} < s^0_{b} - s^0_{1}\right)  \prod_{a \in \{1, b\}} \hspace{-2mm}  \mathbb{P} \left(\pi_{t,a} \geq R_{\min} + \gamma + \omega \right) 
    \prod_{a \in \mathcal{A} \setminus \{1, b\}} \mathbb{P} \left(\pi_{t,a} \leq R_{\min} + \gamma \right) \label{eq:s-byindepc's-2-1} 
     \allowdisplaybreaks  \\ 
    &= \mathbb{P}\left(s_b - s_1 + \omicron \leq \pi_{t, 1} - \pi_{t, b} < s^0_{b} - s^0_{1}\right) \prod_{a \in \{1, b\}} \left(1 - \frac{\gamma + \omega}{\overline{C} - \underline{C}} \right) \prod_{a \in \mathcal{A} \setminus \{1, b\}}  \frac{\gamma}{\overline{C} - \underline{C}}  \label{eq:s-byindepc's-2-2} 
\end{align}
where (\ref{eq:s-byindepc's-2-1}) and (\ref{eq:s-byindepc's-2-2}) follow since $\underline{C} = R_{\min}$ by Assumption \ref{assm1} and $\pi_{t,a}$'s are independent random variables with $\pi_{t,a} \sim \mathcal{U}(\underline{C}, \overline{C}), \forall a \in \mathcal{A}$. Next, we bound the first term in (\ref{eq:s-byindepc's-2-2}).
\begin{align}
    \mathbb{P}\left(s_b - s_1 + \omicron \leq \pi_{t, 1} - \pi_{t, b} < s^0_{b} - s^0_{1}\right) &= \mathbb{P}\left(s_b - s_1 + s^0_b - s^0_b  + \omicron \leq \pi_{t, 1} - \pi_{t, b} < s^0_{b} - s^0_{1}\right)  \label{eq:s-combine2-0}\allowdisplaybreaks  \\
     &\geq \mathbb{P}\left(s^0_{b} - \beta - s_1 + \omicron \leq \pi_{t, 1} - \pi_{t, b} < s^0_{b} - s^0_{1} \right)  \allowdisplaybreaks  \\
     &= \mathbb{P}\left(s^0_{b} - \beta + \omicron \leq \pi_{t, 1} - \pi_{t, b}  < s^0_{b} \right)   
\end{align}
We can further bound the last probability by using the cdf (\ref{eq:cdf}) that is defined as a piecewise function. For that purpose, we consider the following two mutually exclusive cases. 
\begin{itemize}
    \item[--] \textit{Case 1:} $\underline{C}-\overline{C} \leq s^0_{b} < 0$
    \item[--] \textit{Case 2:} $0 \leq s^0_{b} \leq \overline{C}-\underline{C}$ 
\end{itemize}
We also consider the following subcases to bound the two cases above.
\begin{itemize}
    \item[--] \textit{Subcase 1:} $\underline{C}-\overline{C} \leq s^0_{b} < 0$ and $\underline{C}-\overline{C} \leq s^0_{b} - \beta + \omicron < 0$
    \item[--] \textit{Subcase 2:} $0 \leq s^0_{b} \leq \overline{C}-\underline{C}$ and $0 \leq s^0_{b} - \beta + \omicron \leq \overline{C}-\underline{C}$
\end{itemize}
and compute the lower bounds for these subcases as follows.
\begin{align}
    \mathbb{P} \left(s^0_{b} - \beta + \omicron \leq \pi_{t, 1}  - \pi_{t,b} < s^0_{b}, \textit{Subcase 1} \right) &=  \frac{(s^0_{b} + \overline{C} - \underline{C})^2 }{2(\overline{C}-\underline{C})^2} - \frac{(s^0_{b} - \beta + \omicron + \overline{C} - \underline{C})^2 }{2(\overline{C}-\underline{C})^2}  \allowdisplaybreaks  \\
    &= \frac{(s^0_{b})^2 - (s^0_{b} - \beta + \omicron)^2 + 2(s^0_{b} - (s^0_{b} - \beta + \omicron))(\overline{C} - \underline{C})}{2(\overline{C}-\underline{C})^2}  \allowdisplaybreaks  \\
    &= \frac{(s^0_{b})^2 - (s^0_{b} - \beta + \omicron)^2 + 2(\beta - \omicron)(\overline{C} - \underline{C})}{2(\overline{C}-\underline{C})^2}  \allowdisplaybreaks  \\
    &= \frac{(s^0_{b})^2 - (s^0_{b})^2 -(\beta - \omicron)^2 + 2(\beta - \omicron)s^0_{b} + 2(\beta - \omicron)(\overline{C} - \underline{C}) }{2(\overline{C}-\underline{C})^2}  \allowdisplaybreaks  \\
    &= \frac{-(\beta - \omicron)^2 + 2(\beta - \omicron)(s^0_b  + \overline{C} - \underline{C})}{2(\overline{C}-\underline{C})^2}  \allowdisplaybreaks  \\
    &\geq \frac{-(\beta - \omicron)^2 + 2(\beta - \omicron)^2}{2(\overline{C}-\underline{C})^2}  \allowdisplaybreaks  \\
    &= \frac{(\beta - \omicron)^2}{2(\overline{C}-\underline{C})^2} 
\end{align}
where second to the last line follows since we have $0 < \beta - \omicron \leq s^0_b + \overline{C} - \underline{C}$ in this subcase. 
\begin{align}
    \mathbb{P} \left(s^0_{b} - \beta + \omicron \leq \pi_{t, 1}  - \pi_{t,b} < s^0_{b}, \textit{Subcase 2} \right) &= 1 - \frac{(s^0_{b} + \underline{C} - \overline{C})^2 }{2(\overline{C}-\underline{C})^2} - 1 + \frac{(s^0_{b} - \beta + \omicron + \underline{C} - \overline{C})^2 }{2(\overline{C}-\underline{C})^2}  \allowdisplaybreaks  \\
    &=  \frac{(\overline{C} - \underline{C} - s^0_{b} + \beta - \omicron)^2 }{2(\overline{C}-\underline{C})^2} - \frac{(\overline{C} - \underline{C} - s^0_{b})^2 }{2(\overline{C}-\underline{C})^2}  \allowdisplaybreaks  \\
    &= \frac{(\beta - \omicron)^2 + 2(\beta-\omicron)(\overline{C} - \underline{C} - s^0_{b})}{2(\overline{C}-\underline{C})^2}  \allowdisplaybreaks  \\
    &\geq \frac{(\beta - \omicron)^2}{2(\overline{C}-\underline{C})^2} 
\end{align}
where the last inequality follows since we have $\overline{C} - \underline{C} - s^0_{b} \geq 0$ and $\beta - \omicron > 0$ by definition. Next, we use the lower bounds for \textit{Subcase 1} and \textit{Subcase 2} to bound \textit{Case 1} and \textit{Case 2}.
\begin{align}
    &\mathbb{P} \left(s^0_{b} - \beta + \omicron \leq \pi_{t, 1}  - \pi_{t,b} < s^0_{b} \right) \nonumber  \allowdisplaybreaks  \\
    &= \mathbb{P} \left(s^0_{b} - \beta + \omicron \leq\pi_{t, 1}  - \pi_{t,b} < s^0_{b}, \textit{Case 1} \right) + \mathbb{P} \left(s^0_{b} - \beta + \omicron \leq \pi_{t, 1}  - \pi_{t,b} < s^0_{b},\textit{Case 2}\right)  \allowdisplaybreaks  \\
    &\geq  \mathbb{P} \left(s^0_{b} - \beta + \omicron \leq \pi_{t, 1}  - \pi_{t,b} < s^0_{b}, \textit{Subcase 1} \right) + \mathbb{P} \left(s^0_{b} - \beta + \omicron \leq \pi_{t, 1}  - \pi_{t,b} < s^0_{b},\textit{Subcase 2}\right)  \allowdisplaybreaks  \\
    &\geq \frac{(\beta - \omicron)^2}{(\overline{C}-\underline{C})^2}  \label{eq:s-combine2-last}
\end{align}
We conclude our proof by combining the last result with (\ref{eq:s-byindepc's-2-2}) which yields 
\begin{align}
    \mathbb{P}\left(\ell\left(\mathbf{s}, \upsilon_{t}(\boldsymbol{\pi}_t), \boldsymbol{\pi}_t\right) \geq \omicron \Big | \upsilon_t(\boldsymbol{\pi}_t) = \argmax_{a \in \mathcal{A}} \left( s^0_a + \pi_{t, a} \right)\right) &\geq  \frac{(\beta - \omicron)^2}{(\overline{C}-\underline{C})^2} \left(1 - \frac{\gamma + \omega}{\overline{C} - \underline{C}} \right)^2 \left( \frac{\gamma}{\overline{C} - \underline{C}}\right)^{n-2}  
\end{align}
\Halmos \endproof

\proof{\textbf{Proof of Proposition \ref{prop:iden3}.}}
Our proof mainly follows arguments similar to those in the proof of Proposition \ref{prop:iden2}. Recall that $b \neq 1$ (because $\mathbf{s}$ is defined such that $\|\mathbf{s}^0 - \mathbf{s}\|_\infty > \beta$) and that either $s_b >0$ or $s^0_b >0$ holds. We use the definition of $\omega = \sup_{\mathbf{s} \in \mathcal{B}(\mathbf{s}^j, d)} \max_{a \in \mathcal{A}} \{|s^0_a|, |s_a| \}$ and consider the following set of incentives.
\begin{align}
    &\pi_{t,a} < R_{\min} + \gamma + \beta - d \ \text{for all }  a \in \mathcal{A} \setminus \{1, b\} \label{eq:s-smallc's-3-part1} \allowdisplaybreaks  \\
    &\pi_{t,b} \geq R_{\min} + \gamma + \beta - d \label{eq:s-smallc's-3-part2}  \allowdisplaybreaks  \\
    &\pi_{t,1} \geq R_{\min} + \gamma + \omega \label{eq:s-smallc's-3-part3} 
\end{align} 
which are compatible with Assumption \ref{assm1}. By construction of $\mathbf{s}$, we know that $|s_b - s^0_{b}| > |s_1 - s^0_1| = 0$. Then, without loss of generality, we suppose that $s^0_b - s_b > s^0_1 - s_1 = 0$ and $s^0_b - s_b > \beta$.
\begin{align}
    &\mathbb{P}\left(\ell\left(\mathbf{s}, \upsilon_{t}(\boldsymbol{\pi}_t), \boldsymbol{\pi}_t\right) \geq \omicron \Big | \upsilon_t(\boldsymbol{\pi}_t) = \argmax_{a \in \mathcal{A}} \left( s^0_a + \pi_{t, a} \right)\right) \nonumber \allowdisplaybreaks  \\
    &= \mathbb{P}\left(\max_{a \in \mathcal{A}} \left(s_a + \pi_{t,a}\right) - s_{\upsilon_{t}(\boldsymbol{\pi}_t)} - \pi_{t, \upsilon_{t}(\boldsymbol{\pi}_t)} \geq \omicron \Big | \upsilon_t(\boldsymbol{\pi}_t) = \argmax_{a \in \mathcal{A}} \{ s^0_a + \pi_{t, a} \}\right)  \allowdisplaybreaks  \\
    &\geq \mathbb{P}\bigg( \bigcup_{x \in \mathcal{A}, y \in \mathcal{A}, y \neq x} x = \argmax_{a \in \mathcal{A}}\left(s^0_a  + \pi_{t,a} \right), y = \argmax_{a \in \mathcal{A}}\left(s_a  + \pi_{t,a} \right), s_y + \pi_{t,y} - s_x - \pi_{t, x} \geq \omicron \bigg)   \allowdisplaybreaks  \\
    &\geq \mathbb{P}\left(1 = \argmax_{a \in \mathcal{A}} (s_a + \pi_{t,a}), b = \argmax_{a \in \mathcal{A}} (s^0_a + \pi_{t,a}), s_1 + \pi_{t,1} - s_b - \pi_{t, b} \geq \omicron\right)   \allowdisplaybreaks  \\
    &\geq \mathbb{P}\left(1 = \argmax_{a \in \mathcal{A}} (s_a + \pi_{t,a}), b = \argmax_{a \in \mathcal{A}} (s^0_a + \pi_{t,a}), s_1 + \pi_{t,1} - s_b - \pi_{t, b} \geq \omicron \Big |  (\ref{eq:s-smallc's-3-part1}) - (\ref{eq:s-smallc's-3-part3}) \right) \notag  \allowdisplaybreaks  \\
    &\hspace{.5cm} \cdot \mathbb{P}\left((\ref{eq:s-smallc's-3-part1}) - (\ref{eq:s-smallc's-3-part3}) \right)  \allowdisplaybreaks  \\ 
    &= \mathbb{P}\left(s_b - s_1 + \omicron \leq \pi_{t, 1} - \pi_{t, b} < s^0_{b} - s^0_{1}\right) \mathbb{P}\left(\ref{eq:s-smallc's-3-part2}\right)\mathbb{P}\left(\ref{eq:s-smallc's-3-part3}\right) \prod_{a \in \mathcal{A} \setminus \{1, b\}}  \mathbb{P}\left( \pi_{t,a} < R_{\min} + \gamma + \beta - d \right)  \label{eq:s-byindepc's-3}\allowdisplaybreaks  \\ 
    &\geq \mathbb{P}\left(s_b - s_1 + \omicron \leq \pi_{t, 1} - \pi_{t, b} < s^0_{b} - s^0_{1}\right) \mathbb{P}\left(\ref{eq:s-smallc's-3-part2}\right)\mathbb{P}\left(\ref{eq:s-smallc's-3-part3}\right) \prod_{a \in \mathcal{A} \setminus \{1, b\}} \mathbb{P}\left( \pi_{t,a} \leq R_{\min} + \gamma\right)  \allowdisplaybreaks  \\ 
    &= \mathbb{P}\left(s_b - s_1 + \omicron \leq \pi_{t, 1} - \pi_{t, b} < s^0_{b} - s^0_{1}\right) \left(1 - \frac{\gamma + \beta - d}{\overline{C} - \underline{C}} \right) \left(1 - \frac{\gamma + \omega}{\overline{C} - \underline{C}} \right) \prod_{a \in \mathcal{A} \setminus \{1, b\}} \frac{\gamma}{\overline{C} - \underline{C}}   \label{eq:s-combine3}
\end{align}
where (\ref{eq:s-byindepc's-3}) follows as $\pi_{t,a}$'s are independent random variables and (\ref{eq:s-combine3}) follows since we assume that $\underline{C} = R_{\min}$ by Assumption \ref{assm1} and $\pi_{t,a} \sim \mathcal{U}(\underline{C}, \overline{C}), \forall a \in \mathcal{A}$. 

Then, by using similar arguments as in (\ref{eq:s-combine2-0})-(\ref{eq:s-combine2-last}) from the proof of Proposition \ref{prop:iden2}, we get the following lower bound for the first term in (\ref{eq:s-combine3})
\begin{align}
     \mathbb{P}\left(s_b - s_1 + \omicron \leq \pi_{t, 1} - \pi_{t, b} < s^0_{b} - s^0_{1}\right) &\geq \mathbb{P}\left(s^0_{b} - \beta + \omicron \leq \pi_{t, 1} - \pi_{t, b}< s^0_{b} \right) \geq \frac{(\beta-\omicron)^2}{(\overline{C}-\underline{C})^2} 
\end{align}
and obtain the desired result.
\begin{multline}
     \mathbb{P}\left(\ell\left(\mathbf{s}, \upsilon_{t}(\boldsymbol{\pi}_t), \boldsymbol{\pi}_t\right) \geq \omicron \Big | \upsilon_t(\boldsymbol{\pi}_t) = \argmax_{a \in \mathcal{A}} \left( s^0_a + \pi_{t, a} \right)\right) \\ \geq \frac{(\beta-\omicron)^2}{(\overline{C}-\underline{C})^2} \left(1 - \frac{\gamma + \beta - d}{\overline{C} - \underline{C}} \right) \left(1 - \frac{\gamma + \omega}{\overline{C} - \underline{C}} \right) \left( \frac{\gamma}{\overline{C} - \underline{C}}  \right)^{n-2} 
\end{multline}
\Halmos \endproof

\proof{\textbf{Proof of Proposition \ref{prop:iden4}.}}
We derive the desired lower bound by conditioning on the case when the imperfect-knowledge agent selects the true maximizer arm at time $t \in \mathcal{T}$.
\begin{align}
    &\mathbb{P}\left( \ell\left(\mathbf{s}, \upsilon_{t}(\boldsymbol{\pi}_t), \boldsymbol{\pi}_t\right) \geq \omicron \right) \nonumber  \allowdisplaybreaks  \\
    &\geq \mathbb{P}\left(\ell\left(\mathbf{s}, \upsilon_{t}(\boldsymbol{\pi}_t), \boldsymbol{\pi}_t\right) \geq \omicron \Big | \upsilon_t(\boldsymbol{\pi}_t) = \argmax_{a \in \mathcal{A}} (s^0_a + \pi_{t, a}) \right) \mathbb{P}\left(\upsilon_t(\boldsymbol{\pi}_t) = \argmax_{a \in \mathcal{A}} (s^0_a + \pi_{t, a} )\right)  \allowdisplaybreaks  \\
    &= \mathbb{P}\left(\ell\left(\mathbf{s}, \upsilon_{t}(\boldsymbol{\pi}_t), \boldsymbol{\pi}_t\right) \geq \omicron \Big | \upsilon_t(\boldsymbol{\pi}_t) = \argmax_{a \in \mathcal{A}} ( s^0_a + \pi_{t, a} ) \right) (1 - p_t)  \allowdisplaybreaks  \\
    &\geq \mathbb{P}\left(\ell\left(\mathbf{s}, \upsilon_{t}(\boldsymbol{\pi}_t), \boldsymbol{\pi}_t\right) \geq \omicron \Big | \upsilon_t(\boldsymbol{\pi}_t) = \argmax_{a \in \mathcal{A}} (s^0_a + \pi_{t, a}) \right) \left(1 - k \frac{\sqrt{\log 2t}}{\sqrt{t}} \right)
    \label{eq:iden1}
\end{align}
where the last inequality follows by Assumption \ref{assm:agent}. Then, we observe that the following three conditions are mutually exclusive events
\begin{enumerate}
    \item[\textit{i.}] $K^0 \cap K = \emptyset$
    \item[\textit{ii.}] $K^0 \cap K \neq \emptyset$ and $b \notin K^0 \cap K$ 
    \item[\textit{iii.}] $K^0 \cap K \neq \emptyset$ and $b \in K^0 \cap K$ 
\end{enumerate}
which allows us to combine the results of Propositions \ref{prop:iden1}-\ref{prop:iden3} and obtain
\begin{align}
     (\ref{eq:iden1}) &= \sum_{j \in \{i, ii, iii\}} \mathbb{P}\left(\ell\left(\mathbf{s}, \upsilon_{t}(\boldsymbol{\pi}_t), \boldsymbol{\pi}_t\right) \geq \omicron, \ j \Big | \upsilon_t(\boldsymbol{\pi}_t) = \argmax_{a \in \mathcal{A}} (s^0_a + \pi_{t, a}) \right) \left(1 - k \frac{\sqrt{\log 2t}}{\sqrt{t}} \right) \allowdisplaybreaks  \\
     &\geq \alpha (\beta - \omicron)^2  \left(1 - k \frac{\sqrt{\log 2t}}{\sqrt{t}} \right)
\end{align}
for some constant $\alpha > 0$. 
\Halmos \endproof

\proof{\textbf{Proof of Proposition \ref{prop:concen1}.}}
We derive the given concentration bound by using the bounded differences inequality (i.e., McDiarmid’s inequality) \citep{boucheron2013concentration}. So, we begin by showing that the loss function $L^{\Lambda(\widetilde{k}, t)}\left(\mathbf{s}, \Upsilon_t(\boldsymbol{\Pi}_t), \boldsymbol{\Pi}_t\right)$ has the \emph{bounded differences property}. 

We first note that the single-step loss function $\ell\left(\mathbf{s}, \upsilon_\tau(\boldsymbol{\pi}_\tau), \boldsymbol{\pi}_\tau\right) = \max_{a \in \mathcal{A}} \left( s_a + \pi_{\tau, a} - s_{\upsilon_\tau(\boldsymbol{\pi}_\tau)} - \pi_{\tau, \upsilon_\tau(\boldsymbol{\pi}_\tau)} \right)$ is bounded from below and above as
\begin{align}
   \left(2R_{\min} - R_{\max} \right)  - \left(2R_{\max} + \gamma - R_{\min}\right) &\leq \ell\left(\mathbf{s}, \upsilon_\tau(\boldsymbol{\pi}_\tau), \boldsymbol{\pi}_\tau\right)
    \leq \left(2R_{\max} + \gamma - R_{\min}\right) - \left(2R_{\min} - R_{\max} \right)  \allowdisplaybreaks  \\ 
    3R_{\min} - 3 R_{\max} - \gamma &\leq \ell\left(\mathbf{s}, \upsilon_\tau(\boldsymbol{\pi}_\tau), \boldsymbol{\pi}_\tau\right)
    \leq 3R_{\max} - 3 R_{\min} + \gamma \label{eq:boundsonell}
\end{align}
since $s_a \in \mathcal{S} = \left[R_{\min} - R_{\max}, R_{\max} - R_{\min}\right]$ and $\pi_{\tau,a} \in \mathcal{C} = \left[R_{\min}, R_{\max} + \gamma \right]$ for all $a \in \mathcal{A}, \tau \in \mathcal{T}$ by definition.

Now, recall that the sequence of incentives $\boldsymbol{\Pi}_t = \{\boldsymbol{\pi}_1, \ldots, \boldsymbol{\pi}_{\widetilde{\tau}}, \ldots, \boldsymbol{\pi}_{t-1}\} \in \mathcal{C}^{n \times (t-1)}$ includes $(t-1)$ vectors of dimension $n = |\mathcal{A}|$ by definition. We define $\boldsymbol{\pi}'_{\widetilde{\tau}} = \{\pi_{\widetilde{\tau},1}, \ldots, \pi'_{\widetilde{\tau},p}, \ldots, \pi_{\widetilde{\tau},n}\}$ for a particular  $\widetilde{\tau} \in \{1, \ldots, t-1\}$ such that it has the same components with $\boldsymbol{\pi}_{\widetilde{\tau}}$ except the value at index $p$. Then, for $\boldsymbol{\Pi}'_t = \{\boldsymbol{\pi}_1, \ldots, \boldsymbol{\pi}'_{\widetilde{\tau}}, \ldots, \boldsymbol{\pi}_{t-1}\}$, we have 
\begin{align} 
    \left| L^{\Lambda(\widetilde{k}, t)}\left(\mathbf{s}, \Upsilon_t(\boldsymbol{\Pi}_t), \boldsymbol{\Pi}_t\right) -  L^{\Lambda(\widetilde{k}, t)}\left(\mathbf{s}, \Upsilon_t(\boldsymbol{\Pi}'_t), \boldsymbol{\Pi}'_t\right) \right| &= \left|\sum_{\tau\in \Lambda(\widetilde{k}, t)} \ell\left(\mathbf{s}, \upsilon_\tau(\boldsymbol{\pi}_\tau), \boldsymbol{\pi}_\tau \right) - \ell\left(\mathbf{s}, \upsilon_\tau(\boldsymbol{\pi}'_\tau), \boldsymbol{\pi}'_\tau \right) \right|  \allowdisplaybreaks  \\
    &= \left| \ell\left(\mathbf{s}, \upsilon_{\widetilde{\tau}}(\boldsymbol{\pi}_{\widetilde{\tau}}), \boldsymbol{\pi}_{\widetilde{\tau}}\right) - \ell\left(\mathbf{s}, \upsilon_{\widetilde{\tau}}(\boldsymbol{\pi}'_{\widetilde{\tau}}), \boldsymbol{\pi}'_{\widetilde{\tau}}\right) \right|  \allowdisplaybreaks  \\
    &\leq 6R_{\max} - 6R_{\min} + 2\gamma  \label{eq:bdddiff}
\end{align}
where the last result follows by (\ref{eq:boundsonell}). Because this result holds for any $\widetilde{\tau} \in \{1, \ldots, t-1\}$ and any index $p \in \{1, \ldots, n\}$, it shows that the bounded differences property holds for $L^{\Lambda(\widetilde{k}, t)}\left(\mathbf{s}, \Upsilon_t(\boldsymbol{\Pi}_t), \boldsymbol{\Pi}_t\right)$. Therefore, we can directly use the bounded differences inequality \citep{boucheron2013concentration} to obtain 
\begin{multline}
    \mathbb{P} \left(L^{\Lambda(\widetilde{k}, t)}\left(\mathbf{s}, \Upsilon_t(\boldsymbol{\Pi}_t), \boldsymbol{\Pi}_t\right) - \mathbb{E} L^{\Lambda(\widetilde{k}, t)}\left(\mathbf{s}, \Upsilon_t(\boldsymbol{\Pi}_t), \boldsymbol{\Pi}_t\right) \geq \nu \right)  \\ \leq \exp \left(- \frac{2\nu^2}{(\eta(\widetilde{k}, t)-1) n \left(6R_{\max} - 6R_{\min} + 2\gamma\right)^2 } \right)
\end{multline}
for any $\nu > 0$. Because the bounded differences property $(\ref{eq:bdddiff})$ is symmetric, $L^{\Lambda(\widetilde{k}, t)}\left(\mathbf{s}, \Upsilon_t(\boldsymbol{\Pi}_t), \boldsymbol{\Pi}_t\right)$ also satisfies the lower-tail inequality 
\begin{multline}
    \mathbb{P} \left(L^{\Lambda(\widetilde{k}, t)}\left(\mathbf{s}, \Upsilon_t(\boldsymbol{\Pi}_t), \boldsymbol{\Pi}_t\right) - \mathbb{E} L^{\Lambda(\widetilde{k}, t)}\left(\mathbf{s}, \Upsilon_t(\boldsymbol{\Pi}_t), \boldsymbol{\Pi}_t\right) \leq - \nu  \right) \\ \leq \exp \left(- \frac{2\nu^2}{(\eta(\widetilde{k}, t)-1) n \left(6R_{\max} - 6R_{\min} + 2\gamma\right)^2 } \right)
\end{multline}
for any $\nu > 0$. Lastly, combining the last two inequalities gives us desired the concentration bound and completes our proof.
\Halmos \endproof

\proof{\textbf{Proof of Proposition \ref{prop:concen2}.}}
First, we start by recalling that the finite subcover $\{\mathcal{B}(\mathbf{s}^j, d) : \mathbf{s}^j \in \mathcal{F}\}_{j = 1}^q$ of a collection of open balls covering $\mathcal{F}$ for finite $q > 0$ and $d < \beta$. Now, we also define the vector $\overline{\mathbf{s}}^j_t = \argsup_{\mathbf{s} \in \mathcal{B}(\mathbf{s}^j, d)} \left| L^{\Lambda(\widetilde{k}, t)}\left(\mathbf{s}, \Upsilon_t(\boldsymbol{\Pi}_t), \boldsymbol{\Pi}_t\right) - \mathbb{E} L^{\Lambda(\widetilde{k}, t)}\left(\mathbf{s}, \Upsilon_t(\boldsymbol{\Pi}_t), \boldsymbol{\Pi}_t\right) \right|$. Then, we have
\begin{align}
    &\sup_{\mathbf{s} \in \mathcal{F}} \left| L^{\Lambda(\widetilde{k}, t)}\left(\mathbf{s}, \Upsilon_t(\boldsymbol{\Pi}_t), \boldsymbol{\Pi}_t\right) - \mathbb{E} L^{\Lambda(\widetilde{k}, t)}\left(\mathbf{s}, \Upsilon_t(\boldsymbol{\Pi}_t), \boldsymbol{\Pi}_t\right)  \right| \nonumber  \allowdisplaybreaks  \\ 
    &\leq \max_{j \in [q]} \sup_{\mathbf{s} \in \mathcal{B}(\mathbf{s}^j, d)}\left| L^{\Lambda(\widetilde{k}, t)}\left(\mathbf{s}, \Upsilon_t(\boldsymbol{\Pi}_t), \boldsymbol{\Pi}_t\right) - \mathbb{E} L^{\Lambda(\widetilde{k}, t)}\left(\mathbf{s}, \Upsilon_t(\boldsymbol{\Pi}_t), \boldsymbol{\Pi}_t\right)  \right|   \allowdisplaybreaks  \\
    &= \max_{j \in [q]}  \left| L^{\Lambda(\widetilde{k}, t)}\left(\overline{\mathbf{s}}^j_t, \Upsilon_t(\boldsymbol{\Pi}_t), \boldsymbol{\Pi}_t\right) - \mathbb{E} L^{\Lambda(\widetilde{k}, t)}\left(\overline{\mathbf{s}}^j_t, \Upsilon_t(\boldsymbol{\Pi}_t), \boldsymbol{\Pi}_t\right) \right|  \label{eq:conc1-s}
\end{align}
where $[q] = \{1, \ldots, q\}$, and the first inequality follows because $\mathcal{F} \subseteq \bigcup_{j = 1}^q \mathcal{B}(\mathbf{s}^j, d)$ by construction. We then follow by
 \begin{align}
    &\mathbb{P}\left( \sup_{\mathbf{s} \in \mathcal{F}} \left| L^{\Lambda(\widetilde{k}, t)}\left(\mathbf{s}, \Upsilon_t(\boldsymbol{\Pi}_t), \boldsymbol{\Pi}_t\right) - \mathbb{E} L^{\Lambda(\widetilde{k}, t)}\left(\mathbf{s}, \Upsilon_t(\boldsymbol{\Pi}_t), \boldsymbol{\Pi}_t\right) \right| \geq \nu \right)  \nonumber  \allowdisplaybreaks  \\
    &\leq \mathbb{P}\left(\max_{j \in [q]}  \left| L^{\Lambda(\widetilde{k}, t)}\left(\overline{\mathbf{s}}^j_t, \Upsilon_t(\boldsymbol{\Pi}_t), \boldsymbol{\Pi}_t\right) - \mathbb{E} L^{\Lambda(\widetilde{k}, t)}\left(\overline{\mathbf{s}}^j_t, \Upsilon_t(\boldsymbol{\Pi}_t), \boldsymbol{\Pi}_t\right) \right| \geq \nu \right)  \allowdisplaybreaks  \\
    &\leq \mathbb{P}\bigg(\bigcup_{j \in [q]} \left| L^{\Lambda(\widetilde{k}, t)}\left(\overline{\mathbf{s}}^j_t, \Upsilon_t(\boldsymbol{\Pi}_t), \boldsymbol{\Pi}_t\right) - \mathbb{E} L^{\Lambda(\widetilde{k}, t)}\left(\overline{\mathbf{s}}^j_t, \Upsilon_t(\boldsymbol{\Pi}_t), \boldsymbol{\Pi}_t\right) \right| \geq \nu \bigg)  \allowdisplaybreaks  \\
    &\leq \sum_{j \in [q]} \mathbb{P}\left( \left| L^{\Lambda(\widetilde{k}, t)}\left(\overline{\mathbf{s}}^j_t, \Upsilon_t(\boldsymbol{\Pi}_t), \boldsymbol{\Pi}_t\right) - \mathbb{E} L^{\Lambda(\widetilde{k}, t)}\left(\overline{\mathbf{s}}^j_t, \Upsilon_t(\boldsymbol{\Pi}_t), \boldsymbol{\Pi}_t\right) \right| \geq \nu \right)  \label{eq:byunion} \allowdisplaybreaks  \\
    &\leq \sum_{j \in [q]} 2\exp\left(-\frac{2 \nu^2}{(\eta(\widetilde{k}, t)-1)n(6R_{\max} - 6R_{\min} + 2\gamma)^2}\right) \label{eq:byProposition5-S} \allowdisplaybreaks  \\
    &= 2 q \exp\left(-\frac{2 \nu^2}{(\eta(\widetilde{k}, t)-1)n(6R_{\max} - 6R_{\min} + 2\gamma)^2}\right) \label{eq:substituteq}
\end{align}
where (\ref{eq:byunion}) follows by Boole's inequality (i.e., union bound) and (\ref{eq:byProposition5-S}) follows by Proposition \ref{prop:concen1}. Note that Proposition \ref{prop:concen1} holds for any vector $\mathbf{s} \in \mathcal{S}$, and hence, it also holds for $\overline{\mathbf{s}}^j_t$. 

Lastly, it remains to provide an upper bound for the covering number $q$. We compute this bound by using the volume ratios and definition of the set $\mathcal{S} = [R_{\min} - R_{\max}, R_{\max} - R_{\min}]$.
\begin{align}
    q = \mathcal{N}(d, \mathcal{F}, \|\cdot\|) \leq \frac{\mathrm{vol} (\mathcal{F})}{\mathrm{vol} (\mathcal{B}(\mathbf{s}^j, d))} \leq  \frac{\mathrm{vol} (\mathcal{S}^n)}{\mathrm{vol} (\mathcal{B}(\mathbf{s}^j, d))} \leq \frac{(R_{\max} - R_{\min})^n}{d^n} \label{eq:boundonq}
\end{align}
Suppose that we have $d = \sqrt[n]{\beta}$. Then, combining (\ref{eq:substituteq}) and (\ref{eq:boundonq}) gives us the desired result.
\begin{align}
    &\mathbb{P}\left( \sup_{\mathbf{s} \in \mathcal{F}} \left| L^{\Lambda(\widetilde{k}, t)}\left(\mathbf{s}, \Upsilon_t(\boldsymbol{\Pi}_t), \boldsymbol{\Pi}_t\right) - \mathbb{E} L^{\Lambda(\widetilde{k}, t)}\left(\mathbf{s}, \Upsilon_t(\boldsymbol{\Pi}_t), \boldsymbol{\Pi}_t\right) \right| \geq \nu \right) \nonumber  \allowdisplaybreaks  \\  
    &\leq 2 \frac{(R_{\max} - R_{\min})^n}{\beta} \exp\left(-\frac{2 \nu^2}{(\eta(\widetilde{k}, t)-1)n(6R_{\max} - 6R_{\min} + 2\gamma)^2}\right)  \allowdisplaybreaks  \\
    &= 2\exp \left(-\frac{2 \nu^2}{(\eta(\widetilde{k}, t)-1)n(6R_{\max} - 6R_{\min} + 2\gamma)^2} - \log \beta + n \log (R_{\max} - R_{\min})\right)
\end{align} 
\Halmos \endproof

\proof{\textbf{Proof of Lemma \ref{lem:concen1}.}}
The given lower bound can be derived by considering the time steps $\tau \in \Lambda(\widetilde{k}, t)$ where the agent selects a utility-maximizer arm $\upsilon_\tau(\boldsymbol{\pi}_\tau) = \argmax_{a \in \mathcal{A}} s^0_a + \pi_{\tau,a}$.  Further, recall that the set $\Lambda(\widetilde{k}, t)$ consists of random time points which makes $\eta(\widetilde{k}, t)$ a random variable. Thus, we start by computing a lower bound for the conditional expected loss given the set $\Lambda(\widetilde{k}, t)$. 
\begin{align}
   &\mathbb{E} \left[ L^{\Lambda(\widetilde{k}, t)}\left(\mathbf{s}^{\mathcal{F}}_t, \Upsilon_t(\boldsymbol{\Pi}_t), \boldsymbol{\Pi}_t\right) \Big| \Lambda(\widetilde{k}, t) \right] \allowdisplaybreaks  \\
   &=  \mathbb{E} \bigg[\sum_{\tau\in \Lambda(\widetilde{k}, t)} \ell\left(\mathbf{s}^\mathcal{F}_t, \upsilon_\tau(\boldsymbol{\pi}_\tau), \boldsymbol{\pi}_\tau \right) \Big| \Lambda(\widetilde{k}, t)\bigg]  \allowdisplaybreaks  \\
   &\geq \mathbb{E} \bigg[\sum_{\tau\in \Lambda(\widetilde{k}, t)} \ell\left(\mathbf{s}^\mathcal{F}_t, \upsilon_\tau(\boldsymbol{\pi}_\tau), \boldsymbol{\pi}_\tau \right) \cdot \mathbf{1}\left\{\upsilon_\tau(\boldsymbol{\pi}_\tau) = \argmax_{a \in \mathcal{A}} s^0_a + \pi_{\tau,a}\right\} \Big| \Lambda(\widetilde{k}, t) \bigg]  \allowdisplaybreaks  \\
   &\geq \sum_{\tau\in \Lambda(\widetilde{k}, t)} \omicron \mathbb{P} \left(\ell\left(\mathbf{s}^\mathcal{F}_t, \upsilon_\tau(\boldsymbol{\pi}_\tau), \boldsymbol{\pi}_\tau \right) \geq \omicron \right)  \left( 1 - p_\tau \right)   \label{eq:bycondition}\allowdisplaybreaks  \\
   &\geq \alpha \omicron (\beta - \omicron)^2 \sum_{\tau\in \Lambda(\widetilde{k}, t)} \left(1 - k \frac{\sqrt{\log 2\tau}}{\sqrt{\tau}}\right)^2   \label{eq:byProp4-S}  \allowdisplaybreaks  \\
   &\geq  \alpha \omicron (\beta - \omicron)^2 \sum_{\tau\in \Lambda(\widetilde{k}, t)} \left(1 - k \sqrt{\log 2\widetilde{k}}/ \sqrt{\widetilde{k}}\right)^2  \allowdisplaybreaks  \\
   &= \alpha \left(1 - k \sqrt{\log 2\widetilde{k}}/ \sqrt{\widetilde{k}}\right)^2 \omicron (\beta - \omicron)^2 \eta(\widetilde{k}, t)
\end{align}
where (\ref{eq:bycondition}) follows by considering whether the single-step loss function $\ell\left(\mathbf{s}^\mathcal{F}_t, \upsilon_\tau(\boldsymbol{\pi}_\tau), \boldsymbol{\pi}_\tau \right)$ is zero or strictly positive, and (\ref{eq:byProp4-S}) follows by Assumption \ref{assm:agent} and Proposition \ref{prop:iden4} for any $\omicron \in (0, \beta)$. 

Next, we take the expectation of both sides of the last inequality and get
\begin{multline}
    \mathbb{E} \left[ \mathbb{E} \left[ L^{\Lambda(\widetilde{k}, t)}\left(\mathbf{s}^{\mathcal{F}}_t, \Upsilon_t(\boldsymbol{\Pi}_t), \boldsymbol{\Pi}_t\right) \Big| \Lambda(\widetilde{k}, t) \right] \right] = \mathbb{E} L^{\Lambda(\widetilde{k}, t)}\left(\mathbf{s}^{\mathcal{F}}_t, \Upsilon_t(\boldsymbol{\Pi}_t), \boldsymbol{\Pi}_t\right)
    \\ \geq \alpha \left(1 - k \sqrt{\log 2\widetilde{k}}/ \sqrt{\widetilde{k}}\right)^2 \omicron (\beta - \omicron)^2 \mathbb{E}\eta(\widetilde{k}, t) \allowdisplaybreaks
\end{multline}

Lastly, we substitute $\omicron = \beta/3$ to maximize the lower bound that we have in the last line above and conclude our proof by
\begin{align}
    \mathbb{E} L^{\Lambda(\widetilde{k}, t)}\left(\mathbf{s}^{\mathcal{F}}_t, \Upsilon_t(\boldsymbol{\Pi}_t), \boldsymbol{\Pi}_t\right) &\geq \frac{4\alpha \left(1 - k \sqrt{\log 2\widetilde{k}}/ \sqrt{\widetilde{k}}\right)^2}{27} \beta^3 \mathbb{E}\eta(\widetilde{k}, t)  
\end{align}
\Halmos \endproof

\proof{\textbf{Proof of Lemma \ref{lem:concen2}.}}
First, we recall that the single-step loss $\ell\left(\mathbf{s}^0, \upsilon_\tau(\boldsymbol{\pi}_\tau), \boldsymbol{\pi}_\tau\right)$ becomes 0 when the agent selects the true maximizer arm (i.e., $\upsilon_\tau(\boldsymbol{\pi}_\tau) = \argmax_{a \in \mathcal{A}} s^0_a + \pi_{\tau,a}$). Thus, in this proof, it suffices to consider only the time steps where the agent selects an arbitrary non-maximizer arm $\upsilon_\tau(\boldsymbol{\pi}_\tau) \in \mathcal{A}$. Because $\upsilon_\tau(\boldsymbol{\pi}_\tau)$ is not the true maximizer arm, we have $s^0_{\upsilon_\tau(\boldsymbol{\pi}_\tau)} + \pi_{\tau,\upsilon_\tau(\boldsymbol{\pi}_\tau)} < s^0_a  + \pi_{\tau,a}$ for some $a \in \mathcal{A} \setminus \{\upsilon_\tau(\boldsymbol{\pi}_\tau)\}$. Further, we know that by definition 
\begin{align}
    \ell\left(\mathbf{s}^0, \upsilon_\tau(\boldsymbol{\pi}_\tau), \boldsymbol{\pi}_\tau\right) = \max_{a \in \mathcal{A}} \left(s^0_a + \pi_{\tau, a}\right) -  s^0_{\upsilon_\tau(\boldsymbol{\pi}_\tau)} - \pi_{\tau, \upsilon_\tau(\boldsymbol{\pi}_\tau)} \leq 3 (R_{\max} - R_{\min}) + \gamma  
\end{align}
since $s_a \in \mathcal{S} = [R_{\min} - R_{\max}, R_{\max}- R_{\min}]$ and $\pi_{\tau,a} \in \mathcal{C} = [R_{\min}, R_{\max} + \gamma]$ for all $a, \tau$. Then, we have 
\begin{align}
    \mathbb{E}L\left(\mathbf{s}^0, \Upsilon_t(\boldsymbol{\Pi}_t), \boldsymbol{\Pi}_t\right)  &= \sum_{\tau =1}^{t-1} \mathbb{E} \ell\left(\mathbf{s}^0, \upsilon_\tau(\boldsymbol{\pi}_\tau), \boldsymbol{\pi}_\tau\right)  \allowdisplaybreaks  \\
    &= \sum_{\tau =1}^{t-1} \mathbb{E} \ell\left(\mathbf{s}^0, \upsilon_\tau(\boldsymbol{\pi}_\tau), \boldsymbol{\pi}_\tau\right) \mathbf{1}\Big\{\upsilon_\tau(\boldsymbol{\pi}_\tau) \neq \argmax_{a \in \mathcal{A}} s^0_a + \pi_{\tau,a}\Big\} \allowdisplaybreaks  \\
    &\leq \left(3 (R_{\max} - R_{\min}) + \gamma\right) \sum_{\tau =1}^{t-1} p_\tau  \allowdisplaybreaks  \\
    &\leq \left(3 (R_{\max} - R_{\min}) + \gamma\right) \sum_{\tau =1}^{t-1} k \frac{\sqrt{\log 2\tau}}{\sqrt{\tau}} \label{eq:byAss1-S}  \allowdisplaybreaks  \\
    &\leq  k \left(3 (R_{\max} - R_{\min}) + \gamma\right) \left( \sqrt{\log 2}  + \int_{\tau = 1}^{t-1} \frac{\sqrt{\log 2\tau}}{\sqrt{\tau}} d\tau \right) \allowdisplaybreaks  \\
    &= k \left(3 (R_{\max} - R_{\min}) + \gamma\right) \left( \sqrt{\log 2} + \sqrt{2} \sqrt{(2t - 2)
    \log (2t-2)} \right) \allowdisplaybreaks  \\
    &\leq k \left(3 (R_{\max} - R_{\min}) + \gamma\right) \left( \sqrt{\log 2} + 2\sqrt{t\log (2t)} \right) \allowdisplaybreaks  \\
    &\leq 3 k \left(3 (R_{\max} - R_{\min}) + \gamma\right) \sqrt{t\log (2t)}  \allowdisplaybreaks  
\end{align}
where (\ref{eq:byAss1-S}) follows by Assumption \ref{assm:agent}. 
\Halmos \endproof

\proof{\textbf{Proof of Lemma \ref{lem:concen3}.}}
Similar to our argument in the proof of Lemma \ref{lem:concen2}, we note that the single-step loss $\ell\left(\mathbf{s}^0, \upsilon_\tau(\boldsymbol{\pi}_\tau), \boldsymbol{\pi}_\tau\right)$ becomes 0 when the agent selects the true utility-maximizer arm, $\upsilon_\tau(\boldsymbol{\pi}_\tau) = \argmax_{a \in \mathcal{A}} s^0_a + \pi_{\tau,a}$. Therefore, we only need to bound the single-step loss at the time steps where the agent selects an arbitrary non-maximizer arm $\upsilon_\tau(\boldsymbol{\pi}_\tau) \in \mathcal{A}$.
\begin{align}
    L\left(\mathbf{s}^0, \Upsilon_t(\boldsymbol{\Pi}_t), \boldsymbol{\Pi}_t\right) &= \sum_{\tau =1}^{t-1} \ell\left(\mathbf{s}^0, \upsilon_\tau(\boldsymbol{\pi}_\tau), \boldsymbol{\pi}_\tau\right) \mathbf{1}\Big\{\upsilon_\tau(\boldsymbol{\pi}_\tau) \neq \argmax_{a \in \mathcal{A}} s^0_a + \pi_{\tau,a}\Big\}  \allowdisplaybreaks  \\
    &=  \sum_{\tau =1}^{t-1} \max_{a \in \mathcal{A}} \left\{s^0_a + \pi_{\tau, a} -  s^0_{\upsilon_\tau(\boldsymbol{\pi}_\tau)} - \pi_{\tau, \upsilon_\tau(\boldsymbol{\pi}_\tau)} \right\}  \mathbf{1}\Big\{\upsilon_\tau(\boldsymbol{\pi}_\tau) \neq \argmax_{a \in \mathcal{A}} s^0_a + \pi_{\tau,a}\Big\}  \allowdisplaybreaks  
\end{align}   
We next compute the concentration inequality for the sum of identity functions in the last expression above which are independent Bernoulli variables.  
\begin{align}
    &\mathbb{P}\left(\sum_{\tau =1}^{t-1} \mathbf{1}\Big\{\upsilon_\tau(\boldsymbol{\pi}_\tau) \neq \argmax_{a \in \mathcal{A}} s^0_a + \pi_{\tau,a}\Big\} - \mathbb{E} \sum_{\tau =1}^{t-1} \mathbf{1}\Big\{\upsilon_\tau(\boldsymbol{\pi}_\tau) \neq \argmax_{a \in \mathcal{A}} s^0_a + \pi_{\tau,a}\Big\}  \geq \overline{\nu} \right)  \allowdisplaybreaks  \\
    &\leq \exp\left(-\frac{2\overline{\nu}^2}{t-1}\right)
\end{align}
where the last inequality follows by Hoeffding's Inequality \citep{boucheron2013concentration} for any $\overline{\nu} > 0$. Here, we note that $\max_{a \in \mathcal{A}} \left\{s^0_a + \pi_{\tau, a} -  s^0_{\upsilon_\tau(\boldsymbol{\pi}_\tau)} - \pi_{\tau, \upsilon_\tau(\boldsymbol{\pi}_\tau)} \right\}$ for all $\tau \leq t-1$ is a constant value for a given sequence of incentives $\boldsymbol{\Pi}_t = \{\boldsymbol{\pi}_1, \boldsymbol{\pi}_2, \ldots, \boldsymbol{\pi}_{t-1}\}$. Further, we know that $\max_{a \in \mathcal{A}} \left\{s^0_a + \pi_{\tau, a} -  s^0_{\upsilon_\tau(\boldsymbol{\pi}_\tau)} - \pi_{\tau, \upsilon_\tau(\boldsymbol{\pi}_\tau)} \right\} \leq 3R_{\max} - 3R_{\min} + \gamma$ since $s_a \in \mathcal{S} = [R_{\min} - R_{\max}, R_{\max}- R_{\min}]$ and $\pi_{\tau,a} \in \mathcal{C} = [R_{\min}, R_{\max} + \gamma]$ for all $a \in \mathcal{A}, \tau \in \mathcal{T}$. Then, the last result implies that
\begin{align}
    \mathbb{P}\left(L\left(\mathbf{s}^0, \Upsilon_t(\boldsymbol{\Pi}_t), \boldsymbol{\Pi}_t\right) - \mathbb{E}  L\left(\mathbf{s}^0, \Upsilon_t(\boldsymbol{\Pi}_t), \boldsymbol{\Pi}_t\right) \geq \left(3R_{\max} - 3R_{\min} + \gamma\right)\overline{\nu} \right)  \leq \exp\left(-\frac{2\overline{\nu}^2}{t-1}\right)   
\end{align}
Last, replacing $ \nu = \left(3R_{\max} - 3R_{\min} + \gamma\right)\overline{\nu}$ for any $\overline{\nu} > 0$, we obtain the concentration inequality for $L\left(\mathbf{s}^0, \Upsilon_t(\boldsymbol{\Pi}_t), \boldsymbol{\Pi}_t\right)$ as
\begin{align}
    \mathbb{P}\left( L\left(\mathbf{s}^0, \Upsilon_t(\boldsymbol{\Pi}_t), \boldsymbol{\Pi}_t\right) - \mathbb{E}  L\left(\mathbf{s}^0, \Upsilon_t(\boldsymbol{\Pi}_t), \boldsymbol{\Pi}_t\right) \geq \nu \right) &\leq \exp\left(-\frac{2\nu^2}{(t-1)\left(3R_{\max} - 3R_{\min} + \gamma\right)^2}\right)   
\end{align}
for any $\nu > 0$. 
\Halmos \endproof

\proof{\textbf{Proof of Theorem \ref{thm:concen}.}}
For notational convenience, we let $\mathbf{s}^\mathcal{F}_t = \arginf_{\mathbf{s} \in \mathcal{F}} L(\mathbf{s}, \Upsilon_t(\boldsymbol{\Pi}_t), \boldsymbol{\Pi}_t)$. Then, we begin by considering any constant $\nu > 0$ such that $3\nu < \mathbb{E}L^{\Lambda(\widetilde{k}, t)}(\mathbf{s}^\mathcal{F}_t, \Upsilon_t(\boldsymbol{\Pi}_t), \boldsymbol{\Pi}_t) - \mathbb{E} L(\mathbf{s}^0, \Upsilon_t(\boldsymbol{\Pi}_t), \boldsymbol{\Pi}_t)$ where $L^{\Lambda(\widetilde{k}, t)}(\mathbf{s}^\mathcal{F}_t, \Upsilon_t(\boldsymbol{\Pi}_t), \boldsymbol{\Pi}_t)$ is as introduced in (\ref{eq:l-stoc-lambda}). 

If $L(\mathbf{s}^0, \Upsilon_t(\boldsymbol{\Pi}_t), \boldsymbol{\Pi}_t) < \mathbb{E} L(\mathbf{s}^0, \Upsilon_t(\boldsymbol{\Pi}_t), \boldsymbol{\Pi}_t) + \nu$ and $L^{\Lambda(\widetilde{k}, t)}(\mathbf{s}^\mathcal{F}_t, \Upsilon_t(\boldsymbol{\Pi}_t), \boldsymbol{\Pi}_t) > \mathbb{E}L^{\Lambda(\widetilde{k}, t)}(\mathbf{s}^\mathcal{F}_t, \Upsilon_t(\boldsymbol{\Pi}_t), \boldsymbol{\Pi}_t) - \nu$ for the $\nu$ being considered, then it follows that $L(\mathbf{s}^0, \Upsilon_t(\boldsymbol{\Pi}_t), \boldsymbol{\Pi}_t) < L^{\Lambda(\widetilde{k}, t)}(\mathbf{s}^\mathcal{F}_t, \Upsilon_t(\boldsymbol{\Pi}_t), \boldsymbol{\Pi}_t) $. 

Now, taking the contrapositive of the last `if' statement, this means that if we have $L^{\Lambda(\widetilde{k}, t)}(\mathbf{s}^\mathcal{F}_t, \Upsilon_t(\boldsymbol{\Pi}_t), \boldsymbol{\Pi}_t) \leq L\left(\mathbf{s}^0, \Upsilon_t(\boldsymbol{\Pi}_t), \boldsymbol{\Pi}_t\right)$, then either $L(\mathbf{s}^0, \Upsilon_t(\boldsymbol{\Pi}_t), \boldsymbol{\Pi}_t) \geq \mathbb{E} L(\mathbf{s}^0, \Upsilon_t(\boldsymbol{\Pi}_t), \boldsymbol{\Pi}_t) + \nu$ or $L^{\Lambda(\widetilde{k}, t)}(\mathbf{s}^\mathcal{F}_t, \Upsilon_t(\boldsymbol{\Pi}_t), \boldsymbol{\Pi}_t) \leq \mathbb{E}L^{\Lambda(\widetilde{k}, t)}(\mathbf{s}^\mathcal{F}_t, \Upsilon_t(\boldsymbol{\Pi}_t), \boldsymbol{\Pi}_t) - \nu$ will hold for the $\nu$ being considered. 

Further, we know that $L^{\Lambda(\widetilde{k}, t)}\left(\mathbf{s}, \Upsilon_t(\boldsymbol{\Pi}_t), \boldsymbol{\Pi}_t\right) \leq L\left(\mathbf{s}, \Upsilon_t(\boldsymbol{\Pi}_t), \boldsymbol{\Pi}_t\right)$ for any $\mathbf{s} \in \mathcal{S}$ and all $t \in \mathcal{T}$ by definition. Thus, we have 
\begin{align}
    &\mathbb{P}\left(L(\mathbf{s}^\mathcal{F}_t, \Upsilon_t(\boldsymbol{\Pi}_t), \boldsymbol{\Pi}_t) \leq L(\mathbf{s}^0, \Upsilon_t(\boldsymbol{\Pi}_t), \boldsymbol{\Pi}_t)\right) \nonumber \allowdisplaybreaks  \\
    &\leq \mathbb{P}\left(L^{\Lambda(\widetilde{k}, t)}(\mathbf{s}^\mathcal{F}_t, \Upsilon_t(\boldsymbol{\Pi}_t), \boldsymbol{\Pi}_t) \leq L(\mathbf{s}^0, \Upsilon_t(\boldsymbol{\Pi}_t), \boldsymbol{\Pi}_t)\right) \allowdisplaybreaks  \\
    &\leq \mathbb{P}\left(L(\mathbf{s}^0, \Upsilon_t(\boldsymbol{\Pi}_t), \boldsymbol{\Pi}_t) \geq \mathbb{E} L(\mathbf{s}^0, \Upsilon_t(\boldsymbol{\Pi}_t), \boldsymbol{\Pi}_t) + \nu\right) \notag \allowdisplaybreaks \\
    &\hspace{.5cm}+ \mathbb{P}\left(L^{\Lambda(\widetilde{k}, t)}(\mathbf{s}^\mathcal{F}_t, \Upsilon_t(\boldsymbol{\Pi}_t), \boldsymbol{\Pi}_t) \leq \mathbb{E}L^{\Lambda(\widetilde{k}, t)}(\mathbf{s}^\mathcal{F}_t, \Upsilon_t(\boldsymbol{\Pi}_t), \boldsymbol{\Pi}_t) - \nu\right) \allowdisplaybreaks  \\
    &\leq \exp\bigg(-\frac{2\nu^2}{(t-1)\left(3R_{\max} - 3R_{\min} + \gamma\right)^2}\bigg) \notag \allowdisplaybreaks \\
    &\hspace{.5cm}  + \frac{(R_{\max} - R_{\min})^n}{\beta} \exp\bigg(-\frac{2 \nu^2}{(\eta(\widetilde{k}, t)-1)n(6R_{\max} - 6R_{\min} + 2\gamma)^2}\bigg)   \label{eq:concent-results} \allowdisplaybreaks  \\
    &\leq\exp\bigg(-\frac{2\nu^2}{(t-1)n\left(6R_{\max} - 6R_{\min} + 2\gamma\right)^2}\bigg)\notag \allowdisplaybreaks \\
    &\hspace{.5cm}  + \frac{(R_{\max} - R_{\min})^n}{\beta} \exp\bigg(-\frac{2 \nu^2}{(t-1)n(6R_{\max} - 6R_{\min} + 2\gamma)^2}\bigg)   \allowdisplaybreaks  \\
    &=  \exp\bigg(-\frac{2\nu^2}{(t-1)n\left(6R_{\max} - 6R_{\min} + 2\gamma\right)^2}\bigg) \bigg(1 + \frac{(R_{\max} - R_{\min})^n}{\beta}\bigg) \allowdisplaybreaks  \\
    &\leq \exp\bigg(-\frac{2\nu^2}{(t-1)n\left(6R_{\max} - 6R_{\min} + 2\gamma\right)^2}\bigg) \bigg(1 + \frac{(2R_{\max} - 2R_{\min})^n}{\beta}\bigg)
    \intertext{where (\ref{eq:concent-results}) follows by Lemma \ref{lem:concen3} and Proposition \ref{prop:concen2}. Notice that $\beta < 2R_{\max} - 2R_{\min}$ by construction.}
    &\leq 2\exp \left(-\frac{2\nu^2}{(t-1)n(6R_{\max} - 6R_{\min} + 2\gamma)^2}\right) \frac{(2R_{\max} - 2R_{\min})^n}{\beta} \allowdisplaybreaks  \\
    &=  2\exp \left(-\frac{2\nu^2}{(t-1)n(6R_{\max} - 6R_{\min} + 2\gamma)^2} - \log \beta + n \log (2R_{\max} - 2R_{\min})\right) 
    \intertext{for the $\nu$ being considered. Suppose $\nu = (\mathbb{E}L^{\Lambda(\widetilde{k}, t)}(\mathbf{s}^\mathcal{F}_t, \Upsilon_t(\boldsymbol{\Pi}_t), \boldsymbol{\Pi}_t) - \mathbb{E} L(\mathbf{s}^0, \Upsilon_t(\boldsymbol{\Pi}_t), \boldsymbol{\Pi}_t))/4$. Then,}
    &= 2\exp \left(-\frac{2(\mathbb{E}L^{\Lambda(\widetilde{k}, t)}(\mathbf{s}^\mathcal{F}_t, \Upsilon_t(\boldsymbol{\Pi}_t), \boldsymbol{\Pi}_t) - \mathbb{E} L(\mathbf{s}^0, \Upsilon_t(\boldsymbol{\Pi}_t), \boldsymbol{\Pi}_t))^2}{(t-1)16n(6R_{\max} - 6R_{\min} + 2\gamma)^2} - \log \beta + n \log (2R_{\max} - 2R_{\min})\right) \allowdisplaybreaks  \\
    &\leq 2\exp \left(-\frac{2\lambda_t^2}{(t-1)16n(6R_{\max} - 6R_{\min} + 2\gamma)^2} - \log \beta + n \log (2R_{\max} - 2R_{\min})\right)
\end{align}
where the last inequality follows since $\mathbb{E} L^{\Lambda(\widetilde{k}, t)}\left(\mathbf{s}^\mathcal{F}_t, \Upsilon_t(\boldsymbol{\Pi}_t), \boldsymbol{\Pi}_t\right) -  \mathbb{E}L\left(\mathbf{s}^0, \Upsilon_t(\boldsymbol{\Pi}_t), \boldsymbol{\Pi}_t\right) \geq \lambda_t > 0$ by Lemma \ref{lem:concen1} and Lemma \ref{lem:concen2} for $t \in [\widetilde{k}, T]$. 
\Halmos \endproof

\proof{\textbf{Proof of Corollary \ref{cor:concen}.}}
We first recall that the principal's estimator $\widehat{\mathbf{s}}^\mathrm{pr}_t$ (\ref{prblm:estimator}) is defined such that it satisfies $L\left(\widehat{\mathbf{s}}^\mathrm{pr}_t, \Upsilon_t(\boldsymbol{\Pi}_t), \boldsymbol{\Pi}_t\right) \leq L\left(\mathbf{s}^0, \Upsilon_t(\boldsymbol{\Pi}_t), \boldsymbol{\Pi}_t\right)$. Then, we have the following implication 
\begin{align}
    \left\{\|\mathbf{s}^0 - \widehat{\mathbf{s}}^\mathrm{pr}_t \|_\infty > \beta\right\} &\subseteq \left\{\exists \mathbf{s} : \|\mathbf{s}^0 - \mathbf{s}\|_\infty > \beta \text{ and } L\left(\mathbf{s}, \Upsilon_t(\boldsymbol{\Pi}_t), \boldsymbol{\Pi}_t\right) \leq L\left(\mathbf{s}^0, \Upsilon_t(\boldsymbol{\Pi}_t), \boldsymbol{\Pi}_t\right)\right\} \allowdisplaybreaks \\
    &\subseteq \left\{\inf_{\mathbf{s} \in \mathcal{F}} L\left(\mathbf{s}, \Upsilon_t(\boldsymbol{\Pi}_t), \boldsymbol{\Pi}_t\right) \leq L\left(\mathbf{s}^0, \Upsilon_t(\boldsymbol{\Pi}_t), \boldsymbol{\Pi}_t\right) \right\} \allowdisplaybreaks 
\end{align}
which gives us the desired bound as follows.
\begin{align}
    \mathbb{P} \left(\|\mathbf{s}^0 - \widehat{\mathbf{s}}^\mathrm{pr}_t \|_\infty > \beta \right) &\leq \mathbb{P}\bigg(\inf_{\mathbf{s} \in \mathcal{F}} L\left(\mathbf{s}, \Upsilon_t(\boldsymbol{\Pi}_t), \boldsymbol{\Pi}_t\right) \leq L\left(\mathbf{s}^0, \Upsilon_t(\boldsymbol{\Pi}_t), \boldsymbol{\Pi}_t\right)\bigg)  \allowdisplaybreaks  \\
    &\leq 2\exp \left(-\frac{2\lambda_t^2}{(t-1)16n(6R_{\max} - 6R_{\min} + 2\gamma)^2} - \log \beta + n \log (2R_{\max} - 2R_{\min})\right)
\end{align}
where the last inequality follows by Theorem \ref{thm:concen}.
\Halmos \endproof
\subsection{Results in Section \ref{sec:policy}} \label{appendix2}
\proof{\textbf{Proof of Proposition \ref{prop:regret}.}}
We start by recalling the definition of $j^*_t$ in Algorithm \ref{alg:pr} (line \ref{eq:jstar}) which implies 
\begin{align}
     \mathbb{P} \left(j^*_t \neq j^{*,0}\right) &\leq \mathbb{P} \left(\bigcup_{j^*_t \in \mathcal{A}} \widetilde{V}(j^{*,0}, \widehat{\mathbf{s}}^\mathrm{pr}_t; \widehat{\boldsymbol{\theta}}_t) < \widetilde{V}(j^*_t, \widehat{\mathbf{s}}^\mathrm{pr}_t; \widehat{\boldsymbol{\theta}}_t) \right) \allowdisplaybreaks \\
    &\leq \sum_{j^*_t \in \mathcal{A}} \mathbb{P} \left(\widetilde{V}(j^{*,0}, \widehat{\mathbf{s}}^\mathrm{pr}_t; \widehat{\boldsymbol{\theta}}_t) < \widetilde{V}(j^*_t, \widehat{\mathbf{s}}^\mathrm{pr}_t; \widehat{\boldsymbol{\theta}}_t)   \right) \allowdisplaybreaks \\
    &= \sum_{j^*_t \in \mathcal{A}} \mathbb{P} \left(\widehat{\theta}_{t,j^{*,0}} - \widehat{\theta}_{t,j^*_t} < \widehat{s}^\mathrm{pr}_{t,j^*_t} - \widehat{s}^\mathrm{pr}_{t,j^{*,0}} \right) \allowdisplaybreaks \label{eq:prop7-1.}
\end{align}
We observe that by definition of $j^{*,0} = \argmax_{j \in \mathcal{A}} \widetilde{V}(j, \mathbf{s}^0;\boldsymbol{\theta}^0) = \argmax_{j \in \mathcal{A}} \theta^0_j - \left(\max_{a \in \mathcal{A}} s^0_a \right) + s^0_j$, it follows that $\theta^0_{j^*_t} - \theta^0_{j^{*,0}}  \leq s^0_{j^{*,0}} - s^0_{j^*_t}$. Then,
\begin{align}
   (\ref{eq:prop7-1.}) &= \sum_{j^*_t \in \mathcal{A}}  \mathbb{P} \left((\widehat{\theta}_{t,j^{*,0}} - \theta^0_{j^{*,0}}) +  (\theta^0_{j^*_t} - \widehat{\theta}_{t,j^*_t}) < (\widehat{s}^\mathrm{pr}_{t,j^*_t} - s^0_{j^*_t}) + (s^0_{j^{*,0}} - \widehat{s}^\mathrm{pr}_{t,j^{*,0}}) \right) \allowdisplaybreaks \\
   &= \sum_{j^*_t \in \mathcal{A}} \mathbb{P} \left((\widehat{\theta}_{t,j^{*,0}} - \theta^0_{j^{*,0}}) + (\theta^0_{j^*_t} - \widehat{\theta}_{t,j^*_t}) < (\widehat{s}^\mathrm{pr}_{t,j^*_t} - s^0_{j^*_t}) + (s^0_{j^{*,0}} - \widehat{s}^\mathrm{pr}_{t,j^{*,0}}) \Big| \|\mathbf{s}^0 - \widehat{\mathbf{s}}^\mathrm{pr}_t \|_\infty \leq \beta_t \right)  \notag \allowdisplaybreaks \\
   & \hspace{11cm} \cdot \mathbb{P} \left(\|\mathbf{s}^0 - \widehat{\mathbf{s}}^\mathrm{pr}_t \|_\infty \leq \beta_t \right) \notag \allowdisplaybreaks \\
   &\hspace{1cm} + \mathbb{P} \left((\widehat{\theta}_{t,j^{*,0}} - \theta^0_{j^{*,0}}) + (\theta^0_{j^*_t} - \widehat{\theta}_{t,j^*_t}) < (\widehat{s}^\mathrm{pr}_{t,j^*_t} - s^0_{j^*_t}) + (s^0_{j^{*,0}} - \widehat{s}^\mathrm{pr}_{t,j^{*,0}}) \Big| \|\mathbf{s}^0 - \widehat{\mathbf{s}}^\mathrm{pr}_t \|_\infty > \beta_t \right)  \notag \allowdisplaybreaks \\
   &\hspace{11cm} \cdot  \mathbb{P} \left(\|\mathbf{s}^0 - \widehat{\mathbf{s}}^\mathrm{pr}_t \|_\infty > \beta_t \right) \notag \allowdisplaybreaks \\ 
    &\leq \sum_{j^*_t \in \mathcal{A}}  \mathbb{P} \left((\widehat{\theta}_{t,j^{*,0}} - \theta^0_{j^{*,0}}) +  (\theta^0_{j^*_t} - \widehat{\theta}_{t,j^*_t}) \leq 2 \beta_t \right) + n\mathbb{P} \left(\|\mathbf{s}^0 - \widehat{\mathbf{s}}^\mathrm{pr}_t \|_\infty > \beta_t \right) \allowdisplaybreaks \\
    &\leq \sum_{j^*_t \in \mathcal{A}}  \mathbb{P} \left((\widehat{\theta}_{t,j^{*,0}} - \theta^0_{j^{*,0}}) +  (\theta^0_{j^*_t} - \widehat{\theta}_{t,j^*_t}) \leq 2 \beta_t \right) \notag \allowdisplaybreaks \\
    &\hspace{1cm} + 2n\exp \left(-\frac{2\lambda_t^2}{(t-1)16n(6R_{\max} - 6R_{\min} + 2\gamma)^2} - \log \beta_t + n \log (2R_{\max} - 2R_{\min})\right) \label{eq:pff3}
\end{align} 
where the last inequality follows by Theorem \ref{thm:concen}. Then, we can bound the first term in the last result as follows.
\begin{align}
    &\sum_{j^*_t \in \mathcal{A}} \mathbb{P} \left((\widehat{\theta}_{t,j^{*,0}} - \theta^0_{j^{*,0}}) \leq 2 \beta_t - (\theta^0_{j^*_t} - \widehat{\theta}_{t,j^*_t}) \right) \notag \allowdisplaybreaks \\
    &= \sum_{j^*_t \in \mathcal{A}} \mathbb{P} \left((\widehat{\theta}_{t,j^{*,0}} - \theta^0_{j^{*,0}}) \leq 2 \beta_t - (\theta^0_{j^*_t} - \widehat{\theta}_{t,j^*_t}) \Big | \theta^0_{j^*_t} - \widehat{\theta}_{t,j^*_t} < 3\beta_t \right) \mathbb{P} \left(\theta^0_{j^*_t} - \widehat{\theta}_{t,j^*_t} < 3\beta_t \right)  \notag \allowdisplaybreaks \\
    &\quad +  \mathbb{P} \left((\widehat{\theta}_{t,j^{*,0}} - \theta^0_{j^{*,0}}) \leq 2 \beta_t - (\theta^0_{j^*_t} - \widehat{\theta}_{t,j^*_t}) \Big | \theta^0_{j^*_t} - \widehat{\theta}_{t,j^*_t} \geq  3\beta_t \right) \mathbb{P} \left(\theta^0_{j^*_t} - \widehat{\theta}_{t,j^*_t} \geq  3\beta_t \right) \allowdisplaybreaks \\
    &\leq \sum_{j^*_t \in \mathcal{A}}  \mathbb{P} \left(\widehat{\theta}_{t,j^{*,0}} - \theta^0_{j^{*,0}} \leq -\beta_t \right) +  \mathbb{P} \left(\widehat{\theta}_{t,j^*_t} - \theta^0_{j^*_t} \leq -3\beta_t \right) \allowdisplaybreaks \\
    &\leq \sum_{j^*_t \in \mathcal{A}} \mathbb{P} \left(\widehat{\theta}_{t,j^{*,0}} - \theta^0_{j^{*,0}} \leq -\beta_t \right) +  \mathbb{P} \left(\widehat{\theta}_{t,j^*_t} - \theta^0_{j^*_t} \leq -\beta_t \right) \label{eq:pff2} \allowdisplaybreaks
\end{align}
Note that bounding the two probability terms in the right-hand side of the last inequality follows a very similar approach to each other due to the definition of $\widehat{\theta}_{t,a}$'s given in (\ref{eq:theta-hat}). For any $a \in \mathcal{A}$, let $\overline{T}(a, t)$ be the number of time steps $\tau \in [\widetilde{k}, t-1]$ where: the agent picks the true utility-maximizer arm, the principal explores, and arm $a$ is the true utility-maximizer arm for the agent. Then, $\overline{T}(a, t)$ is the sum of $t - \widetilde{k} - 1$ independent indicator variables that are independent Bernoulli random variables with success probabilities $(1 - p_\tau) \epsilon_\tau^{\mathrm{pr}} \mathbb{P} \left(a = \argmax_{a' \in \mathcal{A}} {s^0_{a'} + \pi_{\tau, a'}}\right)$. Further, we note that $T(a, t) \geq \overline{T}(a, t)$. Then, for any $a \in \mathcal{A}$, we have 
\begin{align}
    &\mathbb{P} \left(\widehat{\theta}_{t,a} - \theta^0_{a} \leq -\beta_t \right)  \notag \allowdisplaybreaks \\
    &\leq \mathbb{P} \left(\widehat{\theta}_{t,a} - \theta^0_{a} \leq -\beta_t \Big| \overline{T}(a, t) > \mathbb{E} \overline{T}(a, t)/2\right)\mathbb{P} \left(\overline{T}(a, t) > \mathbb{E} \overline{T}(a, t)/2\right) \notag \allowdisplaybreaks \\
    &\quad + \mathbb{P} \left(\widehat{\theta}_{t,a} - \theta^0_{a} \leq -\beta_t \Big| \overline{T}(a, t) \leq \mathbb{E} \overline{T}(a, t)/2 \right)\mathbb{P} \left(\overline{T}(a, t) \leq \mathbb{E} \overline{T}(a, t)/2 \right)  \allowdisplaybreaks \\
    &\leq \mathbb{P} \left(\widehat{\theta}_{t,a} - \theta^0_{a} \leq -\beta_t \Big| \overline{T}(a, t) > \mathbb{E} \overline{T}(a, t)/2\right) +  \mathbb{P} \left(\overline{T}(a, t) \leq \mathbb{E} \overline{T}(a, t)/2\right)  \allowdisplaybreaks \intertext{Next, we use Hoeffding's Inequality \citep{boucheron2013concentration} and obtain.}
    &\leq \exp\left(- \frac{\mathbb{E} \overline{T}(a, t) \beta^2_{t}}{(\overline{C}  - \underline{C})^2}\right) + \exp \left(- t \frac{\left(\mathbb{E} \overline{T}(a, t)\right)^2}{4}\right) \label{eq:pff}
\end{align}
We recall that our research problems are well-posed by Assumption \ref{assm1} which ensures that the principal is able to provide incentives whose magnitudes are sufficiently large to steer the agent's decisions. This further implies that the rewards of the principal should be large enough to compensate these incentives. Therefore, in the denominator of the first term above, we take the range of $\mu_{\tau, a}$ as $[\underline{C}, \overline{C}]$ in in accordance with Assumption \ref{assm1}. 

The next step is to derive a lower bound for $\mathbb{E} \overline{T}(a, t)$ by using the definition of $\overline{T}(a, t)$ and Assumption \ref{assm:agent}. 
\begin{align}
    \mathbb{E}\overline{T}(a, t) &= \sum_{\tau = \widetilde{k}}^{t-1} (1 - p_\tau) \epsilon_\tau^{\mathrm{pr}} \mathbb{P} \left(a = \argmax_{a' \in \mathcal{A}} {s^0_{a'} + \pi_{\tau, a'}}\right) \allowdisplaybreaks \\
    &\geq \sum_{\tau = \widetilde{k}}^{t-1} \left(1 - k \frac{\sqrt{\log 2\tau}}{\sqrt{\tau}} \right) \frac{m^\mathrm{pr}}{\tau^{(1/2 - w)}}  \mathbb{P} \left(a = \argmax_{a' \in \mathcal{A}} {s^0_{a'} + \pi_{\tau, a'}}\right)   \allowdisplaybreaks
\end{align}
We compute a lower bound on the probability $\mathbb{P} \left(a = \argmax_{a' \in \mathcal{A}} {s^0_{a'} + \pi_{\tau, a'}} \right)$ by using the cdf that we derived in (\ref{eq:cdf}). Since the cdf is a piecewise function, we can consider the case $\underline{C} - \overline{C} \leq s^0_{a} - s^0_{a'} < 0$ to derive a lower bound. 
\begin{align}
    \mathbb{P} \left(a = \argmax_{a' \in \mathcal{A}} {s^0_{a'} + \pi_{\tau, a'}} \right) &= \mathbb{P} \left(s^0_{a'} + \pi_{\tau, a'} < s^0_{a} + \pi_{\tau, a}, \ \forall a' \in \mathcal{A} \setminus \{a\} \right) \allowdisplaybreaks \\
    &= \prod_{a' \in \mathcal{A} \setminus \{a\}} \mathbb{P} \left(\pi_{\tau, a'} - \pi_{\tau, a} < s^0_{a} - s^0_{a'} \right) \allowdisplaybreaks \\
    &\geq \prod_{a' \in \mathcal{A}\setminus \{a\}}   \mathbb{P} \left(\pi_{\tau, a'} - \pi_{\tau, a} < s^0_{a} - s^0_{a'}, \ \underline{C} - \overline{C} \leq s^0_{a} - s^0_{a'} < 0 \right) \allowdisplaybreaks \\
    &= \prod_{a' \in \mathcal{A}\setminus \{a\}}  \frac{(s^0_{a} - s^0_{a'} + \overline{C}- \underline{C})^2}{2(\overline{C}- \underline{C})^2} \allowdisplaybreaks \\
    &\geq \prod_{a' \in \mathcal{A}\setminus \{a\}}   \frac{(s^0_{a} + \gamma)^2}{2(\overline{C}- \underline{C})^2} \allowdisplaybreaks \\
    &= \frac{(s^0_{a} + \gamma)^{2n-2}}{2^{n-1}(\overline{C}- \underline{C})^{2n-2}}  \allowdisplaybreaks
\end{align}
where the last inequality follows since $s^0_{a'} \leq R_{\max} - R_{\min}$ for all $a' \in \mathcal{A}$ and $\overline{C}- \underline{C} = R_{\max} - R_{\min} + \gamma$ by definition. Then, we have
\begin{align}
    \mathbb{E}\overline{T}(a, t) &\geq  \frac{(s^0_{a} + \gamma)^{2n-2}}{2^{n-1}(\overline{C}- \underline{C})^{2n-2}}  \sum_{\tau = \widetilde{k}}^{t-1} \left(1 - k \frac{\sqrt{\log 2\tau}}{\sqrt{\tau}} \right) \frac{m^\mathrm{pr}}{\tau^{(1/2 - w)}} \allowdisplaybreaks \\
    &\geq \frac{(s^0_{a} + \gamma)^{2n-2}}{2^{n-1}(\overline{C}- \underline{C})^{2n-2}} \left( 1  - k \textstyle \frac{ \sqrt{\log 2\widetilde{k}}}{ \sqrt{\widetilde{k}}} \right) m^\mathrm{pr} \sum_{\tau = \widetilde{k}}^{t-1}  \frac{1}{\tau^{(1/2 - w)}} \allowdisplaybreaks \\
    &\geq \frac{(s^0_{a} + \gamma)^{2n-2}}{2^{n-1}(\overline{C}- \underline{C})^{2n-2}} \left( 1  - k \textstyle \frac{ \sqrt{\log 2\widetilde{k}}}{ \sqrt{\widetilde{k}}} \right) m^\mathrm{pr} \int_{\tau = \widetilde{k}}^{t} \frac{1}{\tau^{(1/2 - w)}}d\tau \allowdisplaybreaks \\
    &= \frac{(s^0_{a} + \gamma)^{2n-2}}{2^{n-1}(\overline{C}- \underline{C})^{2n-2}} \left( 1  - k \textstyle \frac{ \sqrt{\log 2\widetilde{k}}}{ \sqrt{\widetilde{k}}} \right) m^\mathrm{pr} \frac{2\left(t^{w + 1/2} - \widetilde{k}^{w + 1/2}\right)}{2w + 1} \\
    &\geq \frac{(s^0_{a} + \gamma)^{2n-2}}{2^{n-1}(\overline{C}- \underline{C})^{2n-2}} \left( 1  - k \textstyle \frac{ \sqrt{\log 2\widetilde{k}}}{ \sqrt{\widetilde{k}}} \right) m^\mathrm{pr} \frac{2 - 2w - 1}{2w + 1} t^{w + 1/2} \end{align}
where the last inequality always holds for $t \geq \widetilde{k}$ since $0 < w < 1/4$ by definition. Combining this last result with (\ref{eq:pff}), we get 
\begin{align}
    &\mathbb{P} \left(\widehat{\theta}_{t,a} - \theta^0_{a} \leq -\beta_t \right) \notag \allowdisplaybreaks \\
    &\leq \exp\left(- \frac{\frac{(s^0_{a} + \gamma)^{2n-2}}{2^{n-1}(\overline{C}- \underline{C})^{2n-2}} \left( 1  - k \textstyle \frac{ \sqrt{\log 2\widetilde{k}}}{ \sqrt{\widetilde{k}}} \right) m^\mathrm{pr} \frac{2 - 2w - 1}{2w + 1} t^{w + 1/2} \beta^2_{t}}{(\overline{C}  - \underline{C})^2}\right) + \frac{1}{t^{2w + 2}} \allowdisplaybreaks 
    \intertext{We now substitute $\beta_t = B\dfrac{\sqrt{\log 2t}}{t^{w/3}}$ as given in Algorithm \ref{alg:pr}. }
    &\leq \exp\left(- \frac{\frac{(s^0_{a} + \gamma)^{2n-2}}{2^{n-1}(\overline{C}- \underline{C})^{2n-2}} \left( 1  - k \textstyle \frac{ \sqrt{\log 2\widetilde{k}}}{ \sqrt{\widetilde{k}}} \right) m^\mathrm{pr} \frac{2 - 2w - 1}{2w + 1} t^{w + 1/2} \frac{9 k^2 \left(R_{\max} - R_{\min} + \gamma\right)^{2n} \sqrt[3]{32n}}{\left(1 - k \sqrt{\log 2\widetilde{k}}/ \sqrt{\widetilde{k}}\right)^2} t^{w + 1/2} \frac{\log 2t}{t^{2w/3}}}{(\overline{C}  - \underline{C})^2}\right) \allowdisplaybreaks  \\
    &\leq \exp\left(- \frac{(s^0_{a} + \gamma)^{2n-2}}{2^{n-1}} m^\mathrm{pr} \frac{2 - 2w - 1}{2w + 1} \frac{9 k^2  \sqrt[3]{32n}}{\left(1 - k \sqrt{\log 2\widetilde{k}}/ \sqrt{\widetilde{k}}\right)} t^{w + 1/2} \frac{\log 2t}{t^{2w/3}}\right) \allowdisplaybreaks  
    \intertext{where the last inequality follows since $\overline{C}- \underline{C} = R_{\max} - R_{\min} + \gamma$ where $\gamma > 0$ by definition.}
\end{align}
Now, notice that we can bound each of the constant terms in the last line from below by 1 since we have $n \geq 2$, $m^\mathrm{pr} \geq 1$, $0 < \gamma \leq R_{\max} - R_{\min} - 1$, $s^0_{a} \geq R_{\min} - R_{\max}$, $0 < w < 1/4$, $k \geq 1$, and $k \sqrt{\log 2\widetilde{k}} < \sqrt{\widetilde{k}}$ by definition. Then, 
\begin{align}
   \mathbb{P} \left(\widehat{\theta}_{t,a} - \theta^0_{a} \leq -\beta_t \right)  &\leq \exp\left(-  t^{w + 1/2} \frac{\log 2t}{t^{2w/3}}\right) \leq \exp(-\log 2t) = \frac{1}{2t}
\end{align}
Substituting this upper bound in (\ref{eq:pff2}), we obtain 
\begin{align}
    \sum_{j^*_t \in \mathcal{A}}  \mathbb{P} \left((\widehat{\theta}_{t,j^{*,0}} - \theta^0_{j^{*,0}}) +  (\theta^0_{j^*_t} - \widehat{\theta}_{t,j^*_t}) \leq 2 \beta_t \right) \leq \frac{n}{t} \label{eq:prop7-1}\allowdisplaybreaks
\end{align}
Next, we compute an upper bound for the second part of (\ref{eq:pff3}). We recall the definition of $\lambda_t$ as given in (\ref{eq:lambda_t}). 
\begin{align}
    \lambda_t = \frac{4\alpha \left(1 - k \sqrt{\log 2\widetilde{k}}/ \sqrt{\widetilde{k}}\right)^2}{27} \beta_t^3 \mathbb{E}\eta(\widetilde{k}, t) - 3 k \left(3 (R_{\max} - R_{\min}) + \gamma\right) \sqrt{t\log (2t)} \label{eq:lambda_proof_0}
\end{align}
where $\eta(\widetilde{k}, t)$ is defined in (\ref{eq:eta}) as the number of time steps within the time interval $[\widetilde{k}, t-1]$ where the principal chooses each incentive $\pi_{t,a}$ uniformly randomly from the compact set $\mathcal{C}$. By this definition, we can compute a lower bound for $\mathbb{E}\eta(\widetilde{k}, t)$ as follows. 
\begin{align}
     \mathbb{E}\eta(\widetilde{k}, t) = \sum_{\tau = \widetilde{k}}^{t-1} \min \left\{1, m^\mathrm{pr} \frac{1}{\tau^{(1/2 - w)}} \right\} \geq m^\mathrm{pr} \sum_{\tau = \widetilde{m}}^{t-1} \frac{1}{\tau^{(1/2 - w)}} &\geq m^\mathrm{pr} \int_{\tau = \widetilde{m}}^{t} \frac{1}{\tau^{(1/2 - w)}} d\tau \allowdisplaybreaks  \\
     &= \frac{2m^\mathrm{pr}}{2w + 1} \left( t^{w + 1/2} -  \widetilde{m}^{w + 1/2}\right) \allowdisplaybreaks  \\
     &\geq \frac{2m^\mathrm{pr} - 2w  - 1}{2w + 1} t^{w + 1/2}  \allowdisplaybreaks 
\end{align}
where $\widetilde{m} \geq \widetilde{k}$ is the minimum value satisfying $m^\mathrm{pr} \leq \widetilde{m}^{(1/2 - w)}$. Then, the last inequality always holds for $t \geq \widetilde{m}$, $0 < w < 1/4$, and $m^\mathrm{pr} \geq 1$. Using this lower bound for $\mathbb{E}\eta(\widetilde{k}, t)$, we obtain
\begin{align}
    \lambda_t &\geq \sqrt{t}\Bigg(\frac{4\alpha \big(1 - k \sqrt{\log 2\widetilde{k}}/ \sqrt{\widetilde{k}}\big)^2 (2m^\mathrm{pr} - 2w  - 1)}{27(2w + 1)} \beta^3_t t^{w}  - 3 k \left(3 (R_{\max} - R_{\min}) + \gamma\right) \sqrt{ \log 2t} \Bigg) \allowdisplaybreaks
\intertext{For the specified choice of $\beta_t$, we further have}
    &\geq 3 k \left(3 (R_{\max} - R_{\min}) + \gamma\right) \sqrt{t \log 2t}  \Bigg(\textstyle \frac{4\alpha (2m^\mathrm{pr} - 2w  - 1) k^2 \left(3 (R_{\max} - R_{\min}) + \gamma\right)^{3n-1} \sqrt{32n}}{3(2w + 1) \big(1 - k \sqrt{\log 2\widetilde{k}}/ \sqrt{\widetilde{k}}\big)} \log (2t) - 1 \Bigg) \\
    &\geq 3 k \left(3 (R_{\max} - R_{\min}) + \gamma\right) \sqrt{t \log 2t}  \left(\sqrt{32n} \log (2t) - 1 \right) \label{eq:alpha}\\
    &\geq 3 k \left(3 (R_{\max} - R_{\min}) + \gamma\right) \sqrt{t \log 2t}  \left(\sqrt{32n} \log (2t) - \frac{\sqrt{32n}}{3} \log (2t) \right) \\
    &\geq 2\sqrt{32n} k \left(3 (R_{\max} - R_{\min}) + \gamma\right) \sqrt{t \log 2t} \log (2t)
\end{align}
where (\ref{eq:alpha}) follows by the fact that we can bound each constant term that appears in the coefficient of $\log (2t)$ from below by 1 since we have $n \geq 2$, $k \geq 1$, $0 < w < 1/4$, $m^\mathrm{pr} \geq 1$, $k \sqrt{\log 2\widetilde{k}} < \sqrt{\widetilde{k}}$, and $\alpha = \mathrm{constant} / \left(R_{\max} - R_{\min} + \gamma\right)^n$ for some constant $ > 0$ as introduced in Proposition \ref{prop:iden4}. Also, second to the last inequality above follows for all $t \geq 1$ since $n \geq 2$ by definition. Then, 
\begin{align}
    &\exp \left(-\frac{2\lambda_t^2}{16n(6R_{\max} - 6R_{\min} + 2\gamma)^2 (t-1)} - \log \beta_t + n \log (2R_{\max} - 2R_{\min})\right)  \notag \allowdisplaybreaks \\
    &\leq \exp \left(-\frac{\lambda_t^2}{32n(3R_{\max} - 3R_{\min} + \gamma)^2 t } - \log \beta_t + n \log (2R_{\max} - 2R_{\min})\right)  \allowdisplaybreaks \\
    &\leq \exp \left(-4 k^2  (\log 2t)^3  - \log \left( \frac{3 k \left(3 (R_{\max} - R_{\min}) + \gamma\right)^n \sqrt[6]{32n}}{1 - k \sqrt{\log 2\widetilde{k}}/ \sqrt{\widetilde{k}}}\frac{\sqrt{\log 2t}}{t^{w/3}} \right) + n \log (2R_{\max} - 2R_{\min})\right)  \allowdisplaybreaks \\
    &\leq \exp \left(- (\log 2t)^3  - \log \left( \frac{3 k \left(3 (R_{\max} - R_{\min}) + \gamma\right)^n \sqrt[6]{32n}}{1 - k \sqrt{\log 2\widetilde{k}}/ \sqrt{\widetilde{k}}}\frac{\sqrt{\log 2t}}{t^{w/3}} \right) + n \log (2R_{\max} - 2R_{\min})\right)  \allowdisplaybreaks \\
    &\leq \frac{(2R_{\max} - 2R_{\min})^n \left(1 - k \sqrt{\log 2\widetilde{k}}/ \sqrt{\widetilde{k}}\right)}{3 k \left(3 (R_{\max} - R_{\min}) + \gamma\right)^n \sqrt[6]{32n}} \frac{1}{t^{w/3 + 1/2}}  \frac{t^{w/3}} {\sqrt{\log 2t}} \allowdisplaybreaks \\
    &= \frac{2^n }{3^{n + 1} k \sqrt[6]{32n}}  \frac{1}{\sqrt{t \log 2t}} \label{eq:lambda_proof_last}
\end{align}
where second to the last inequality holds for all $t \geq 1$ and $0 < w < 1/4$ since $\exp( - (\log 2t)^3) \leq \exp( - (\log 2t)) = \dfrac{1}{t} \leq  \dfrac{1}{t^{w/3 + 1/2}}$.

Lastly, substituting the upper bounds in (\ref{eq:prop7-1}) and (\ref{eq:lambda_proof_last}) into (\ref{eq:pff3}), we get
\begin{align}
     \mathbb{P} \left(j^*_t \neq j^{*,0} \right) &\leq \frac{n}{t} + \frac{n^{5/6} 2^{n+1}}{3^{n + 1} k \sqrt[6]{32}} \frac{1}{\sqrt{t \log 2t}}\allowdisplaybreaks 
\end{align}
\Halmos \endproof

\proof{\textbf{Proof of Theorem \ref{thm:regret}.}}
We start by recalling the definition of our regret notion given in (\ref{eq:regretdefn}) and decompose it into two main parts: 1) total costs incurred due to the offered incentives, 2) total expected rewards collected through the arms chosen by the agent.
\begin{align}
     \mathrm{Regret}\left(\Pi_{\epsilon, T} \right) &= \sum_{t \in \mathcal{T}} V(\mathbf{c}(\boldsymbol{\theta}^0, \mathbf{s}^0); \boldsymbol{\theta}^0) - V_t(\boldsymbol{\pi}_t; \boldsymbol{\theta}^0)  \allowdisplaybreaks  \\
    &= \sum_{t \in \mathcal{T}} \sum_{a \in \mathcal{A}} \left[\pi_{t,a} - c_{a}(\boldsymbol{\theta}^0, \mathbf{s}^0) \right]  + \sum_{t \in \mathcal{T}} \left[\theta^0_{\upsilon(\mathbf{c}(\boldsymbol{\theta}^0, \mathbf{s}^0))} -\theta^0_{\upsilon_t(\boldsymbol{\pi}_t)} \right]  \allowdisplaybreaks  \label{eq:stoc-regret0}
\end{align}
We let $\mathcal{T}^{\mathrm{pr-xplore}} \in \mathcal{T}$ and $\mathcal{T}^{\mathrm{pr-xploit}} \in \mathcal{T}$ be the set of random time steps that the principal's algorithm (\ref{alg:pr}) performs exploration (lines \ref{alg-explore1}-\ref{alg-explorelast}) and exploitation (lines \ref{alg-exploit1}-\ref{alg-exploitlast}), respectively. First, we bound the first part of (\ref{eq:stoc-regret0}) as follows. 
\begin{align}
   \sum_{t \in \mathcal{T}} \sum_{a \in \mathcal{A}} \left[\pi_{t,a} - c_{a}(\boldsymbol{\theta}^0, \mathbf{s}^0) \right] &= \sum_{t \in \mathcal{T}^\mathrm{pr-xplore}} \sum_{a \in \mathcal{A}} \left[\pi_{t,a} - c_{a}(\boldsymbol{\theta}^0, \mathbf{s}^0) \right] + \sum_{t \in \mathcal{T}^\mathrm{pr-xploit}} \sum_{a \in \mathcal{A}} \left[ c_a(\widehat{\boldsymbol{\theta}}_t, \widehat{\mathbf{s}}^\mathrm{pr}_t)  -  c_{a}(\boldsymbol{\theta}^0, \mathbf{s}^0) \right] \label{eq:stoc-regret1} \allowdisplaybreaks
\end{align}
Notice that the cardinalities of the defined random sets, $|\mathcal{T}^\mathrm{pr-xplore}|$ and $|\mathcal{T}^\mathrm{pr-xploit}|$, are random variables. Then, the first part of (\ref{eq:stoc-regret1}) is bounded by considering the following conditional expectation.
\begin{align}
    \mathbb{E} \bigg[ \sum_{t \in \mathcal{T}^\mathrm{pr-xplore}} \sum_{a \in \mathcal{A}} \pi_{t,a} - c_{a}(\boldsymbol{\theta}^0, \mathbf{s}^0) \Big | \mathcal{T}^\mathrm{pr-xplore}\bigg] \leq  n (\overline{C} - \underline{C}) |\mathcal{T}^\mathrm{pr-xplore}|  \allowdisplaybreaks \label{eq:stoc-explore-1}
\end{align}
Taking the expectation of both sides in the last inequality, we obtain 
\begin{align}
    \sum_{t \in \mathcal{T}^\mathrm{pr-xplore}} \sum_{a \in \mathcal{A}} \pi_{t,a} - c_{a}(\boldsymbol{\theta}^0, \mathbf{s}^0) \leq n (\overline{C} - \underline{C}) \mathbb{E} |\mathcal{T}^\mathrm{pr-xplore}| &= n (\overline{C} - \underline{C}) \sum_{t=1}^T \epsilon^\mathrm{pr}_t \allowdisplaybreaks \\
    &= n (\overline{C} - \underline{C}) \sum_{t=1}^T \min \left\{1, \frac{m^\mathrm{pr}}{ t^{(1/2 - w)}} \right\}  \allowdisplaybreaks \\
    &\leq n m^\mathrm{pr} (\overline{C} - \underline{C}) \sum_{t=1}^T  \frac{1}{ t^{1/2 - w}}  \allowdisplaybreaks \\
    &\leq m^\mathrm{pr} n (\overline{C} - \underline{C})\left(1 + \int_{t=1}^T \frac{1}{ t^{1/2 - w}}  dt \right) \allowdisplaybreaks \\
    &=  m^\mathrm{pr} n (\overline{C} - \underline{C})\left(\frac{2}{2w + 1} T^{w + 1/2} + \frac{2w - 1}{2w + 1}\right) \allowdisplaybreaks \label{eq:stoc-explore-last}
\end{align}
Next, we bound the second part of (\ref{eq:stoc-regret1}) again by considering a conditional expectation.
\begin{align}
    &\mathbb{E} \bigg[ \sum_{t \in \mathcal{T}^\mathrm{pr-xploit}} \sum_{a \in \mathcal{A}} c_a(\widehat{\boldsymbol{\theta}}_t, \widehat{\mathbf{s}}^\mathrm{pr}_t) - c_{a}(\boldsymbol{\theta}^0, \mathbf{s}^0) \Big | \mathcal{T}^\mathrm{pr-xploit}\bigg] \nonumber \allowdisplaybreaks \\
    &= \sum_{t \in \mathcal{T}^\mathrm{pr-xploit}} \sum_{a \in \mathcal{A}}  \mathbb{E} \left[ c_a(\widehat{\boldsymbol{\theta}}_t, \widehat{\mathbf{s}}^\mathrm{pr}_t) - c_{a}(\boldsymbol{\theta}^0, \mathbf{s}^0) \Big| \mathcal{T}^\mathrm{pr-xploit}, \|\mathbf{s}^0 - \widehat{\mathbf{s}}^{\mathrm{pr}}_t \|_\infty \leq \beta_t \right] \mathbb{P}\left(\|\mathbf{s}^0 - \widehat{\mathbf{s}}^{\mathrm{pr}}_t \|_\infty  \leq \beta_t\right) \nonumber \allowdisplaybreaks \\
    &\quad + \sum_{t \in \mathcal{T}^\mathrm{pr-xploit}} \sum_{a \in \mathcal{A}} \mathbb{E} \left[ c_a(\widehat{\boldsymbol{\theta}}_t, \widehat{\mathbf{s}}^\mathrm{pr}_t) - c_{a}(\boldsymbol{\theta}^0, \mathbf{s}^0) \Big| \mathcal{T}^\mathrm{pr-xploit}, \|\mathbf{s}^0 - \widehat{\mathbf{s}}^{\mathrm{pr}}_t \|_\infty > \beta_t \right] \mathbb{P}\left(\|\mathbf{s}^0 - \widehat{\mathbf{s}}^{\mathrm{pr}}_t \|_\infty > \beta_t\right) \allowdisplaybreaks \\
    &\leq \sum_{t \in \mathcal{T}^\mathrm{pr-xploit}} \sum_{a \in \mathcal{A}}  \mathbb{E} \left[ c_a(\widehat{\boldsymbol{\theta}}_t, \widehat{\mathbf{s}}^\mathrm{pr}_t) - c_{a}(\boldsymbol{\theta}^0, \mathbf{s}^0) \Big| \mathcal{T}^\mathrm{pr-xploit}, \|\mathbf{s}^0 - \widehat{\mathbf{s}}^{\mathrm{pr}}_t \|_\infty \leq \beta_t \right] \nonumber \allowdisplaybreaks \\
    &\quad + 2 n (\overline{C} - \underline{C}) \sum_{t \in \mathcal{T}^\mathrm{pr-xploit}} \exp \left(-\frac{2\lambda_t^2}{(t-1)16n(6R_{\max} - 6R_{\min} + 2\gamma)^2} - \log \beta_t + n \log (2R_{\max} - 2R_{\min})\right) \label{eq:stoc-regret2}
\end{align}
where the last inequality follows by Corollary \ref{cor:concen}. For the first term in (\ref{eq:stoc-regret2}), we have 
\begin{align}
    &\sum_{t \in \mathcal{T}^\mathrm{pr-xploit}} \sum_{a \in \mathcal{A}}  \mathbb{E} \left[ c_a(\widehat{\boldsymbol{\theta}}_t, \widehat{\mathbf{s}}^\mathrm{pr}_t) - c_{a}(\boldsymbol{\theta}^0, \mathbf{s}^0) \Big| \mathcal{T}^\mathrm{pr-xploit}, \|\mathbf{s}^0 - \widehat{\mathbf{s}}^{\mathrm{pr}}_t \|_\infty \leq \beta_t \right]\notag \allowdisplaybreaks \\
    &= \sum_{t \in \mathcal{T}^\mathrm{pr-xploit}} \sum_{a \in \mathcal{A}}  \mathbb{E} \left[ c_a(\widehat{\boldsymbol{\theta}}_t, \widehat{\mathbf{s}}^\mathrm{pr}_t) - c_{a}(\boldsymbol{\theta}^0, \mathbf{s}^0) \Big| \mathcal{T}^\mathrm{pr-xploit}, \|\mathbf{s}^0 - \widehat{\mathbf{s}}^{\mathrm{pr}}_t \|_\infty \leq \beta_t, \ j^*_t = j^{*,0}\right] \mathbb{P}\left(j^*_t = j^{*,0}\right)  \notag \allowdisplaybreaks \\
    &\hspace{2.5cm} + \mathbb{E} \left[ c_a(\widehat{\boldsymbol{\theta}}_t, \widehat{\mathbf{s}}^\mathrm{pr}_t) - c_{a}(\boldsymbol{\theta}^0, \mathbf{s}^0) \Big| \mathcal{T}^\mathrm{pr-xploit}, \|\mathbf{s}^0 - \widehat{\mathbf{s}}^{\mathrm{pr}}_t \|_\infty \leq \beta_t, \ j^*_t \neq j^{*,0}\right] \mathbb{P}\left(j^*_t \neq j^{*,0}\right)  \allowdisplaybreaks \\
    &\leq  \sum_{t \in \mathcal{T}^\mathrm{pr-xploit}} \mathbb{E} \Big[ \max_{a \in \mathcal{A}} \widehat{s}^\mathrm{pr}_{t,a} - \widehat{s}^\mathrm{pr}_{t,j^*_t} + 2\beta_t - \max_{a \in \mathcal{A}} s^0_a + s^0_{j^{*,0}} \Big | \mathcal{T}^\mathrm{pr-xploit}, \|\mathbf{s}^0 - \widehat{\mathbf{s}}^{\mathrm{pr}}_t \|_\infty \leq \beta_t, \ j^*_t = j^{*,0}\Big] \notag \allowdisplaybreaks \\
    &\hspace{2.5cm} + n (\overline{C} - \underline{C}) \mathbb{P}\left(j^*_t \neq j^{*,0}\right) \allowdisplaybreaks
    \intertext{For convenience, we continue by using the indices $\kappa_t \in \argmax_{a \in \mathcal{A}} \widehat{s}^\mathrm{pr}_{t,a} $ and $\kappa^0 \in \argmax_{a \in \mathcal{A}} s^0_a $.}
    &=  \sum_{t \in \mathcal{T}^\mathrm{pr-xploit}} \mathbb{E} \left[ \widehat{s}^\mathrm{pr}_{t,\kappa_t} - \widehat{s}^\mathrm{pr}_{t,j^*_t} + 2\beta_t - s^0_{\kappa^0} + s^0_{j^{*,0}} \Big | \mathcal{T}^\mathrm{pr-xploit}, \|\mathbf{s}^0 - \widehat{\mathbf{s}}^{\mathrm{pr}}_t \|_\infty \leq \beta_t, \ j^*_t = j^{*,0}\right] \notag \allowdisplaybreaks \\
    &\hspace{2.5cm} + n (\overline{C} - \underline{C}) \mathbb{P}\left(j^*_t \neq j^{*,0}\right) \allowdisplaybreaks \allowdisplaybreaks \\
    &=  \sum_{t \in \mathcal{T}^\mathrm{pr-xploit}} \hspace{-.2cm} \mathbb{E} \left[ (\widehat{s}^\mathrm{pr}_{t,\kappa_t} - s^0_{\kappa_t}) + (s^0_{\kappa_t} - s^0_{\kappa^0}) + (s^0_{j^{*,0}} - \widehat{s}^\mathrm{pr}_{t,j^*_t}) + 2\beta_t \Big | \mathcal{T}^\mathrm{pr-xploit}, \|\mathbf{s}^0 - \widehat{\mathbf{s}}^{\mathrm{pr}}_t \|_\infty \leq \beta_t, \ j^*_t = j^{*,0} \right] \notag \allowdisplaybreaks \\
    &\hspace{2.5cm} + n (\overline{C} - \underline{C}) \mathbb{P}\left(j^*_t \neq j^{*,0}\right)   \allowdisplaybreaks \allowdisplaybreaks \\
    &\leq  \sum_{t \in \mathcal{T}^\mathrm{pr-xploit}} 4 \beta_t + n (\overline{C} - \underline{C}) \mathbb{P}\left(j^*_t \neq j^{*,0}\right) \allowdisplaybreaks \\
    &\leq  \sum_{t \in \mathcal{T}^\mathrm{pr-xploit}} 4 \beta_t + \frac{n^2 (\overline{C} - \underline{C}) }{t} + \frac{n^{11/6} 2^{n+1} (\overline{C} - \underline{C}) }{3^{n + 1} k \sqrt[6]{32}} \frac{1}{\sqrt{t \log 2t}} \allowdisplaybreaks 
    \intertext{where the last inequality follows by Proposition \ref{prop:regret}. Substituting $\beta_t = B \dfrac{\sqrt{\log 2t}}{t^{w/3}}$, we have}
    &= \sum_{t \in \mathcal{T}^\mathrm{pr-xploit}} 4B \frac{\sqrt{\log 2t}}{t^{w/3}} + \frac{n^2 (\overline{C} - \underline{C}) }{t} + \frac{n^{11/6} 2^{n+1} (\overline{C} - \underline{C}) }{3^{n + 1} k \sqrt[6]{32}} \frac{1}{\sqrt{t \log 2t}} \allowdisplaybreaks \label{eq:firstpart-0}\\
    &= \frac{12 B}{3 - w}B |\mathcal{T}^\mathrm{pr-xploit}|^{(1 - w/3)} \sqrt{\log 2|\mathcal{T}^\mathrm{pr-xploit}|} + n^2 (\overline{C} - \underline{C}) \log |\mathcal{T}^\mathrm{pr-xploit}| \notag \allowdisplaybreaks \\
    & \hspace{9cm} + \frac{n^{11/6} 2^{n+2} (\overline{C} - \underline{C}) }{3^{n + 1} k \sqrt[6]{32}} \sqrt{|\mathcal{T}^\mathrm{pr-xploit}|}  \allowdisplaybreaks \\
    &\leq \frac{12 B}{3 - w}T^{1 - w/3} \sqrt{\log 2T} + n^2 (\overline{C} - \underline{C}) \log T + \frac{n^{11/6} 2^{n+2} (\overline{C} - \underline{C}) }{3^{n + 1} k \sqrt[6]{32}} \sqrt{T}  \allowdisplaybreaks\label{eq:firstpart-2}
\end{align} 
Now, for the same choice of $\beta_t$, we can bound the second term in (\ref{eq:stoc-regret2}) by following the same arguments as in (\ref{eq:lambda_proof_0})-(\ref{eq:lambda_proof_last}) and obtain
\begin{align}
    &2 n (\overline{C} - \underline{C}) \sum_{t \in \mathcal{T}^\mathrm{pr-xploit}} \exp \left(-\frac{2\lambda_t^2}{(t-1)16n(6R_{\max} - 6R_{\min} + 2\gamma)^2} - \log \beta_t + n \log (2R_{\max} - 2R_{\min})\right)\notag \allowdisplaybreaks \\
    &\leq \frac{2^{n+1} n^{5/6} (\overline{C} - \underline{C}) }{3^{n + 1} k \sqrt[6]{32}}  \sum_{t \in \mathcal{T}^\mathrm{pr-xploit}} \frac{1}{\sqrt{t \log 2t}} \allowdisplaybreaks \\
    &\leq  \frac{2^{n+1} n^{5/6} (\overline{C} - \underline{C}) }{3^{n + 1} k \sqrt[6]{32}} \sqrt{|\mathcal{T}^\mathrm{pr-xploit}|} \allowdisplaybreaks \\
    &\leq  \frac{2^{n+1} n^{5/6} (\overline{C} - \underline{C}) }{3^{n + 1} k \sqrt[6]{32}} \sqrt{T} \allowdisplaybreaks
\end{align}
Combining these upper bounds with (\ref{eq:stoc-regret2}), we get \begin{multline}
       \mathbb{E} \bigg[ \sum_{t \in \mathcal{T}^\mathrm{pr-xploit}} \sum_{a \in \mathcal{A}} c_a(\widehat{\boldsymbol{\theta}}_t, \widehat{\mathbf{s}}^\mathrm{pr}_t) - c_{a}(\boldsymbol{\theta}^0, \mathbf{s}^0) \Big | \mathcal{T}^\mathrm{pr-xploit}\bigg] \\ \leq \frac{12 B}{3 - w}T^{1 - w/3} \sqrt{\log 2T} + n^2 (\overline{C} - \underline{C}) \log T  + \frac{2^{n+1} n^{5/6} (\overline{C} - \underline{C})\left(1 + 2n \right)}{3^{n + 1} k \sqrt[6]{32}} \sqrt{T} 
\end{multline}
Taking the expectation of both sides in the last inequality, we obtain
\begin{multline}
    \sum_{t \in \mathcal{T}^\mathrm{pr-xploit}} \sum_{a \in \mathcal{A}} c_a(\widehat{\boldsymbol{\theta}}_t, \widehat{\mathbf{s}}^\mathrm{pr}_t) - c_{a}(\boldsymbol{\theta}^0, \mathbf{s}^0) \\ \leq \frac{12 B}{3 - w}T^{1 - w/3} \sqrt{\log 2T} + n^2 (\overline{C} - \underline{C}) \log T  + \frac{2^{n+1} n^{5/6} (\overline{C} - \underline{C})\left(1 + 2n \right)}{3^{n + 1} k \sqrt[6]{32}} \sqrt{T}
\end{multline}
Substituting this last result and the result in (\ref{eq:stoc-explore-last}) into (\ref{eq:stoc-regret1}), we provide the upper bound for the first part of our regret notion as follows.
\begin{align}
    \sum_{t \in \mathcal{T}} \sum_{a \in \mathcal{A}} \left[\pi_{t,a} - c_{a}(\boldsymbol{\theta}^0, \mathbf{s}^0) \right] &\leq m^\mathrm{pr} n (\overline{C} - \underline{C})\left(\frac{2}{2w + 1} T^{w + 1/2} + \frac{2w - 1}{2w + 1}\right) + \frac{12 B}{3 - w}T^{1 - w/3} \sqrt{\log 2T} \notag \allowdisplaybreaks \\
    & \quad + n^2 (\overline{C} - \underline{C}) \log T  + \frac{2^{n+1} n^{5/6} (\overline{C} - \underline{C})\left(1 + 2n \right)}{3^{n + 1} k \sqrt[6]{32}} \sqrt{T}
\end{align}
Next, we compute an upper bound for the second part of the regret decomposed in (\ref{eq:stoc-regret0}). 
\begin{align}
    \sum_{t \in \mathcal{T}} \left[\theta^0_{\upsilon(\mathbf{c}(\boldsymbol{\theta}^0, \mathbf{s}^0))} -\theta^0_{\upsilon_t(\boldsymbol{\pi}_t)} \right]  &=  \sum_{t \in \mathcal{T}^\mathrm{pr-xplore}} \left[\theta^0_{\upsilon(\mathbf{c}(\boldsymbol{\theta}^0, \mathbf{s}^0))} -\theta^0_{\upsilon_t(\boldsymbol{\pi}_t)} \right] +  \sum_{t \in \mathcal{T}^\mathrm{pr-xploit}} \left[\theta^0_{\upsilon(\mathbf{c}(\boldsymbol{\theta}^0, \mathbf{s}^0))} -\theta^0_{\upsilon_t(\boldsymbol{\pi}_t)} \right] \label{eq:stoc-regret3}
\end{align}
Because the principal's expected rewards belong to a known compact set $\Theta$, we let $\Theta^{\max}$ to be the upper bound on $\theta^0_a$'s. Then, we can bound the first term in the last inequality above by following a similar argument as in (\ref{eq:stoc-explore-1})-(\ref{eq:stoc-explore-last}). 
\begin{align}
    \sum_{t \in \mathcal{T}^\mathrm{pr-xplore}} \left[\theta^0_{\upsilon(\mathbf{c}(\boldsymbol{\theta}^0, \mathbf{s}^0))} -\theta^0_{\upsilon_t(\boldsymbol{\pi}_t)} \right] &\leq \Theta^{\max} \mathbb{E} |\mathcal{T}^\mathrm{pr-xplore}|\allowdisplaybreaks \\
    &\leq \Theta^{\max}m^\mathrm{pr}\left(\frac{2}{2w + 1} T^{w + 1/2} + \frac{2w - 1}{2w + 1}\right)  \label{eq:secondpart-1} \allowdisplaybreaks 
\end{align}
To bound the second term in (\ref{eq:stoc-regret3}), we again start by considering a conditional expectation. 
\begin{align}
    &\mathbb{E} \left[\sum_{t \in \mathcal{T}^\mathrm{pr-xploit}} \theta^0_{\upsilon(\mathbf{c}(\boldsymbol{\theta}^0, \mathbf{s}^0))} -\theta^0_{\upsilon_t(\boldsymbol{\pi}_t)} \Big | \mathcal{T}^\mathrm{pr-xploit} \right] \notag \allowdisplaybreaks \\
    &=  \mathbb{E} \left[\sum_{t \in \mathcal{T}^\mathrm{pr-xploit}} \theta^0_{\upsilon(\mathbf{c}(\boldsymbol{\theta}^0, \mathbf{s}^0))} -\theta^0_{\upsilon_t(\boldsymbol{\pi}_t)} \Big | \mathcal{T}^\mathrm{pr-xploit} \right] \mathbb{P} \left(\upsilon_t(\boldsymbol{\pi}_t) \neq \upsilon(\mathbf{c}(\boldsymbol{\theta}^0, \mathbf{s}^0)) \right)  \allowdisplaybreaks \\
    &\leq \Theta^{\max} \sum_{t \in \mathcal{T}^\mathrm{pr-xploit}}\mathbb{P} \left(\upsilon_t(\boldsymbol{\pi}_t) \neq \upsilon(\mathbf{c}(\boldsymbol{\theta}^0, \mathbf{s}^0)) \right) \label{eq:stoc-regret4}
\end{align} 
Now, we need to derive an upper bound for the probability that the arm chosen by the agent in response to the principal's exploitation incentives is different than the arm chosen by the perfect-knowledge agent in response to the oracle incentives. Recall that the perfect-knowledge agent always picks the true utility-maximizer arm, i.e.,  $\upsilon(\mathbf{c}(\boldsymbol{\theta}^0, \mathbf{s}^0)) = \argmax_{a \in \mathcal{A}} s^0_a + c_a^0(\boldsymbol{\theta}^0, \mathbf{s}^0) $ by construction. On the other hand, for an imperfect-knowledge learning agent, we need to take into account whether the agent picks the true utility-maximizer arm or not at a certain time step. 
\begin{align}
    &\sum_{t \in \mathcal{T}^\mathrm{pr-xploit}} \mathbb{P} \left(\upsilon_t(\boldsymbol{\pi}_t) \neq \upsilon(\mathbf{c}(\boldsymbol{\theta}^0, \mathbf{s}^0)) \right) \notag \allowdisplaybreaks \\
    &= \sum_{t \in \mathcal{T}^\mathrm{pr-xploit}} \mathbb{P} \left(\upsilon_t(\boldsymbol{\pi}_t) \neq \upsilon(\mathbf{c}(\boldsymbol{\theta}^0, \mathbf{s}^0)) \Big| \upsilon_t(\boldsymbol{\pi}_t) \neq \argmax_{a \in \mathcal{A}} s^0_a + c_a(\widehat{\boldsymbol{\theta}}_t, \widehat{\mathbf{s}}^\mathrm{pr}_t) \right) p_t  \notag \allowdisplaybreaks \\
    &\hspace{2cm} + \mathbb{P} \left(\upsilon_t(\boldsymbol{\pi}_t) \neq \upsilon(\mathbf{c}(\boldsymbol{\theta}^0, \mathbf{s}^0)) \Big| \upsilon_t(\boldsymbol{\pi}_t) = \argmax_{a \in \mathcal{A}} s^0_a + c_a(\widehat{\boldsymbol{\theta}}_t, \widehat{\mathbf{s}}^\mathrm{pr}_t) \right) (1 - p_t) \allowdisplaybreaks \\
    &\leq \sum_{t \in \mathcal{T}^\mathrm{pr-xploit}}  p_t + \sum_{t \in \mathcal{T}^\mathrm{pr-xploit}} \mathbb{P} \left(\upsilon(\mathbf{c}(\boldsymbol{\theta}^0, \mathbf{s}^0)) \Big| \upsilon_t(\boldsymbol{\pi}_t) =  \argmax_{a \in \mathcal{A}} s^0_a + c_a(\widehat{\boldsymbol{\theta}}_t, \widehat{\mathbf{s}}^\mathrm{pr}_t) \right)  \allowdisplaybreaks \\
    &\leq \widetilde{k} + \sum_{t \in \mathcal{T}^\mathrm{pr-xploit}, t \geq \widetilde{k}} k \frac{\sqrt{\log 2t}}{\sqrt{t}}  + \sum_{t \in \mathcal{T}^\mathrm{pr-xploit}} \mathbb{P} \left(\upsilon_t(\boldsymbol{\pi}_t) \neq \upsilon(\mathbf{c}(\boldsymbol{\theta}^0, \mathbf{s}^0)) \Big| \upsilon_t(\boldsymbol{\pi}_t) =  \argmax_{a \in \mathcal{A}} s^0_a + c_a(\widehat{\boldsymbol{\theta}}_t, \widehat{\mathbf{s}}^\mathrm{pr}_t) \right)  \allowdisplaybreaks 
    \intertext{where the last inequality follows by Assumption \ref{assm:agent}.}
    &\leq \widetilde{k} + 2k \sqrt{|\mathcal{T}^{\mathrm{pr-xploit}}| \log 2|\mathcal{T}^{\mathrm{pr-xploit}}|} \notag \allowdisplaybreaks \\
    &\hspace{3cm} + \sum_{t \in \mathcal{T}^\mathrm{pr-xploit}} \mathbb{P} \left(\upsilon_t(\boldsymbol{\pi}_t) \neq \upsilon(\mathbf{c}(\boldsymbol{\theta}^0, \mathbf{s}^0)) \Big| \upsilon_t(\boldsymbol{\pi}_t) =  \argmax_{a \in \mathcal{A}} s^0_a + c_a(\widehat{\boldsymbol{\theta}}_t, \widehat{\mathbf{s}}^\mathrm{pr}_t) \right)  \allowdisplaybreaks \\
    &\leq \widetilde{k} + 2k \sqrt{T \log 2T} + \sum_{t \in \mathcal{T}^\mathrm{pr-xploit}} \mathbb{P} \left(\upsilon_t(\boldsymbol{\pi}_t) \neq \upsilon(\mathbf{c}(\boldsymbol{\theta}^0, \mathbf{s}^0)) \Big| \upsilon_t(\boldsymbol{\pi}_t) =  \argmax_{a \in \mathcal{A}} s^0_a + c_a(\widehat{\boldsymbol{\theta}}_t, \widehat{\mathbf{s}}^\mathrm{pr}_t) \right) \label{eq:sbs} \allowdisplaybreaks 
\end{align}
We now compute an upper bound for the summation in the last line above by recalling that $\upsilon(\mathbf{c}(\boldsymbol{\theta}^0, \mathbf{s}^0))= j^{*,0}$ as introduced in Section \ref{subsubsec:oracle}. 
\begin{align}
    &\sum_{t \in \mathcal{T}^\mathrm{pr-xploit}} \mathbb{P} \left(\upsilon_t(\boldsymbol{\pi}_t) \neq \upsilon(\mathbf{c}(\boldsymbol{\theta}^0, \mathbf{s}^0)) \Big| \upsilon_t(\boldsymbol{\pi}_t) =  \argmax_{a \in \mathcal{A}} s^0_a + c_a(\widehat{\boldsymbol{\theta}}_t, \widehat{\mathbf{s}}^\mathrm{pr}_t) \right) \notag \allowdisplaybreaks \\
    &= \sum_{t \in \mathcal{T}^\mathrm{pr-xploit}} \mathbb{P} \left(\argmax_{a \in \mathcal{A}} s^0_a + c_a(\widehat{\boldsymbol{\theta}}_t, \widehat{\mathbf{s}}^\mathrm{pr}_t) \neq j^{*,0} \right)  \allowdisplaybreaks \\
    &= \sum_{t \in \mathcal{T}^\mathrm{pr-xploit}}\mathbb{P} \left( \bigcup_{a \in \mathcal{A} \setminus \{j^{*,0}\}} s^0_{a} + c_{a}(\widehat{\boldsymbol{\theta}}_t, \widehat{\mathbf{s}}^\mathrm{pr}_t) > s^0_{j^{*,0}} + c_{j^{*,0}}(\widehat{\boldsymbol{\theta}}_t, \widehat{\mathbf{s}}^\mathrm{pr}_t) \right)  \allowdisplaybreaks \\
    &= \sum_{t \in \mathcal{T}^\mathrm{pr-xploit}} \mathbb{P} \left( \bigcup_{a \in \mathcal{A} \setminus \{j^{*,0}\}}  s^0_{a} + c_{a}(\widehat{\boldsymbol{\theta}}_t, \widehat{\mathbf{s}}^\mathrm{pr}_t) > s^0_{j^{*,0}} + c_{j^{*,0}}(\widehat{\boldsymbol{\theta}}_t, \widehat{\mathbf{s}}^\mathrm{pr}_t)  \Big | j^*_t = j^{*,0} \right) \mathbb{P} \left( j^*_t = j^{*,0} \right) \notag \allowdisplaybreaks \\
    &\hspace{2cm} + \mathbb{P} \left( \bigcup_{a \in \mathcal{A} \setminus \{j^{*,0}\}}  s^0_{a} + c_{a}(\widehat{\boldsymbol{\theta}}_t, \widehat{\mathbf{s}}^\mathrm{pr}_t) > s^0_{j^{*,0}} + c_{j^{*,0}}(\widehat{\boldsymbol{\theta}}_t, \widehat{\mathbf{s}}^\mathrm{pr}_t)  \Big | j^*_t \neq j^{*,0} \right) \mathbb{P} \left( j^*_t \neq j^{*,0}\right) \allowdisplaybreaks \\
    &\leq \sum_{t \in \mathcal{T}^\mathrm{pr-xploit}} \mathbb{P}  \left( \bigcup_{a \in \mathcal{A} \setminus \{j^{*,0}\}} s^0_{a} + c_{a}(\widehat{\boldsymbol{\theta}}_t, \widehat{\mathbf{s}}^\mathrm{pr}_t) > s^0_{j^{*,0}} + c_{j^{*,0}}(\widehat{\boldsymbol{\theta}}_t, \widehat{\mathbf{s}}^\mathrm{pr}_t)  \Big | j^*_t = j^{*,0} \right) \notag \allowdisplaybreaks \\
    &\quad + \sum_{t \in \mathcal{T}^\mathrm{pr-xploit}} \sum_{a \in \mathcal{A} \setminus \{j^{*,0}\}} \frac{n}{t} + \frac{n^{5/6} 2^{n+1}}{3^{n + 1} k \sqrt[6]{32}} \frac{1}{\sqrt{t \log 2t}}\allowdisplaybreaks \\
    &\leq \sum_{t \in \mathcal{T}^\mathrm{pr-xploit}}\mathbb{P} \left(\bigcup_{a \in \mathcal{A} \setminus \{j^{*,0}\}} s^0_{a} + c_{a}(\widehat{\boldsymbol{\theta}}_t, \widehat{\mathbf{s}}^\mathrm{pr}_t) > s^0_{j^{*,0}} + c_{j^{*,0}}(\widehat{\boldsymbol{\theta}}_t, \widehat{\mathbf{s}}^\mathrm{pr}_t)  \Big | j^*_t = j^{*,0} \right)\notag \allowdisplaybreaks \\
    &\quad + n^2 \log T + \frac{n^{11/6} 2^{n+2}}{3^{n + 1} k \sqrt[6]{32}} \sqrt{T} \label{eq:sbs2} \allowdisplaybreaks 
\end{align}
where second to the last line follows by Proposition \ref{prop:regret} and the last line follows by repeating the same arguments we had in lines (\ref{eq:firstpart-0})-(\ref{eq:firstpart-2}) at the first part of this proof. To bound the first term of the last inequality, we use the definition of the exploitation incentives $c_{a}(\widehat{\boldsymbol{\theta}}_t, \widehat{\mathbf{s}}^\mathrm{pr}_t)$ introduced in Algorithm \ref{alg:pr}.
\begin{align}
     &\sum_{t \in \mathcal{T}^\mathrm{pr-xploit}} \mathbb{P} \left( \bigcup_{a \in \mathcal{A} \setminus \{j^{*,0}\}} s^0_{a} + c_{a}(\widehat{\boldsymbol{\theta}}_t, \widehat{\mathbf{s}}^\mathrm{pr}_t) > s^0_{j^{*,0}} + c_{j^{*,0}}(\widehat{\boldsymbol{\theta}}_t, \widehat{\mathbf{s}}^\mathrm{pr}_t)  \Big | j^*_t = j^{*,0} \right) \notag \allowdisplaybreaks \\
     &= \sum_{t \in \mathcal{T}^\mathrm{pr-xploit}}  \mathbb{P} \left( \bigcup_{a \in \mathcal{A} \setminus \{j^*_t\}}  s^0_{a} + c_{a}(\widehat{\boldsymbol{\theta}}_t, \widehat{\mathbf{s}}^\mathrm{pr}_t) > s^0_{ j^*_t} + c_{ j^*_t}(\widehat{\boldsymbol{\theta}}_t, \widehat{\mathbf{s}}^\mathrm{pr}_t)  \right) \allowdisplaybreaks \\
     &= \sum_{t \in \mathcal{T}^\mathrm{pr-xploit}}  \mathbb{P} \left( \bigcup_{a \in \mathcal{A} \setminus \{j^*_t\}}  s^0_{a}  > s^0_{ j^*_t} + \max_{a' \in \mathcal{A}} \widehat{s}^\mathrm{pr}_{t,a} - \widehat{s}^\mathrm{pr}_{t,j^*_t} + 2\beta_t \right) \allowdisplaybreaks 
     \intertext{Recall the indices $\kappa_t \in \argmax_{a \in \mathcal{A}} \widehat{s}^\mathrm{pr}_{t,a} $, $\kappa^0 \in \argmax_{a \in \mathcal{A}} s^0_a $ defined earlier for notational convenience.}
     &\leq \sum_{t \in \mathcal{T}^\mathrm{pr-xploit}} \mathbb{P} \left( s^0_{\kappa^0} > s^0_{ j^*_t} +  \widehat{s}^\mathrm{pr}_{t,\kappa_t} - \widehat{s}^\mathrm{pr}_{t,j^*_t} + 2\beta_t  \right) \allowdisplaybreaks  \\
     &= \sum_{t \in \mathcal{T}^\mathrm{pr-xploit}} \mathbb{P} \left( (s^0_{\kappa^0} - \widehat{s}^\mathrm{pr}_{t,\kappa^0}) + (\widehat{s}^\mathrm{pr}_{t,\kappa^0}  - \widehat{s}^\mathrm{pr}_{t,\kappa_t}) + (\widehat{s}^\mathrm{pr}_{t,j^*_t} - s^0_{ j^*_t})  > 2\beta_t \right) \allowdisplaybreaks  
     \intertext{Notice that $\widehat{s}^\mathrm{pr}_{t,\kappa^0} - \widehat{s}^\mathrm{pr}_{t,\kappa_t} \leq 0$ by definition. Then,}
     &\leq \sum_{t \in \mathcal{T}^\mathrm{pr-xploit}} \mathbb{P} \left( 2 \max_{a \in \mathcal{A}}|s^0_{a}  - \widehat{s}^\mathrm{pr}_{t,a}| > 2\beta_t \right) \allowdisplaybreaks  \\
     &= \sum_{t \in \mathcal{T}^\mathrm{pr-xploit}}  \mathbb{P} \left( \|\mathbf{s}^0 - \widehat{\mathbf{s}}^\mathrm{pr}_t\|_\infty > \beta_t\right) \allowdisplaybreaks  \\
     &\leq  2 \sum_{t \in \mathcal{T}^\mathrm{pr-xploit}} \exp \left(-\frac{2\lambda_t^2}{(t-1)16n(6R_{\max} - 6R_{\min} + 2\gamma)^2} - \log \beta_t + n \log (2R_{\max} - 2R_{\min})\right) \allowdisplaybreaks  
     \intertext{which follows by Corollary \ref{cor:concen}. Then, we follow the same arguments as in (\ref{eq:lambda_proof_0})-(\ref{eq:lambda_proof_last}) and continue as}
     &\leq \frac{2^{n+1}}{3^{n + 1} k \sqrt[6]{32n}}   \sum_{t \in \mathcal{T}^\mathrm{pr-xploit}} \frac{1}{\sqrt{t \log 2t}}  \allowdisplaybreaks  \\
     &\leq \frac{2^{n+1}}{3^{n + 1} k \sqrt[6]{32n}}  \sqrt{|\mathcal{T}^\mathrm{pr-xploit}|}  \allowdisplaybreaks  \\
     &\leq \frac{2^{n+1}}{3^{n + 1} k \sqrt[6]{32n}}  \sqrt{T}
\end{align}
We substitute this upper bound into first (\ref{eq:sbs2}), then (\ref{eq:sbs}), and lastly (\ref{eq:stoc-regret4}). Then, taking the expectation of both sides of the obtained inequality in (\ref{eq:stoc-regret4}) gives us  
\begin{align}
   &\sum_{t \in \mathcal{T}^\mathrm{pr-xploit}} \theta^0_{\upsilon(\mathbf{c}(\boldsymbol{\theta}^0, \mathbf{s}^0))} -\theta^0_{\upsilon_t(\boldsymbol{\pi}_t)} \notag \allowdisplaybreaks \\
   &\leq \Theta^{\mathrm{\max}} \left( \widetilde{k} + 2k \sqrt{T \log 2T} +  \frac{2^{n+1}}{3^{n + 1} k \sqrt[6]{32n}}  \sqrt{T} + n^2 \log T + \frac{n^{11/6} 2^{n+2}}{3^{n + 1} k \sqrt[6]{32}} \sqrt{T} \right) \allowdisplaybreaks  \\
   &= \Theta^{\mathrm{\max}} \left( \widetilde{k} + 2k \sqrt{T \log 2T} +  \frac{2^{n+1} \left(2n^{11/6} + 1/\sqrt[6]{n}\right)}{3^{n + 1} k \sqrt[6]{32}}  \sqrt{T} + n^2 \log T  \right) 
\end{align}
This result together with (\ref{eq:secondpart-1}) gives us the upper bound for the second part of the regret. We conclude by combining everything with (\ref{eq:stoc-regret0}) and get the desired regret bound. 
\begin{align}
  \mathrm{Regret}\left(\Pi_{\epsilon, T} \right) &\leq \frac{12 B}{3 - w}T^{1 - w/3} \sqrt{\log 2T}  + m^\mathrm{pr} \left(n (\overline{C} - \underline{C}) + \Theta^{\max}\right) \left(\frac{2}{2w + 1} T^{w + 1/2} + \frac{2w - 1}{2w + 1}\right)\\
   & + 2k \Theta^{\mathrm{\max}} \sqrt{T \log 2T} + \frac{2^{n+1} \left(\Theta^{\mathrm{\max}} (2n^{11/6} + 1/\sqrt[6]{n}) + n^{5/6} (\overline{C} - \underline{C})\left(1 + 2n \right) \right)}{3^{n + 1} k \sqrt[6]{32}}  \sqrt{T} \\
   &+  n^2 \left(\overline{C} - \underline{C} + \Theta^{\mathrm{\max}}\right) \log T + \Theta^{\mathrm{\max}} \widetilde{k} 
\end{align}
\Halmos \endproof
\subsection{Results in Section \ref{sec:agent}} \label{appendix3}
\proof{\textbf{Proof of Lemma \ref{lem:validate1}.}}
According to the proposed principal's algorithm (\ref{alg:pr}) and agent's algorithm (\ref{alg:ag}), we know that the arm chosen by the agent at time $t \in \mathcal{T}^{\mathrm{ag-xploit}}$ is defined as $\upsilon_t(\boldsymbol{\pi}_t)  = \argmax_{a \in \mathcal{A}} \widehat{s}^\mathrm{ag}_{t,a} + \pi_{t, a} $ and the incentives provided by the principal at time $t \in \mathcal{T}^{\mathrm{pr-xploit}}$ is given by $\boldsymbol{\pi}_t =  \mathbf{c}(\widehat{\boldsymbol{\theta}}_t, \widehat{\mathbf{s}}^\mathrm{pr}_t)$. We start our proof by using these definitions. 
\begin{align}
    & \mathbb{P}\Big(\upsilon_t(\boldsymbol{\pi}_t) \neq \argmax_{a \in \mathcal{A}} s^0_a + \pi_{t, a}  \Big | t \in \mathcal{T}^{\mathrm{ag-xploit}} \cap  \mathcal{T}^{\mathrm{pr-xploit}} \Big) \nonumber \allowdisplaybreaks \\
    &= \mathbb{P}\left(\upsilon_t(\boldsymbol{\pi}_t) \neq \argmax_{a \in \mathcal{A}} s^0_a + \pi_{t, a} \Big | \upsilon_t(\boldsymbol{\pi}_t)  = \argmax_{a \in \mathcal{A}} \widehat{s}^\mathrm{ag}_{t,a} + \pi_{t, a},  \ \boldsymbol{\pi}_t =  \mathbf{c}(\widehat{\boldsymbol{\theta}}_t, \widehat{\mathbf{s}}^\mathrm{pr}_t) \right) \allowdisplaybreaks \\
    &= \mathbb{P}\left(\bigcup_{a' \in \mathcal{A}} \bigcup_{a \in \mathcal{A}} s^0_a + c_{t, a}(\widehat{\boldsymbol{\theta}}_t, \widehat{\mathbf{s}}^\mathrm{pr}_t) > s^0_{a'} + c_{t, a'}(\widehat{\boldsymbol{\theta}}_t, \widehat{\mathbf{s}}^\mathrm{pr}_t) \Big | a'  = \argmax_{a^{''} \in \mathcal{A}} \widehat{s}^\mathrm{ag}_{t,a^{''}} + c_{t, a^{''}}(\widehat{\boldsymbol{\theta}}_t, \widehat{\mathbf{s}}^\mathrm{pr}_t)  \right) \allowdisplaybreaks \\
    &\leq \sum_{a' \in \mathcal{A}} \sum_{a \in \mathcal{A}} \mathbb{P}\left(s^0_a + c_{t, a}(\widehat{\boldsymbol{\theta}}_t, \widehat{\mathbf{s}}^\mathrm{pr}_t) > s^0_{a'} + c_{t, a'}(\widehat{\boldsymbol{\theta}}_t, \widehat{\mathbf{s}}^\mathrm{pr}_t) \Big | a' = \argmax_{a^{''} \in \mathcal{A}} \widehat{s}^\mathrm{ag}_{t,a^{''}} + c_{t, a^{''}}(\widehat{\boldsymbol{\theta}}_t, \widehat{\mathbf{s}}^\mathrm{pr}_t) \right) \allowdisplaybreaks \label{eq:pt-union}
\end{align}
where (\ref{eq:pt-union}) follows by the Boole's inequality (a.k.a. union bound). We further condition on whether the chosen arm $\upsilon_t(\boldsymbol{\pi}_t)$ is same as the desired arm by the principal ($j^*_t$) or not.
\begin{align}
    &= \sum_{a' \in \mathcal{A}} \sum_{a \in \mathcal{A}} \mathbb{P}\Big(c_{t, a}(\widehat{\boldsymbol{\theta}}_t, \widehat{\mathbf{s}}^\mathrm{pr}_t) - c_{t, a'}(\widehat{\boldsymbol{\theta}}_t, \widehat{\mathbf{s}}^\mathrm{pr}_t) > s^0_{a'} - s^0_a \Big | j^*_t = a', a' = \argmax_{a^{''} \in \mathcal{A}} \widehat{s}^\mathrm{ag}_{t,a^{''}} + c_{t, a^{''}}(\widehat{\boldsymbol{\theta}}_t, \widehat{\mathbf{s}}^\mathrm{pr}_t) \Big) \nonumber \allowdisplaybreaks \\
    &\hspace{2cm} \cdot \mathbb{P}\Big(j^*_t = a' \Big | a' = \argmax_{a^{''} \in \mathcal{A}} \widehat{s}^\mathrm{ag}_{t,a^{''}} + c_{t, a^{''}}(\widehat{\boldsymbol{\theta}}_t, \widehat{\mathbf{s}}^\mathrm{pr}_t)  \Big) \nonumber \allowdisplaybreaks \\
    &\hspace{1.5cm} + \mathbb{P}\Big(c_{t, a}(\widehat{\boldsymbol{\theta}}_t, \widehat{\mathbf{s}}^\mathrm{pr}_t) - c_{t, a'}(\widehat{\boldsymbol{\theta}}_t, \widehat{\mathbf{s}}^\mathrm{pr}_t) > s^0_{a'} - s^0_a \Big | j^*_t \neq a', a' = \argmax_{a^{''} \in \mathcal{A}} \widehat{s}^\mathrm{ag}_{t,a^{''}} + c_{t, a^{''}}(\widehat{\boldsymbol{\theta}}_t, \widehat{\mathbf{s}}^\mathrm{pr}_t) \Big) \nonumber \allowdisplaybreaks \\
    &\hspace{2cm} \cdot \mathbb{P}\Big(j^*_t \neq a' \Big | a' = \argmax_{a^{''} \in \mathcal{A}} \widehat{s}^\mathrm{ag}_{t,a^{''}} + c_{t, a^{''}}(\widehat{\boldsymbol{\theta}}_t, \widehat{\mathbf{s}}^\mathrm{pr}_t) \Big) \allowdisplaybreaks \\
    &\leq \sum_{a' \in \mathcal{A}} \sum_{a \in \mathcal{A}} \mathbb{P}\big(c_{t, a}(\widehat{\boldsymbol{\theta}}_t, \widehat{\mathbf{s}}^\mathrm{pr}_t) - c_{t, j^*_t}(\widehat{\boldsymbol{\theta}}_t, \widehat{\mathbf{s}}^\mathrm{pr}_t) > s^0_{j^*_t} - s^0_a\big) + \mathbb{P}\Big(j^*_t \neq a' \Big | a' = \argmax_{a^{''} \in \mathcal{A}} \widehat{s}^\mathrm{ag}_{t,a^{''}} + c_{t, a^{''}}(\widehat{\boldsymbol{\theta}}_t, \widehat{\mathbf{s}}^\mathrm{pr}_t) \Big)  \allowdisplaybreaks \label{eq:pt-prop-1}
\end{align}
We can bound the first term in (\ref{eq:pt-prop-1}) from above by using the definition of the principal's exploitation incentives $\mathbf{c}(\widehat{\boldsymbol{\theta}}_t, \widehat{\mathbf{s}}^\mathrm{pr}_t)$ introduced in Algorithm \ref{alg:pr}. 
\begin{align}
     \sum_{a' \in \mathcal{A}} \sum_{a \in \mathcal{A}} \mathbb{P}\big(c_{t, a}(\widehat{\boldsymbol{\theta}}_t, \widehat{\mathbf{s}}^\mathrm{pr}_t) - c_{t, j^*_t}(\widehat{\boldsymbol{\theta}}_t, \widehat{\mathbf{s}}^\mathrm{pr}_t) > s^0_{j^*_t} - s^0_a\big)  &= \sum_{a' \in \mathcal{A}} \sum_{a \in \mathcal{A}}  \mathbb{P}\big( \widehat{s}^\mathrm{pr}_{t,j^*_t} - s^0_{j^*_t} + s^0_a - 2\beta_t  >  \max_{a'' \in \mathcal{A}} \widehat{s}^\mathrm{pr}_{t,a''}\big) \\
     &\leq \sum_{a' \in \mathcal{A}} \sum_{a \in \mathcal{A}}  \mathbb{P}\big( \widehat{s}^\mathrm{pr}_{t,j^*_t} - s^0_{j^*_t} + s^0_a - 2\beta_t  >  \widehat{s}^\mathrm{pr}_{t,a}\big) \\
     &= n \sum_{a \in \mathcal{A}}  \mathbb{P}\big( \widehat{s}^\mathrm{pr}_{t,j^*_t} - s^0_{j^*_t} + s^0_a - \widehat{s}^\mathrm{pr}_{t,a} > 2\beta_t\big) \label{eq:conditionAandB}
\end{align}
We continue by conditioning the last probability on the intersection of the two events $A = \{\widehat{s}^\mathrm{pr}_{t,j^*_t} - s^0_{j^*_t} \leq \beta_t\}$ and $B = \{s^0_{a} - \widehat{s}^\mathrm{pr}_{t,a} \leq \beta_t\}$. Recall that the complement of $A \cap B$ is equal to the union of their complements $\overline{A} \cup \overline{B}$.
\begin{align}
     (\ref{eq:conditionAandB}) &= n \sum_{a \in \mathcal{A}}  \mathbb{P}\big(\widehat{s}^\mathrm{pr}_{t,j^*_t} - s^0_{j^*_t} + s^0_a - \widehat{s}^\mathrm{pr}_{t,a} > 2\beta_t  \big | A \cap B\big) \mathbb{P}\big(A \cap B\big) \notag \allowdisplaybreaks \\
     &\quad + n \sum_{a \in \mathcal{A}} \mathbb{P}\big( \widehat{s}^\mathrm{pr}_{t,j^*_t} - s^0_{j^*_t} + s^0_a - \widehat{s}^\mathrm{pr}_{t,a} > 2\beta_t  \big |\overline{A} \cup \overline{B}\big) \mathbb{P}\big(\overline{A} \cup \overline{B} \big) \\
     &\leq n \sum_{a \in \mathcal{A}} \mathbb{P}\big(\overline{A} \cup \overline{B} \big) \allowdisplaybreaks \\
     &\leq n \sum_{a \in \mathcal{A}}  \mathbb{P}\big(\widehat{s}^\mathrm{pr}_{t,j^*_t} - s^0_{j^*_t} > \beta_t \big) + \mathbb{P}\big(\widehat{s}^\mathrm{pr}_{t,a} - s^0_{a} < -\beta_t\big)  \allowdisplaybreaks \\
     &\leq 2n^2 \mathbb{P}\big(\|\widehat{\mathbf{s}}^\mathrm{pr}_t - \mathbf{s}^0\|_\infty > \beta_t \big)   \allowdisplaybreaks \\
     &\leq 4n^2 \exp \left(-\frac{2\lambda_t^2}{(t-1)16n(6R_{\max} - 6R_{\min} + 2\gamma)^2} - \log \beta_t + n \log (2R_{\max} - 2R_{\min})\right)
\end{align}
where the last result follows by Corollary \ref{cor:concen}. Next, we bound the second term in (\ref{eq:pt-prop-1}). 
\begin{align}
    &\sum_{a' \in \mathcal{A}} \sum_{a \in \mathcal{A}} \mathbb{P}\Big(j^*_t \neq a' \Big | a' = \argmax_{a^{''} \in \mathcal{A}} \widehat{s}^\mathrm{ag}_{t,a^{''}} + c_{t, a^{''}}(\widehat{\boldsymbol{\theta}}_t, \widehat{\mathbf{s}}^\mathrm{pr}_t)  \Big) \nonumber \allowdisplaybreaks \\
    &\leq n^2 \mathbb{P}\Big(j^*_t \neq \argmax_{a^{''} \in \mathcal{A}} \widehat{s}^\mathrm{ag}_{t,a^{''}} + c_{t, a^{''}}(\widehat{\boldsymbol{\theta}}_t, \widehat{\mathbf{s}}^\mathrm{pr}_t) \Big) \allowdisplaybreaks \\ 
    &= n^2 \mathbb{P}\Big(\bigcup_{a \in \mathcal{A}} \widehat{s}^\mathrm{ag}_{t,a}+ c_{t, a}(\widehat{\boldsymbol{\theta}}_t, \widehat{\mathbf{s}}^\mathrm{pr}_t) > \widehat{s}^\mathrm{ag}_{t,j^*_t} + c_{t, j^*_t}(\widehat{\boldsymbol{\theta}}_t, \widehat{\mathbf{s}}^\mathrm{pr}_t)\Big) \label{eq:unionnn}\allowdisplaybreaks \\ 
    &\leq n^2 \sum_{a \in \mathcal{A}} \mathbb{P}\Big(\widehat{s}^\mathrm{ag}_{t,a} + c_{t, a}(\widehat{\boldsymbol{\theta}}_t, \widehat{\mathbf{s}}^\mathrm{pr}_t) >\widehat{s}^\mathrm{ag}_{t,j^*_t} + c_{t, j^*_t}(\widehat{\boldsymbol{\theta}}_t, \widehat{\mathbf{s}}^\mathrm{pr}_t)\Big) \allowdisplaybreaks \\ 
    &= n^2 \sum_{a \in \mathcal{A}} \mathbb{P}\Big(\widehat{s}^\mathrm{pr}_{t,j^*_t} + \widehat{s}^\mathrm{ag}_{t,a} - \widehat{s}^\mathrm{ag}_{t,j^*_t} - 2\beta_t > \max_{a' \in \mathcal{A}} \widehat{s}^\mathrm{pr}_{t,a'} \Big) \allowdisplaybreaks \label{eq:pt-bydefofincentives}\\ 
    &\leq n^2 \sum_{a \in \mathcal{A}} \mathbb{P}\Big(\widehat{s}^\mathrm{pr}_{t,j^*_t} + \widehat{s}^\mathrm{ag}_{t,a} - \widehat{s}^\mathrm{ag}_{t,j^*_t} - 2\beta_t > \widehat{s}^\mathrm{pr}_{t,a} \Big) \allowdisplaybreaks \\ 
    &= n^2 \sum_{a \in \mathcal{A}} \mathbb{P}\Big(\widehat{s}^\mathrm{pr}_{t,j^*_t} + \widehat{s}^\mathrm{ag}_{t,a} - \widehat{s}^\mathrm{ag}_{t,j^*_t} - \widehat{s}^\mathrm{pr}_{t,a} + s^0_a - s^0_a + s^0_{j^*_t} - s^0_{j^*_t} > 2\beta_t \Big) \allowdisplaybreaks \\ 
    &= n^2 \sum_{a \in \mathcal{A}} \mathbb{P}\Big(\big( \widehat{s}^\mathrm{pr}_{t,j^*_t} - s^0_{j^*_t} \big) + \big(\widehat{s}^\mathrm{ag}_{t,a}- s^0_a\big) + \big(s^0_{j^*_t} - \widehat{s}^\mathrm{ag}_{t,j^*_t}\big) + \big(s^0_a - \widehat{s}^\mathrm{pr}_{t,a}\big) > 2\beta_t \Big) \allowdisplaybreaks \label{eq:conditionCandD}
\end{align}
where (\ref{eq:unionnn}) follows by the Boole's inequality (a.k.a. union bound) and (\ref{eq:pt-bydefofincentives}) follows by definition of $\mathbf{c}(\widehat{\boldsymbol{\theta}}_t, \widehat{\mathbf{s}}^\mathrm{pr}_t)$ as in Algorithm \ref{alg:pr}. We next condition the last probability on the intersection of the two events $C = \{\widehat{s}^\mathrm{pr}_{t,j^*_t} - s^0_{j^*_t} \leq \beta_t/2, \ \widehat{s}^\mathrm{ag}_{t,a}- s^0_a \leq \beta_t/2\}$ and $D = \{s^0_{j^*_t} - \widehat{s}^\mathrm{ag}_{t,j^*_t} \leq \beta_t/2, \ s^0_a - \widehat{s}^\mathrm{pr}_{t,a} \leq \beta_t/2\}$. 
\begin{align}
    (\ref{eq:conditionCandD}) &= n^2 \sum_{a \in \mathcal{A}} \mathbb{P}\Big(\big( \widehat{s}^\mathrm{pr}_{t,j^*_t} - s^0_{j^*_t} \big) + \big(\widehat{s}^\mathrm{ag}_{t,a}- s^0_a\big) + \big(s^0_{j^*_t} - \widehat{s}^\mathrm{ag}_{t,j^*_t}\big) + \big(s^0_a - \widehat{s}^\mathrm{pr}_{t,a}\big) > 2\beta_t \Big | C \cap D\Big) \mathbb{P}\Big(C \cap D\Big) \notag \allowdisplaybreaks \\
    &\quad + \mathbb{P}\Big(\big( \widehat{s}^\mathrm{pr}_{t,j^*_t} - s^0_{j^*_t} \big) + \big(\widehat{s}^\mathrm{ag}_{t,a}- s^0_a\big) + \big(s^0_{j^*_t} - \widehat{s}^\mathrm{ag}_{t,j^*_t}\big) + \big(s^0_a - \widehat{s}^\mathrm{pr}_{t,a}\big) > 2\beta_t  \Big | \overline{C} \cup \overline{D}\Big) \mathbb{P}\Big(\overline{C} \cup \overline{D}\Big) \allowdisplaybreaks \\ 
    &\leq n^2 \sum_{a \in \mathcal{A}} \mathbb{P}\big(\overline{C} \cup \overline{D}\big)\allowdisplaybreaks \\ 
    &\leq n^2 \sum_{a \in \mathcal{A}}  \mathbb{P}\big(\widehat{s}^\mathrm{pr}_{t,j^*_t} - s^0_{j^*_t} > \beta_t/2 \big) + \mathbb{P}\big( \widehat{s}^\mathrm{pr}_{t,a} - s^0_a < - \beta_t/2 \big) \notag \allowdisplaybreaks \\
    &\quad + n^2 \sum_{a \in \mathcal{A}} \mathbb{P}\big( \widehat{s}^\mathrm{ag}_{t,a}- s^0_a > \beta_t/2 \big) + \mathbb{P}\big( \widehat{s}^\mathrm{ag}_{t,j^*_t} - s^0_{j^*_t} < -\beta_t/2 \big) \allowdisplaybreaks \\ 
    &\leq 2n^2 \mathbb{P}\big(\|\widehat{\mathbf{s}}^\mathrm{pr}_t - \mathbf{s}^0\|_\infty > \beta_t/2 \big) + n^2 \sum_{a \in \mathcal{A}} \mathbb{P}\big( \widehat{s}^\mathrm{ag}_{t,a}- s^0_a > \beta_t/2 \big) + \mathbb{P}\big( \widehat{s}^\mathrm{ag}_{t,j^*_t} - s^0_{j^*_t} < -\beta_t/2 \big) \allowdisplaybreaks \\ 
    &\leq 4n^2\exp \left(-\frac{2\lambda_t^2}{(t-1)16n(6R_{\max} - 6R_{\min} + 2\gamma)^2} - \log \frac{\beta_t}{2} + n \log(2R_{\max} - 2R_{\min})\right) \notag \allowdisplaybreaks \\
    &\quad + n^2 \sum_{a \in \mathcal{A}}  \mathbb{P}\big( \widehat{s}^\mathrm{ag}_{t,a}- s^0_a > \beta_t/2 \big) + \mathbb{P}\big( \widehat{s}^\mathrm{ag}_{t,j^*_t} - s^0_{j^*_t} < -\beta_t/2  \big) \allowdisplaybreaks \label{eq:pt-prop-2}
\end{align}
where the last inequality follows by Corollary \ref{cor:concen}. We can bound the first probability term in (\ref{eq:pt-prop-2}) by recalling that $s^0_{a} = \mathbb{E} \widehat{s}_{t,a}^{\mathrm{ag}}$ by definition as given in (\ref{def:s-hat-agent}). Further, we observe that $T(a, t) \geq T^{\mathrm{ag-xplore}}(a, t)$ for any $a \in \mathcal{A}$, and obtain 
\begin{align}
    n^2 \sum_{a \in \mathcal{A}} \mathbb{P}\big(\widehat{s}^\mathrm{ag}_{t,a}-s^0_{a} > \beta_t/2 \big) &= n^2 \sum_{a \in \mathcal{A}} \mathbb{P}\left(\widehat{s}^\mathrm{ag}_{t,a}-s^0_{a} > \beta_t/2 \Big| T^{\mathrm{ag-xplore}}(a, t) > \mathbb{E} T^{\mathrm{ag-xplore}}(a, t) -  \frac{2(m^{\mathrm{ag}})^2}{n}\right) \notag \allowdisplaybreaks \\
    &\qquad \qquad \cdot \mathbb{P}\left(T^{\mathrm{ag-xplore}}(a, t) > \mathbb{E} T^{\mathrm{ag-xplore}}(a, t) - \frac{2(m^{\mathrm{ag}})^2}{n} \right) \nonumber \allowdisplaybreaks \\
    &\quad + n^2 \sum_{a \in \mathcal{A}} \mathbb{P}\left(\widehat{s}^\mathrm{ag}_{t,a}-s^0_{a} > \beta_t/2 \Big| T^{\mathrm{ag-xplore}}(a, t) \leq \mathbb{E} T^{\mathrm{ag-xplore}}(a, t) - \frac{2(m^{\mathrm{ag}})^2}{n}\right)  \notag \allowdisplaybreaks \\
    &\qquad \qquad \cdot \mathbb{P}\left(T^{\mathrm{ag-xplore}}(a, t) \leq \mathbb{E} T^{\mathrm{ag-xplore}}(a, t) - \frac{2(m^{\mathrm{ag}})^2}{n}\right) \allowdisplaybreaks \label{eq:agent-hoeffding-1}\\
    &\leq n^2 \sum_{a \in \mathcal{A}} \mathbb{P}\left(\widehat{s}^\mathrm{ag}_{t,a}-s^0_{a} > \beta_t/2 \Big| T^{\mathrm{ag-xplore}}(a, t) > \mathbb{E} T^{\mathrm{ag-xplore}}(a, t) - \frac{2(m^{\mathrm{ag}})^2}{n}\right) \nonumber \allowdisplaybreaks \\
    &\quad + n^2 \sum_{a \in \mathcal{A}} \mathbb{P}\left(T^{\mathrm{ag-xplore}}(a, t) \leq \mathbb{E} T^{\mathrm{ag-xplore}}(a, t) - \frac{2(m^{\mathrm{ag}})^2}{n}\right) \allowdisplaybreaks \label{eq:hoeffding}
\end{align}
Now, we can use Hoeffding's Inequality \citep{boucheron2013concentration} to bound the the first probability term in the last inequality above. For the second probability term, we notice that by construction of Algorithm \ref{alg:ag}, we have $T^{\mathrm{ag-xplore}}(a, t) = \sum_{\tau = 1}^{t-1} \mathbf{1}(\upsilon_\tau(\boldsymbol{\pi}_\tau) = a)$ where indicator variables $\mathbf{1}(\upsilon_\tau(\boldsymbol{\pi}_\tau) = a)$'s are defined as independent Bernoulli random variables with success probabilities $\epsilon^\mathrm{ag}_\tau/n$. Hence, Hoeffding's Inequality can also be used for the second term. 
\begin{align}
    (\ref{eq:hoeffding}) &\leq n^2 \sum_{a \in \mathcal{A}} \exp\left(- \frac{\left(\mathbb{E} T^{\mathrm{ag-xplore}}(a, t) - 2(m^{\mathrm{ag}})^2/n\right) \beta^2_{t}}{8(R_{\max} - R_{\min})^2}\right) + n^2 \sum_{a \in \mathcal{A}} \exp \left(- \frac{4(m^{\mathrm{ag}})^4}{n^2} (t-1) \right) \allowdisplaybreaks \\
    &\leq n^2 \sum_{a \in \mathcal{A}} \exp\left(- \frac{\left(\mathbb{E} T^{\mathrm{ag-xplore}}(a, t) - 2(m^{\mathrm{ag}})^2/n\right) \beta^2_{t}}{8(R_{\max} - R_{\min})^2}\right) + \frac{n^3}{t-1}\label{eq:pt-prop-3}
\end{align}
Now, we need to compute a lower bound for $\mathbb{E} T^{\mathrm{ag-xplore}}(a, t)$. 
\begin{align}
    \mathbb{E} T^{\mathrm{ag-xplore}}(a, t)  = \sum_{\tau = 1}^{t-1} \frac{\epsilon^\mathrm{ag}_\tau}{n} =  \frac{1}{n} \sum_{\tau = 1}^{t-1} \min \left\{1, \frac{m^{\mathrm{ag}}}{\sqrt{\tau}}\right\} \geq \frac{m^{\mathrm{ag}}}{n}\sum_{\tau = (m^{\mathrm{ag}})^2}^{t-1} \frac{1}{\sqrt{\tau}} &\geq  \frac{m^{\mathrm{ag}}}{n} \int_{(m^{\mathrm{ag}})^2}^{t} \frac{1}{\sqrt{\tau}} d\tau  \allowdisplaybreaks \\ 
    &= \frac{2m^{\mathrm{ag}}}{n} (\sqrt{t} - m^{\mathrm{ag}}) \allowdisplaybreaks
\end{align}
Using this lower bound on $\mathbb{E} T^{\mathrm{ag-xplore}}(a, t)$, we obtain 
\begin{align}
    (\ref{eq:pt-prop-3}) &\leq n^3 \exp\left( \frac{-\frac{m^{\mathrm{ag}}}{n}\sqrt{t}\beta_t^2 + \frac{2(m^{\mathrm{ag}})^2}{n} \beta_t^2}{4(R_{\max} - R_{\min})^2}\right) + \frac{n^3}{t-1} \label{eq:agent-hoeffding-last}
    \intertext{Recall $\beta_t = B \frac{\sqrt{\log 2t}}{t^{w/3}}$ with $B = \frac{3 k \left(3 (R_{\max} - R_{\min}) + \gamma\right)^n \sqrt[6]{32n}}{1 - k \sqrt{\log 2\widetilde{k}}/ \sqrt{\widetilde{k}}}$. Then,}
    &= n^3 \exp\left( \frac{-\frac{m^{\mathrm{ag}}}{n} B^2 \sqrt{t} \dfrac{\log 2t}{t^{2w/3}} + \frac{2(m^{\mathrm{ag}})^2}{n} B^2 \dfrac{\log 2t}{t^{2w/3}}}{4(R_{\max} - R_{\min})^2}\right)  + \frac{n^3}{t-1} \\
    &\leq n^3 \exp\left( -\sqrt{t} \dfrac{\log 2t}{t^{2w/3}}  + \frac{2(m^{\mathrm{ag}})^2B^2}{4n(R_{\max} - R_{\min})^2}\right)  + \frac{n^3}{t-1} \\ 
    &\leq \frac{n^3 \left(\exp\Big(\frac{2(m^{\mathrm{ag}})^2B^2}{4n(R_{\max} - R_{\min})^2}\Big) + 1\right)}{t-1} 
\end{align} 
where the second to the last inequality holds for all $t \geq 2$. This gives us 
\begin{align}
     n^2 \sum_{a \in \mathcal{A}} \mathbb{P}\big(\widehat{s}^\mathrm{ag}_{t,a}-s^0_{a} > \beta_t/2 \big) \leq \frac{n^3 \left(\exp\Big(\frac{2(m^{\mathrm{ag}})^2B^2}{4n(R_{\max} - R_{\min})^2}\Big) + 1\right)}{t-1} 
\end{align}
Similarly, we have 
\begin{equation}
      n^2 \sum_{a \in \mathcal{A}} \mathbb{P}\big( \widehat{s}^\mathrm{ag}_{t,j^*_t} - s^0_{j^*_t} < -\beta_t/2  \big) \leq \frac{n^3 \left(\exp\Big(\frac{2(m^{\mathrm{ag}})^2B^2}{4n(R_{\max} - R_{\min})^2}\Big) + 1\right)}{t-1} 
\end{equation}
Substituting these upper bounds into (\ref{eq:pt-prop-2}) and combining everything, we conclude as
\begin{align}
    &\mathbb{P}\Big(\upsilon_t(\boldsymbol{\pi}_t) \neq \argmax_{a \in \mathcal{A}} s^0_a + \pi_{t, a}  \Big | t \in \mathcal{T}^{\mathrm{ag-xploit}} \cap  \mathcal{T}^{\mathrm{pr-xploit}} \Big)  \nonumber \\
    &\leq \frac{2n^3 \left(\exp\Big(\frac{2(m^{\mathrm{ag}})^2B^2}{4n(R_{\max} - R_{\min})^2}\Big) + 1\right)}{t-1}  \notag \allowdisplaybreaks \\
    & \quad + 8n^2\exp \left(-\frac{2\lambda_t^2}{(t-1)16n(6R_{\max} - 6R_{\min} + 2\gamma)^2} - \log \frac{\beta_t}{2} + n \log(2R_{\max} - 2R_{\min})\right) \allowdisplaybreaks \label{eq:pt-prop-4}
\end{align}
\Halmos \endproof

\proof{\textbf{Proof of Lemma \ref{lem:validate2}.}}
By construction of the principal's algorithm (\ref{alg:pr}) and agent's algorithm (\ref{alg:ag}), at any time point $t \in \mathcal{T}^{\mathrm{ag-xploit}} \cap \mathcal{T}^{\mathrm{pr-xplore}}$, we know that  the arm chosen by the agent is given as $\upsilon_t(\boldsymbol{\pi}_t)  = \argmax_{a \in \mathcal{A}} \widehat{s}^\mathrm{ag}_{t,a} + \pi_{t, a}$ and the incentives provided by the principal are uniformly randomly selected from set $\mathcal{C}$. Then, we start with 
\begin{align}
    &\mathbb{P}\left(\upsilon_t(\boldsymbol{\pi}_t) \neq \argmax_{a \in \mathcal{A}} s^0_a + \pi_{t, a}  \Big | t \in \mathcal{T}^{\mathrm{ag-xploit}} \cap \mathcal{T}^{\mathrm{pr-xplore}}\right)  \notag \\
    &=\mathbb{P}\left(\upsilon_t(\boldsymbol{\pi}_t) \neq \argmax_{a \in \mathcal{A}}  s^0_a + \pi_{t, a} \Big |  \upsilon_t(\boldsymbol{\pi}_t) = \argmax_{a \in \mathcal{A}} \widehat{s}^\mathrm{ag}_{t,a} + \pi_{t, a} , \ \pi_{t, a} \sim \mathcal{U}(\underline{C}, \overline{C}) \  \forall a \in \mathcal{A} \right) \allowdisplaybreaks \\
    &= \mathbb{P}\left(\bigcup_{a'' \in \mathcal{A}} \bigcup_{a' \in \mathcal{A}} s^0_{a''} + \pi_{t, a''} < s^0_{a'} + \pi_{t, a'}  \Big |  \widehat{s}^\mathrm{ag}_{t,a} + \pi_{t, a} \leq \widehat{s}^\mathrm{ag}_{t,a''} + \pi_{t, a''}  \  \forall a \in \mathcal{A}, \ \pi_{t, a} \sim \mathcal{U}(\underline{C}, \overline{C}) \ \forall a \in \mathcal{A} \right) \allowdisplaybreaks \\
    &\leq \sum_{a'' \in \mathcal{A}} \sum_{a' \in \mathcal{A}} \mathbb{P}\left(s^0_{a''} + \pi_{t, a''} < s^0_{a'} + \pi_{t, a'}  \Big| \widehat{s}^\mathrm{ag}_{t,a} + \pi_{t, a} \leq \widehat{s}^\mathrm{ag}_{t,a''} + \pi_{t, a''}  \  \forall a \in \mathcal{A}, \ \pi_{t, a} \sim \mathcal{U}(\underline{C}, \overline{C}) \ \forall a \in \mathcal{A} \right) \allowdisplaybreaks  \label{eq:union} \\
    &\leq \sum_{a'' \in \mathcal{A}} \sum_{a' \in \mathcal{A}} \mathbb{P}\left(\widehat{s}^\mathrm{ag}_{t,a'} - \widehat{s}^\mathrm{ag}_{t,a''}  \leq \pi_{t, a''} - \pi_{t, a'}  < s^0_{a'} - s^0_{a''} \Big | \pi_{t, a} \sim \mathcal{U}(\underline{C}, \overline{C}), \forall a \in \mathcal{A} \right) \label{eq:validate2-0}
\end{align}
where (\ref{eq:union}) follows by the Boole's inequality (a.k.a. union bound). Now, we recall that $s^0_{a} = \mathbb{E} \widehat{s}_{t,a}^{\mathrm{ag}}$ for any $a \in \mathcal{A}$ by definition given in (\ref{def:s-hat-agent}) and continue our derivation by conditioning on the intersection of the two events $E = \{\widehat{s}^\mathrm{ag}_{t,a''} - \mathbb{E}\widehat{s}^\mathrm{ag}_{t,a''} < \varphi_t \} $ and $F = \{\mathbb{E}\widehat{s}^\mathrm{ag}_{t,a'} - \widehat{s}^\mathrm{ag}_{t,a'} < \varphi_t\}$ for some $\varphi_t >0$.
\begin{align}
    &= \sum_{a'' \in \mathcal{A}} \sum_{a' \in \mathcal{A}} \mathbb{P}\left(\widehat{s}^\mathrm{ag}_{t,a'} - \widehat{s}^\mathrm{ag}_{t,a''}  \leq \pi_{t, a''} - \pi_{t, a'}  < \mathbb{E} \widehat{s}^\mathrm{ag}_{t,a'} - \mathbb{E}\widehat{s}^\mathrm{ag}_{t,a''}  \Big | \pi_{t, a} \sim \mathcal{U}(\underline{C}, \overline{C}), \forall a \in \mathcal{A},  E \cap F \right)\mathbb{P}(E \cap F) \notag \allowdisplaybreaks \\
    &\quad + \sum_{a'' \in \mathcal{A}} \sum_{a' \in \mathcal{A}} \mathbb{P}\left(\widehat{s}^\mathrm{ag}_{t,a'} - \widehat{s}^\mathrm{ag}_{t,a''}  \leq \pi_{t, a''} - \pi_{t, a'}  < \mathbb{E} \widehat{s}^\mathrm{ag}_{t,a'} - \mathbb{E}\widehat{s}^\mathrm{ag}_{t,a''} \Big | \pi_{t, a} \sim \mathcal{U}(\underline{C}, \overline{C}), \forall a \in \mathcal{A}, \overline{E} \cup \overline{F}  \right)\mathbb{P}(\overline{E} \cup \overline{F} ) \allowdisplaybreaks \\ 
    &\leq  \sum_{a'' \in \mathcal{A}} \sum_{a' \in \mathcal{A}} \mathbb{P}\left(\widehat{s}^\mathrm{ag}_{t,a'} - \widehat{s}^\mathrm{ag}_{t,a''}  \leq \pi_{t, a''} - \pi_{t, a'} < \mathbb{E} \widehat{s}^\mathrm{ag}_{t,a'} - \mathbb{E}\widehat{s}^\mathrm{ag}_{t,a''} \Big | \pi_{t, a} \sim \mathcal{U}(\underline{C}, \overline{C}), \forall a \in \mathcal{A},  E \cap F \right) \notag \allowdisplaybreaks \\
    &\quad + \sum_{a'' \in \mathcal{A}} \sum_{a' \in \mathcal{A}} \mathbb{P}(\overline{E})  + \mathbb{P}(\overline{F})  \allowdisplaybreaks  \label{eq:validate2-1}
\end{align}
For the first term in (\ref{eq:validate2-1}), we recall that the difference of the two Uniform random variables, $\pi_{t, a''} - \pi_{t, a'}$, follows a triangular distribution whose cdf is derived in (\ref{eq:cdf}). Because the cdf is in the form of a piecewise function, we can compute an upper bound to the first term above by considering the case with the highest probability, that is $\widehat{s}^\mathrm{ag}_{t,a'} - \widehat{s}^\mathrm{ag}_{t,a''}  <0$ and $\mathbb{E} \widehat{s}^\mathrm{ag}_{t,a'} - \mathbb{E}\widehat{s}^\mathrm{ag}_{t,a''} \geq 0$ for any $a', a'' \in \mathcal{A}$. Then, it follows that 
\begin{align}
    &\sum_{a'' \in \mathcal{A}} \sum_{a' \in \mathcal{A}} \mathbb{P}\left(\widehat{s}^\mathrm{ag}_{t,a'} - \widehat{s}^\mathrm{ag}_{t,a''}  \leq \pi_{t, a''} - \pi_{t, a'}  < \mathbb{E} \widehat{s}^\mathrm{ag}_{t,a'} - \mathbb{E}\widehat{s}^\mathrm{ag}_{t,a''} \Big | \pi_{t, a} \sim \mathcal{U}(\underline{C}, \overline{C}), \forall a \in \mathcal{A},  E \cap F \right) \notag \allowdisplaybreaks \\ 
    &\leq \sum_{a'' \in \mathcal{A}} \sum_{a' \in \mathcal{A}} \left[ 1 - \frac{(\mathbb{E} \widehat{s}^\mathrm{ag}_{t,a'} - \mathbb{E}\widehat{s}^\mathrm{ag}_{t,a''} + \underline{C} - \overline{C})^2}{2 (\overline{C} - \underline{C})^2} - \frac{(\widehat{s}^\mathrm{ag}_{t,a'} - \widehat{s}^\mathrm{ag}_{t,a''} + \overline{C} - \underline{C})^2}{2 (\overline{C} - \underline{C})^2} \Big | E \cap F \right ]\\
    &= \sum_{a'' \in \mathcal{A}} \sum_{a' \in \mathcal{A}} \left[ \frac{-(\mathbb{E} \widehat{s}^\mathrm{ag}_{t,a'} - \mathbb{E}\widehat{s}^\mathrm{ag}_{t,a''} )^2  - (\widehat{s}^\mathrm{ag}_{t,a'} - \widehat{s}^\mathrm{ag}_{t,a''})^2 + 2(\overline{C} - \underline{C})(\mathbb{E} \widehat{s}^\mathrm{ag}_{t,a'} - \mathbb{E}\widehat{s}^\mathrm{ag}_{t,a''}  -  \widehat{s}^\mathrm{ag}_{t,a'} + \widehat{s}^\mathrm{ag}_{t,a''})}{2 (\overline{C} - \underline{C})^2} \Big | E \cap F \right ] \\
    &\leq \sum_{a'' \in \mathcal{A}} \sum_{a' \in \mathcal{A}} \left[ \frac{\mathbb{E} \widehat{s}^\mathrm{ag}_{t,a'} - \mathbb{E}\widehat{s}^\mathrm{ag}_{t,a''} -  \widehat{s}^\mathrm{ag}_{t,a'} + \widehat{s}^\mathrm{ag}_{t,a''}}{\overline{C} - \underline{C}} \Big | E \cap F \right ] \allowdisplaybreaks \\ 
    &\leq \sum_{a'' \in \mathcal{A}} \sum_{a' \in \mathcal{A}} \frac{2\varphi_t}{\overline{C} - \underline{C}} \allowdisplaybreaks \\ 
    &= \frac{2n^2\varphi_t}{\overline{C} - \underline{C}} \label{eq:validate3}
\end{align}
For the second and third terms in (\ref{eq:validate2-1}) (which are essentially same with each other), we follow the same arguments as in between (\ref{eq:agent-hoeffding-1})-(\ref{eq:agent-hoeffding-last}) from the proof of Lemma \ref{lem:validate1}. Then, we get 
\begin{align}
    \sum_{a'' \in \mathcal{A}} \sum_{a' \in \mathcal{A}} \mathbb{P}(\overline{E})  + \mathbb{P}(\overline{F})  &= \sum_{a'' \in \mathcal{A}} \sum_{a' \in \mathcal{A}}  \mathbb{P} \left( \widehat{s}^\mathrm{ag}_{t,a''} - \mathbb{E}\widehat{s}^\mathrm{ag}_{t,a''} \geq \varphi_t \right) + \mathbb{P} \left( \mathbb{E}\widehat{s}^\mathrm{ag}_{t,a'} - \widehat{s}^\mathrm{ag}_{t,a'} \geq \varphi_t  \right) \allowdisplaybreaks \\
    &\leq 2n^2 \exp\left( \frac{-\frac{m^{\mathrm{ag}}}{n}\sqrt{t}\varphi_t^2 + \frac{2(m^{\mathrm{ag}})^2}{n} \varphi_t^2}{4(R_{\max} - R_{\min})^2}\right) + \frac{2n^2}{t-1} \allowdisplaybreaks \\
    \intertext{Suppose $\varphi_t = \dfrac{2\sqrt{n}(R_{\max} - R_{\min})\sqrt{\log 2t}}{\sqrt{m^{\mathrm{ag}}}\sqrt[4]{t}}$. Then,}
    &\leq 2n^2\exp\Big(-\log 2t + 2m^{\mathrm{ag}}\frac{\log 2t}{\sqrt{t}}\Big)  + \frac{2n^2}{t-1} \allowdisplaybreaks \\
    &\leq 2n^2\exp\big(-\log 2t + 2m^{\mathrm{ag}}\big) + \frac{2n^2}{t-1} \allowdisplaybreaks \\
    &= 2n^2\frac{(\exp(2m^{\mathrm{ag}}) + 1)}{t-1} \allowdisplaybreaks
\end{align} 
For the same choice of $\varphi_t$ above, we combine the last result with (\ref{eq:validate3}) and obtain
\begin{align}
    &\mathbb{P}\left(\upsilon_t(\boldsymbol{\pi}_t) \neq \argmax_{a \in \mathcal{A}} s^0_a + \pi_{t, a}  \Big | t \in \mathcal{T}^{\mathrm{ag-xploit}} \cap \mathcal{T}^{\mathrm{pr-xplore}}\right) \notag \allowdisplaybreaks \\
    &\leq \frac{4n^2\sqrt{n}(R_{\max} - R_{\min})\sqrt{\log 2t}}{\sqrt{m^{\mathrm{ag}}}(\overline{C} - \underline{C})\sqrt[4]{t}} + \frac{2n^2(\exp(2m^{\mathrm{ag}}) + 1)}{t-1} \\
    &\leq \frac{4n^2\sqrt{n}\sqrt{\log 2t}}{\sqrt{m^{\mathrm{ag}}}\sqrt[4]{t}} + \frac{2n^2(\exp(2m^{\mathrm{ag}}) + 1)}{t-1}
\end{align}
where the last inequality follows since $R_{\max} - R_{\min} < \overline{C} - \underline{C}$ by Assumption \ref{assm1}. 
\Halmos \endproof

\proof{\textbf{Proof of Proposition \ref{prop:validate}.}}
We prove this result by using the induction technique. 

\underline{\textit{Base Case:}} Consider $t = 1$. We have $1 \leq (\text{constant} )\sqrt{\log 2}$ for any $(\text{constant}) \geq 1.2$. Then, since $p_1 \leq 1$ by definition, it trivially satisfies Proposition \ref{prop:validate}.

\underline{\textit{Induction Step:}} Consider any $t \in [\widetilde{k}, T]$. Suppose that Proposition \ref{prop:validate} holds for all $p_{\tau}$ such that $\tau \in [1, t-1]$. Then, we have 
\begin{align}
p_t &=\mathbb{P}\left(\upsilon_t(\boldsymbol{\pi}_t) \neq \argmax_{a \in \mathcal{A}} s^0_a + \pi_{t, a} \right) \allowdisplaybreaks  \\
    &= \mathbb{P}\left(\upsilon_t(\boldsymbol{\pi}_t) \neq \argmax_{a \in \mathcal{A}} s^0_a + \pi_{t, a}  \Big | t \in \mathcal{T}^{\mathrm{ag-xploit}} \cap  \mathcal{T}^{\mathrm{pr-xploit}} \right) \mathbb{P}\left(t \in \mathcal{T}^{\mathrm{ag-xploit}} \cap  \mathcal{T}^{\mathrm{pr-xploit}} \right) \nonumber \allowdisplaybreaks  \\ 
    &\quad + \mathbb{P}\left(\upsilon_t(\boldsymbol{\pi}_t) \neq \argmax_{a \in \mathcal{A}} s^0_a + \pi_{t, a} \big | t \in \mathcal{T}^{\mathrm{ag-xploit}} \cap  \mathcal{T}^{\mathrm{pr-xplore}} \right) \mathbb{P}\left(t \in \mathcal{T}^{\mathrm{ag-xploit}} \cap  \mathcal{T}^{\mathrm{pr-xplore}} \right) \nonumber \allowdisplaybreaks  \\ 
    &\quad + \mathbb{P}\left(\upsilon_t(\boldsymbol{\pi}_t) \neq \argmax_{a \in \mathcal{A}} s^0_a + \pi_{t, a} \big | t \in \mathcal{T}^{\mathrm{ag-xplore}}\right) \mathbb{P}\left(t \in \mathcal{T}^{\mathrm{ag-xplore}}\right)  \allowdisplaybreaks  \\ 
    &\leq \mathbb{P}\left(\upsilon_t(\boldsymbol{\pi}_t) \neq \argmax_{a \in \mathcal{A}} s^0_a + \pi_{t, a}  \Big | t \in \mathcal{T}^{\mathrm{ag-xploit}} \cap  \mathcal{T}^{\mathrm{pr-xploit}} \right)  \nonumber \allowdisplaybreaks  \\
    &\quad +  \mathbb{P}\left(\upsilon_t(\boldsymbol{\pi}_t) \neq \argmax_{a \in \mathcal{A}} s^0_a + \pi_{t, a} \big | t \in \mathcal{T}^{\mathrm{ag-xploit}} \cap  \mathcal{T}^{\mathrm{pr-xplore}} \right) \mathbb{P}\left(t \in \mathcal{T}^{\mathrm{pr-xplore}}\right) \notag \allowdisplaybreaks  \\ 
    &\quad +  \mathbb{P}\left(t \in \mathcal{T}^{\mathrm{ag-xplore}}\right)  \allowdisplaybreaks  \\ 
    &\leq \frac{2n^3 \left(\exp\Big(\frac{2(m^{\mathrm{ag}})^2B^2}{4n(R_{\max} - R_{\min})^2}\Big) + 1\right)}{t-1} \notag \allowdisplaybreaks \\
    &\quad + 8n^2\exp \left(-\frac{2\lambda_t^2}{(t-1)16n(6R_{\max} - 6R_{\min} + 2\gamma)^2} - \log \frac{\beta_t}{2} + n \log(2R_{\max} - 2R_{\min})\right)    \nonumber \allowdisplaybreaks  \\ 
    &\quad + \left( \frac{4n^2\sqrt{n}\sqrt{\log 2t}}{\sqrt{m^{\mathrm{ag}}}\sqrt[4]{t}} + \frac{2n^2(\exp(2m^{\mathrm{ag}}) + 1)}{t-1}\right) \frac{m^{\mathrm{pr}}}{t^{1/2 - w}}  +  \frac{m^{\mathrm{ag}}}{\sqrt{t}}\label{eq:pt-prop-0} \allowdisplaybreaks 
\end{align}
where the last inequality follows by the results of Lemma \ref{lem:validate1} and Lemma \ref{lem:validate2} and by the constructions of Algorithm \ref{alg:pr} and Algorithm \ref{alg:ag}. 

Further, we can bound the second term in the right-hand side of the last inequality by following the same arguments as in between (\ref{eq:lambda_proof_0})-(\ref{eq:lambda_proof_last}) and obtain 
\begin{align}
    &\exp \left(-\frac{2\lambda_t^2}{(t-1)16n(6R_{\max} - 6R_{\min} + 2\gamma)^2} - \log \frac{\beta_t}{2} + n \log (2R_{\max} - 2R_{\min})\right) \leq \frac{2^{n+1}}{3^{n + 1} k \sqrt[6]{32n}}  \frac{1}{\sqrt{t \log 2t}} \allowdisplaybreaks 
\end{align}
Combining this upper bound with (\ref{eq:pt-prop-0}), we obtain the following upper bound on $p_t$. 
\begin{align}
    p_t = \mathbb{P}\left(\upsilon_t(\boldsymbol{\pi}_t) \neq \argmax_{a \in \mathcal{A}} s^0_a + \pi_{t, a} \right) &\leq \frac{2n^3 \left(\exp\Big(\frac{2(m^{\mathrm{ag}})^2B^2}{4n(R_{\max} - R_{\min})^2}\Big) + 1\right)}{t-1} +  \frac{2^{n+4}n^{11/6}}{3^{n + 1} k \sqrt[6]{32}}  \frac{1}{\sqrt{t \log 2t}}  \notag \allowdisplaybreaks \\
    &\quad + \left( \frac{4n^2\sqrt{n}\sqrt{\log 2t}}{\sqrt{m^{\mathrm{ag}}}\sqrt[4]{t}} + \frac{2n^2(\exp(2m^{\mathrm{ag}}) + 1)}{t-1}\right) \frac{m^{\mathrm{pr}}}{t^{1/2 - w}}  +  \frac{m^{\mathrm{ag}}}{\sqrt{t}} \allowdisplaybreaks  
\end{align}
which implies $p_t = O\Big(\dfrac{\sqrt{\log 2t}}{\sqrt{t}}\Big)$.

\underline{\textit{Conclusion:}} As both the base case and the inductive step have been proven as true, Proposition \ref{prop:validate} is established by mathematical induction.
\Halmos \endproof
\section{Parameters for Numerical Experiments} \label{appendix4}

In our numerical experiments, we test the performance of our data-driven incentive policy for different values of the cardinality of the aggregator's MAB model ($n = |\mathcal{A}|$). The closed intervals that the expected rewards of the aggregator and the utility company are set to $\Theta = [0, 100]$ and $\mathcal{R} = [-20, 50]$, respectively, and the entries of the expected reward vectors $\mathbf{r}^0$ and $\boldsymbol{\theta}^0$ are uniformly randomly generated from these intervals as reported below. 
\begin{table}[h]
\begin{center}
\begin{tabular}{|c | c | c |}  
 \hline
 $n$ & $\boldsymbol{\theta}^0$ & $\mathbf{r}^0$  \\ [0.5ex] 
 \hline
 5 &  (29, 1, 14, 26, 15) & (14, -24, -4, 19, 29)   \\ 
 \hline
 10 & (0, 44, 51, 65, 9, 35, 69, 91, 51, 44)  &  (-4, 8, 22, -12, -2, 46, -8, 16, 38, 14)  \\
 \hline
\end{tabular}
\end{center}
\caption{Expected rewards in different settings of the size of the aggregator's MAB model}
\label{table:parameters}
\end{table}

\end{APPENDICES}

\end{document}